\documentclass[lettersize,journal]{IEEEtran}
\usepackage{amsmath,amsfonts}
\usepackage{algorithmic}
\usepackage{array}
\usepackage[caption=false,font=normalsize,labelfont=sf,textfont=sf]{subfig}
\usepackage{textcomp}
\usepackage{stfloats}
\usepackage{url}
\usepackage{verbatim}
\usepackage{graphicx}
\usepackage{soul}
\usepackage{hyperref}
\usepackage[table]{xcolor}
\usepackage[sort&compress,numbers]{natbib}

\newcommand{\ApproxSign}{\raise.17ex\hbox{$\scriptstyle\sim$}}

\usepackage{float}

\usepackage{csquotes}
\usepackage{multirow}
\usepackage{amsmath}

\usepackage{tikz}

\usepackage{mathtools}

\usepackage{array}
\newcolumntype{C}[1]{>{\centering\let\newline\\\arraybackslash\hspace{0pt}}m{#1}}
\def\BibTeX{{\rm B\kern-.05em{\sc i\kern-.025em b}\kern-.08em
    T\kern-.1667em\lower.7ex\hbox{E}\kern-.125emX}}
\usepackage{balance}
\begin{document}

%\title{Hardware Acceleration of Transformers: A Survey}

\title{A Survey of Techniques for Optimizing Transformer Inference}

%A Survey of Techniques for Optimizing Inference of Transformers

\author{Krishna Teja Chitty-Venkata$^1$$\star$, Sparsh Mittal$^2$$\star$,  Murali Emani$^3$, \\ Venkatram Vishwanath$^3$ and Arun K. Somani$^1$ \\
$^1$ Iowa State Univeristy, Ames, IA, USA\\
$^2$ Indian Institute of Technology Roorkee, Uttarakhand, India\\
$^3$ Argonne National Laboratory, Lemont, IL, USA\\
krishnat@iastate.edu, sparsh.mittal@ece.iitr.ac.in, memani@anl.gov, venkat@anl.gov, arun@iastate.edu \\

\thanks{$\star$Equal Contribution}
%\thanks{Manuscript received April 19, 2021; revised August 16, 2021.}

}

%\thanks{Manuscript created October, 2020; This work was developed by the IEEE Publication Technology Department. This work is distributed under the \LaTeX \ Project Public License (LPPL) ( http://www.latex-project.org/ ) version 1.3. A copy of the LPPL, version 1.3, is included in the base \LaTeX \ documentation of all distributions of \LaTeX \ released 2003/12/01 or later. The opinions expressed here are entirely that of the author. No warranty is expressed or implied. User assumes all risk.}

%\newcommand{\ask}[1]{{\color{red}#1 }}

\markboth{Journal of \LaTeX\ Class Files,~Vol.~18, No.~9, September~2020}
{How to Use the IEEEtran \LaTeX \ Templates}

\maketitle
\begin{abstract}
Recent years have seen a phenomenal rise in performance and applications of transformer neural networks. The family of transformer networks, including Bidirectional Encoder Representations from Transformer (BERT), Generative Pretrained Transformer (GPT) and Vision Transformer (ViT), have shown their effectiveness across Natural Language Processing (NLP) and Computer Vision (CV) domains. Transformer-based networks such as ChatGPT have impacted the lives of common men. 
However, the quest for high predictive performance has led to an exponential increase in transformers' memory and compute footprint. Researchers have proposed techniques to optimize transformer inference at all levels of abstraction. This paper presents a comprehensive survey of techniques for optimizing the inference phase of transformer networks. 
We survey techniques such as knowledge distillation, pruning, quantization, neural architecture search and lightweight network design at the algorithmic level. 
We further review hardware-level optimization techniques and the design of novel hardware accelerators for transformers. We summarize the quantitative results on the number of parameters/FLOPs and accuracy of several models/techniques to showcase the tradeoff exercised by them. 
We also outline future directions in this rapidly evolving field of research. We believe that this survey will educate both novice and seasoned researchers and also spark a plethora of research efforts in this field. 
\end{abstract}


%\section{Introduction}

\section{Introduction} \label{sec:introduction}
Artificial intelligence (AI) has achieved tremendous success in a wide range of applications due to its automatic representation capability. The global AI market was valued at USD 136B in 2022 and is expected to reach USD 1,591B by 2030 \cite{precedenceresearch}. The availability of large datasets, efficient network design and hardware architecture optimization have driven this progress. The advancements in architectural design and the development of innovative topologies such as convolutional neural networks (CNNs), recurrent neural networks (RNNs), graph neural networks, and transformers \cite{vaswani2017attention} have pushed its applications into interdisciplinary domains.

By virtue of modeling long-range dependencies, transformers \cite{vaswani2017attention} have achieved state-of-the-art performance on various Natural Language Processing (NLP) and Computer Vision (CV) tasks. The field of NLP has advanced  significantly due to the emergence of large-scale Pretrained Language Models, which include Bidirectional Encoder Representations from Transformer (BERT) and Generative Pre-trained Transformer (GPT). These models have improved the efficiency of NLP tasks and also enabled new applications, including ChatGPT \cite{schulman2022chatgpt}, BARD \cite{Google_BARD} and content generation. In fact, researchers have recently used Large Language Models (LLMs) \cite{zvyagin2022genslms} to identify potential COVID-19 variants of concerns. Similarly, vision-transformer (ViT) \cite{dosovitskiy2020image}, and subsequent models have shown remarkable effectiveness on computer vision tasks such as the image classification \cite{liu2021swin} and object detection \cite{carion2020end}, and have outperformed CNNs.

\begin{figure*}[h]
    \centering
    \includegraphics[scale=0.55]{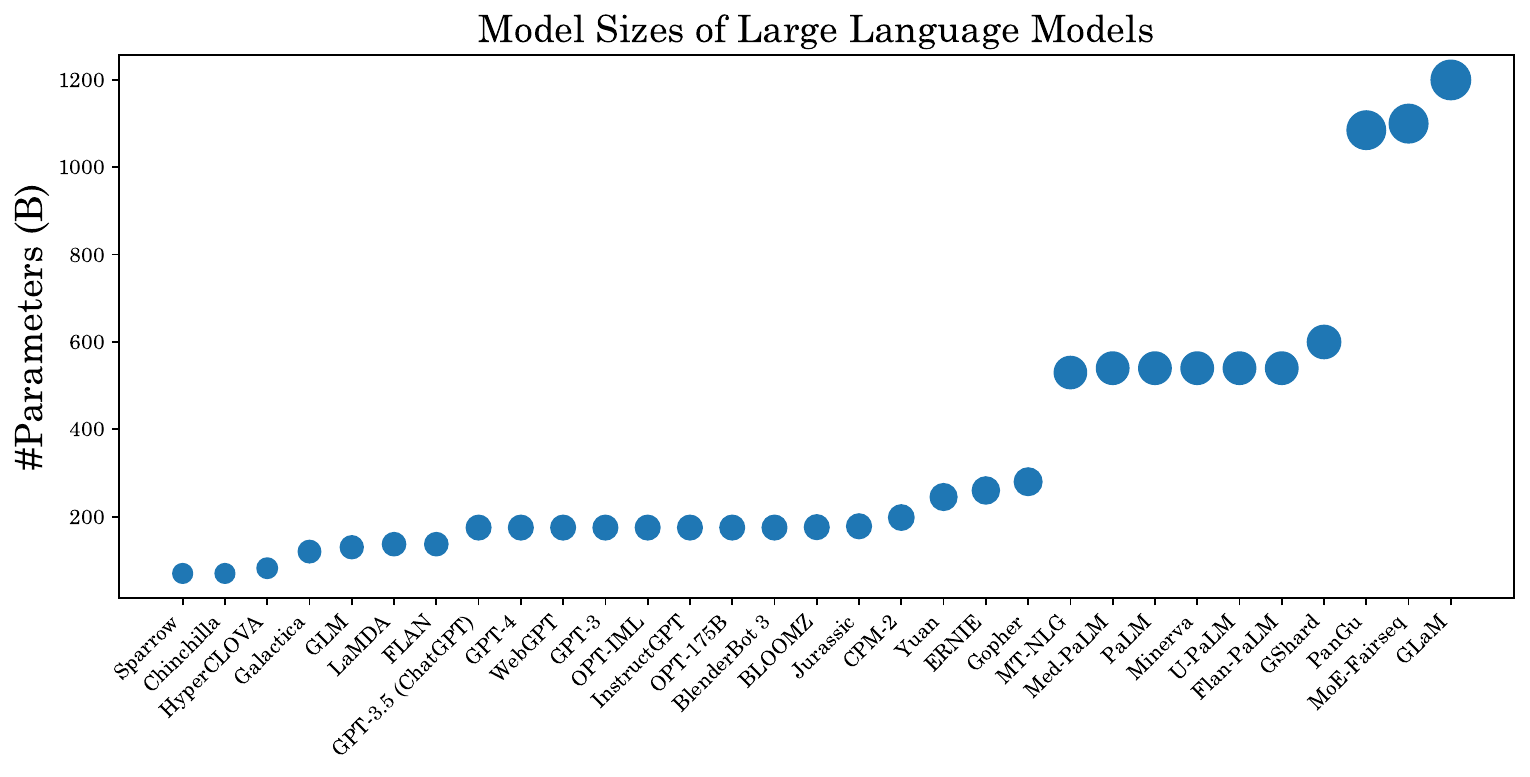}
    \captionsetup{justification=centering}
    \caption{Model size of SOTA large language models (Sparrow \cite{glaese2022improving},
Chinchilla \cite{hoffmann2022training}, 
HyperCLOVA \cite{kim2021changes},
Galactica \cite{taylor2022galactica},
GLM \cite{zeng2022glm},
LaMDA \cite{thoppilan2022lamda}, 
FLAN \cite{wei2021finetuned}, 
GPT-3.5 (ChatGPT) \cite{brown2020language}, 
GPT-4 \cite{brown2020language}, 
WebGPT \cite{nakano2021webgpt},
GPT-3 \cite{brown2020language}, 
OPT-IML \cite{iyer2022opt}, 
InstructGPT \cite{ouyang2022training},
OPT-175B \cite{zhang2022opt}, 
BlenderBot 3 \cite{shuster2022blenderbot}, 
BLOOMZ \cite{muennighoff2022crosslingual},
Jurassic \cite{lieber2021jurassic},
CPM-2 \cite{zhang2021cpm},
Yuan \cite{wu2021yuan},
ERNIE \cite{sun2021ernie},
Gopher \cite{rae2021scaling}, 
MT-NLG \cite{smith2022using}, 
Med-PaLM \cite{singhal2022large}, 
PaLM \cite{chowdhery2022palm}, 
Minerva \cite{lewkowycz2022solving}, 
U-PaLM \cite{tay2022transcending}, 
Flan-PaLM \cite{chung2022scaling}, 
GShard \cite{lepikhin2020gshard},
PanGu \cite{ren2023pangu},
MoE-Fairseq \cite{artetxe2021efficient}, 
GLaM \cite{du2022glam})}
    \label{fig:LLM_motivation}
\end{figure*}

The enhancement in predictive performance and scope of application has come at the cost of a steep increase in memory and computation overheads. Figure \ref{fig:LLM_motivation} illustrates the number of parameters for state-of-the-art (SOTA) language models. Clearly, SOTA models have up to 1.2 trillion parameters! The sizes will increase even further as more powerful hardware platforms are developed. 
ChatGPT inference consumes 500ml water for a simple conversation of nearly 50 questions and answers \cite{li2023making}.
Also, recent work has shown that vision transformers can be scaled up to 22 billion model parameters \cite{dehghani2023scaling}.

These factors call for efficient model compression techniques and hardware acceleration methods to facilitate the deployment and usage of such large models in practical settings. Additionally, given the high computational cost associated with training and fine-tuning large models, there is a growing demand for more robust and scalable computing infrastructure.

These challenges have motivated researchers to propose techniques for reducing transformers' size, latency and energy consumption for efficient inference for a wide range of applications. The methods include pruning \cite{han2015learning, han2015deep}, Quantization \cite{zhou2016dorefa, jacob2018quantization, krishnamoorthi2018quantizing}, Knowledge Distillation \cite{hinton2015distilling} and Neural Architecture Search \cite{chen2021autoformer}. These methods allow better scalability and environment-friendliness. Orthogonal to advances in model compression, the design of hardware architecture tailored for transformers is a promising solution to overcome the computational limitations of the transformer models. This involves identifying the computational bottlenecks in the transformer model, such as the self-attention operator and fully connected network, and developing hardware architectures that can accelerate these modules. 
This can be accomplished through efficient mapping of transformer models on FPGAs and ASICs and through optimization techniques such as parallelization, pipelining and avoiding redundant/ineffectual computations.

\textbf{Scope and outline of this paper:} In this paper, we survey several optimization methods for \textit{efficient inference} of transformer architectures and their family of architectures, such as BERT, GPT, and ViT. We discuss the challenges, advances and future opportunities in this ever-growing space of transformer research, whose goal is to reduce inference time, minimize memory requirements, and enhance hardware performance. To provide a comprehensive synopsis of key advances, we limit our discussion to inference-related optimizations and, thus, exclude training-related techniques. We also forecast possible future directions in this fast-evolving field of research.
The following list summarizes different dimensions of transformer optimization/compression/acceleration methods and provides high-level definitions and the paper organization:

1. In Section \ref{sec:Transformer}, we provide a background on the fundamentals of the transformer model, including embedding, general attention and multi-headed attention (MHA). We also discuss the networks used in NLP and computer vision domains, such as BERT, GPT and vision transformer. 

2. Section \ref{sec:motivation} presents several motivating factors and challenges for optimizing transformer models. The motivating factors include increasing model size and the need for improved performance. The challenges include the availability of computing resources and transformer-specific data/weight distribution. 

3. Knowledge Distillation (KD) is a model compression technique where a relatively small student model is trained to mimic the behavior of a large pre-trained teacher network. For example, using KD, DistilBERT \cite{sanh2019distilbert} compresses the BERT-base model by 40\% while retaining 97\% of its language capabilities. In Section \ref{sec:Knowledge_Distillation}, we first present an overview of distillation methods and distillation loss functions and then summarize KD techniques for transformers. 

4. The transformer models are often large and heavily over-parameterized \cite{michel2019sixteen}. Pruning refers to the process of identifying and removing redundant or unimportant parameters in such a way that the predictive performance is minimally affected. For instance, oBERT \cite{kurtic2022optimal} compresses the BERT model and attains 10$\times$ inference speedup on Intel CPU with less than 1\% accuracy drop. In Section \ref{sec:Pruning}, we first provide a taxonomy of pruning schemes and then review pruning techniques organized along several categories, such as weight, node, neuron, filter, head, and token pruning. We also review post-training pruning techniques and hardware-aware pruning techniques. 

5. During the training process, the weights and activations are generally stored in 32-bit floating-point precision. However, the inference can be performed at a lower precision, such as an 8-bit integer. Quantization reduces their precision/bitwidth to 16-bit, 8-bit or even 1-bit. Thus, while pruning reduces the number of parameters, quantization reduces the storage precision of each parameter. For example, Q8BERT \cite{zafrir2019q8bert} quantizes the weights and activations of the BERT model from 32-bit precision to 8 bits, thereby achieving model size reduction by 4$\times$ without compromising accuracy. In Section \ref{sec:Quantization}, we present a comprehensive discussion on quantization procedures, the taxonomy of transformer quantization, and binarization methods, followed by summarizing prominent transformer, BERT and ViT-centric low-precision acceleration methods. 

6. MHA operation has quadratic time complexity, and hence, it is the crucial performance bottleneck in a transformer. Several methods have been proposed to simplify this operation. MobileViT \cite{mehta2021mobilevit} is an example of such lightweight ViT, which attains six percentage points better accuracy than the DeiT model \cite{touvron2021training}, with 3.4M fewer parameters. 
Section \ref{sec:MCU_Mobile} summarizes such efficient and lightweight architectural design methods for NLP and vision applications. We further analyze the accuracy vs. parameter counts of several tiny ViT models.

7. Neural Architecture Search (NAS) is a process of automating the design process of a neural architecture for the given application and dataset. Hardware-aware NAS (HW-NAS) searches for a network with the highest possible accuracy on a dataset and compute-performance on a target hardware. For instance, Hardware-aware Transformers (HAT) \cite{wang2020hat} developed a methodology to search transformer models which have better validation metrics than original transformer \cite{vaswani2017attention} while having lower latency on CPU, GPU and mobile platforms. In Section \ref{sec:NAS}, we first present an overview of general NAS methods, followed by classifying the transformer NAS and HW-NAS methods based on search space, and search technique. We then review the use of search methods for model compression.   
 
8. The high computational demands of transformers calls for hardware optimization techniques and designs of novel hardware accelerators. For example, a hardware-unaware pruning technique may compress a model to 0.2$\times$ the original size. Yet, such a model is unlikely to provide a 5$\times$ reduction in latency, memory accesses or energy on conventional computing systems. In fact, due to its random sparsity patterns, such a pruned model may forgo vectorization and tiling and hence, incur higher latency than the uncompressed model. Similarly, while approximate computing requires fewer operations than exact computation, the former incurs higher latency on a GPU \cite{ham2021elsa}. Evidently, there is a need to synergistically design the processing system and transformer to obtain optimal performance in both worlds. For example, novel dataflows can expose reuse opportunities and structured pruning techniques can lead to hardware-friendly memory accesses Section \ref{sec:HardwareOptimization} reviews hardware-level techniques for compute- and memory-optimization.

\textbf{Contributions:} The three main contributions of this paper are as follows:

\textbf{1. Comprehensive overview:} We provide a high-level overview of the SOTA enhancement techniques for transformer inference, covering various network and hardware optimization strategies. To make the survey self-contained and thus useful for both beginners and seasoned researchers, we include the essential background on transformer architecture and transformer-based models. Our goal is to help readers understand the wide landscape of optimization strategies along with their challenges and limitations. Our paper is useful for both neural network enthusiasts and hardware practitioners. 

\textbf{2. Taxonomy and tradeoffs :} We provide a taxonomy of methods for optimized transformer inference based on several key factors, including the type of optimization technique, granularity within the transformer model, type of transformer architecture and domain. The categorization helps readers with a clear and organized framework for understanding different approaches and allows them to easily identify and compare different optimization techniques and understand each approach's strengths and weaknesses. We discuss the practical considerations for each optimization technique, such as the tradeoffs between accuracy and efficiency.
In addition to qualitative insights, we also present quantitative results on the number of parameters/FLOPs and the accuracy of several optimization techniques. This provides insights into the tradeoff exercised by those techniques. 

\textbf{3. Future directions:} Finally, we identify the gaps in current literature and promising future research directions, such as developing more efficient hardware architectures, investigating the benefits of co-design, combining different optimization techniques and the need for novel benchmarks. Overall, this paper will be a valuable resource for the research community and industry practitioners seeking to optimize transformer inference efficiency for real-world deployment.

 In this paper, we use ``predictive performance'' to refer to metrics such as accuracy and compute performance to refer to latency/energy/power metrics. Unless mentioned otherwise, performance refers to predictive performance. We use ViT to refer to the vision transformer proposed by Dosovitskiy et al. \cite{dosovitskiy2020image}. We use ``CV transformer'' and ``NLP transformer'' to refer to the broad family of transformers in CV and NLP areas, respectively.

\section{Background on transformer networks} \label{sec:Transformer}
\begin{figure}
    \centering
\includegraphics[scale=0.24]{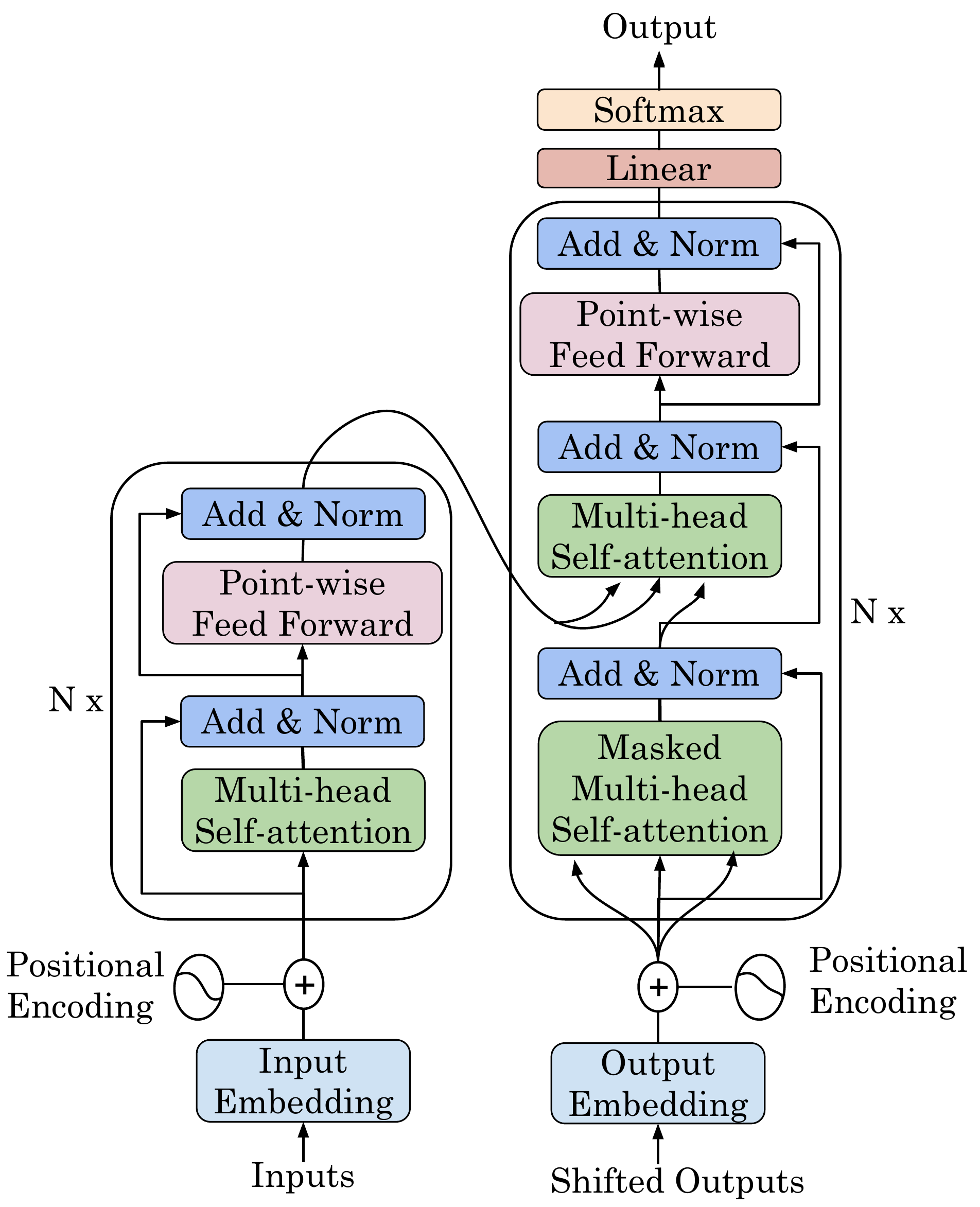}
    \captionsetup{justification=centering}
    \caption{Transformer architecture \cite{vaswani2017attention}}
    \label{fig:transformer}
\end{figure}

%Self-Attention and Transformer Model
The transformer model \cite{vaswani2017attention}  learns global dependencies in the input through attention mechanism in a pairwise correlation manner. 
The model, depicted in Figure \ref{fig:transformer}, has $N$ identical encoder and decoder modules. The primitive modules in these two units are Input and Output Embedding, Positional Embedding, MHA and Pointwise Feed-Forward Network (FFN). In this paper, the vanilla transformer refers to the transformer model with both encoder and decoder units.

\subsection{Basic Modules} \label{subsec:basic_modules}
\subsubsection{Embedding Layer} \label{subsubsec:inp_oup_embedding}
The embedding layer translates the tokens into a sequence of dense vector representation, which are fed to the attention mechanism.  

\subsubsection{Positional Embedding} \label{subsubsec:position_embedding}  
Since transformers lack recurrence or convolution operations, they need a mechanism to remember the relative positional information of the words in the input sequence \cite{dufter2022position}. The positional information is induced using sin and cosine functions at even and odd positions, respectively, in the input sequence.

\subsubsection{Self-Attention} \label{subsubsec:self_attention}

Transformer architectures rely on the self-attention mechanism, which exhibits better model parallelism compared to recurrent layers and require minimal inductive bias compared to a convolution network. This mechanism enables the model to focus on different parts of the input sequence dynamically, establishes pairwise correlation and models long-range dependencies between the elements of the input data sequence. In self-attention, the model calculates the attention weights for each position in the sequence, which reflects the importance of each position with others. This allows the model to attend to different parts of the sequence depending on the input. The input of the attention module is fed to three distinct fully-connected (FC) layers, which are learned during training, to produce Query (Q), Key (K), and Value (V) tensors. The scaled dot-product attention (A), as given in Equation \ref{Eq:scaled_dot_product}, represents the influence of each word in Query with respect to other words in the Key matrix.  
\begin{equation} \label{Eq:scaled_dot_product}
    \text{A} = \text{softmax} \left(  \frac{QK^{T}}{\sqrt{D_{k}}} \right)
\end{equation}

\begin{figure}
    \centering
\includegraphics[scale=0.25]{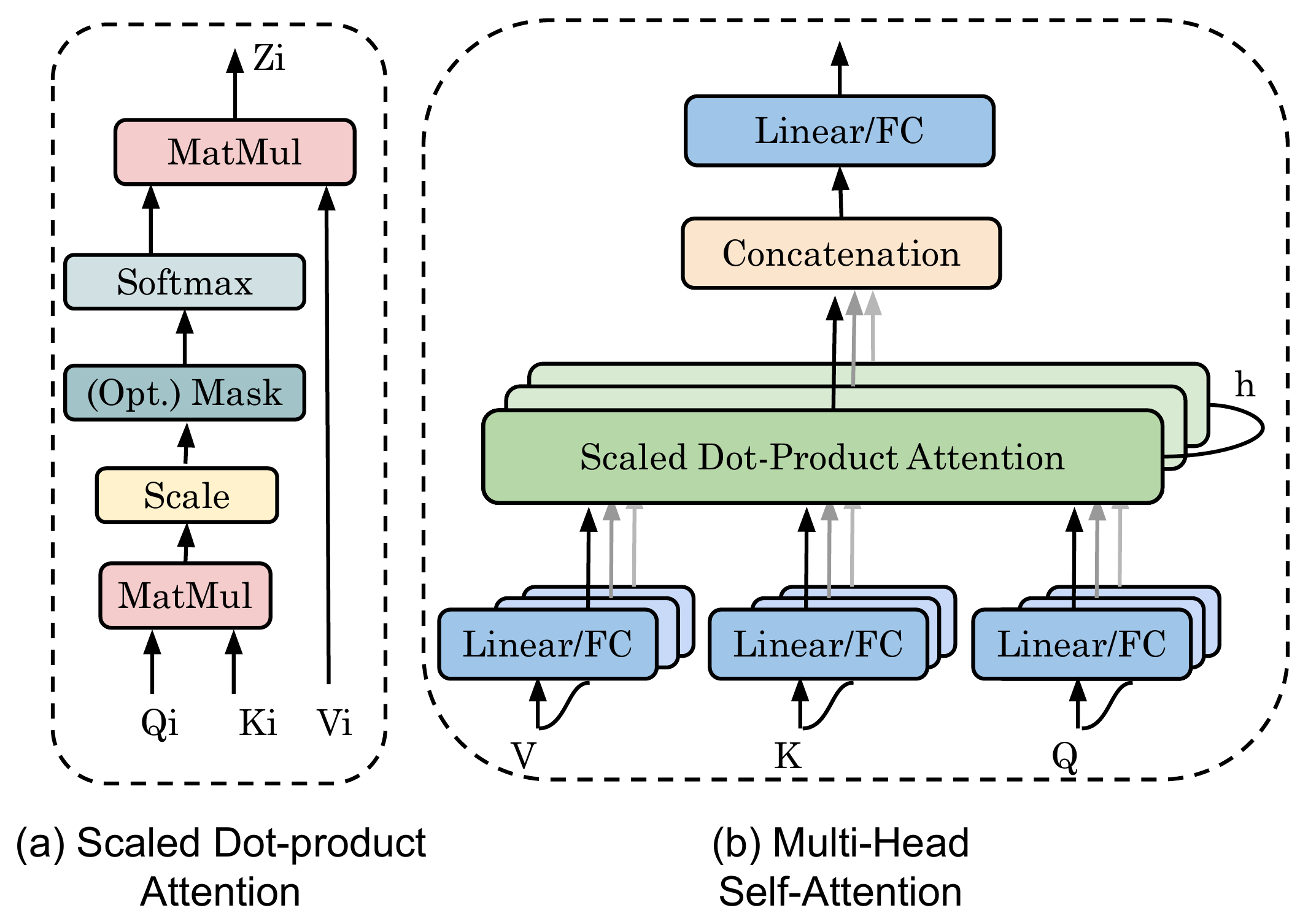}
    \captionsetup{justification=centering}
    \caption{Attention mechanism in transformers}
    \label{fig:SA_MHSA}
\end{figure}

The Query and Key are multiplied in an element-by-element manner to produce a score matrix, which is divided by $\sqrt{D_{k}}$, the square root of output dimensions of the Key matrix to alleviate the gradient vanishing problem. The softmax function boosts high score values and dampens lower score values. The attention score  is finally obtained by multiplying the attention and value matrix, as given in Equation \ref{Eq:attention_Q_K_V}. The schematic of self-attention is depicted in Figure \ref{fig:SA_MHSA}(a).

\begin{equation} \label{Eq:attention_Q_K_V}
    \text{Attention (\textbf{Q}, \textbf{K}, \textbf{V})} = AV
\end{equation}

\subsubsection{Multi-Head Self-Attention (MHA)} \label{subsubsec:MHSA}  
The MHA module includes several \enquote{heads}, each of which concurrently computes attention operations. As depicted in Figure \ref{fig:SA_MHSA}(b), the input to the MHA module is replicated across all the heads. The input (X) to a head (head$_{i}$) is processed across three FC layers (W$^{Q_{i}}$, W$^{K_{i}}$, W$^{V_{i}}$) to obtain one set of Query (Q$_{i}$), Key (K$_{i}$), and Value (V$_{i}$) vectors on each head, as per Equation \ref{Eq:MHSA_QKV}. 
\begin{equation}\label{Eq:MHSA_QKV}
    \textit{Q}_{i} = \text{XW}^{\text{Q}_{i}}, \textit{K}_{i} = \text{XW}^{\text{K}_{i}}, \textit{V}_{i} = \text{XW}^{\text{V}_{i}}
\end{equation}

The output (Z$_{i}$) of each head is computed using Q$_{i}$, K$_{i}$, and V$_{i}$ vectors through the self-attention mechanism, as per Equation \ref{Eq:MHSA_SA}.
\begin{equation}\label{Eq:MHSA_SA}
    head_{i} = \text{Self-attention} \text{(Q}_{i}, \text{K}_{i}, \text{V}_{i}\text{),  i = 1,2,..h}
\end{equation}

The independent outputs from all the heads \{head$_{1}$, head$_{2}$, ... head$_{i}$\} are concatenated depthwise and linearly transformed using an FC layer, as per Equation \ref{Eq:MHSA_concat} to produce the output of MHA module.

\begin{equation}\label{Eq:MHSA_concat}
    \text{MHA(\textit{Q}, \textit{K}, \textit{V})} = [head_{1};...head_{h}] * \textbf{W}^{O}
\end{equation}

\subsubsection{Pointwise Feed Forward Network (FFN)} \label{subsubsec:FFN}  

FFN or multi-layer perceptron (MLP) unit is a series of two fully connected (FC) layers with ReLU \cite{agarap2018deep} or GELU \cite{hendrycks2016gaussian} activation function. FFN learns position-specific information with respect to different sets of input sequences. The output of MHA is fed to pointwise FFNs, which is further processed using a normalization (Norm) operation. 

\subsection{Encoder and Decoder} \label{subsec:Encoder_and_Decoder}  

The vanilla transformer proposed by Vaswani et al. \cite{vaswani2017attention} consists of encoder and decoder modules. The encoder processes the input sequence to generate a fixed-length representation, which contains the essential details of the input data. The decoder utilizes the context of the encoder and the attention mechanism to generate an output sequence. This is used for sequence-to-sequence tasks such as machine translation.

\subsubsection{Encoder} \label{subsubsec:encoder}

The encoder and decoder modules are built by stacking identical layers, each with two sub-layers: MHA and FFN, as shown in Figure \ref{fig:transformer}. The input to the first MHA module is the positional embedding and that to subsequent modules is the output of the previous layer. The output tensor of MHA is processed further through a Normalization layer \cite{ba2016layer} and fed to the FFN block to enhance the expressiveness of the input sequence. The output vector of the FFN unit is added to the output of MHA using a residual connection and normalized to generate the encoder output.

\subsubsection{Decoder} \label{subsubsec:decoder}
The decoder follows a similar structure to the encoder and is built using an identical stack of three sub-layers. The first sub-layer is a masked MHA unit. Its operation is equivalent to MHA, except that the future positions in the sequence are masked as they are yet to be predicted by the network. The second sub-layer is a multi-headed cross-attention unit, where the output of the encoder is mixed with the output of the first sub-layer (masked MHA). This cross-attention scheme utilizes the previously generated sequence from the encoder and focuses on essential information in the sequence. The third sub-layer is an FFN, which learns position-specific information of the processed sequence, followed by an FC layer.

\subsection{Family of Transformer Architectures} \label{subsec:Transformer_family}  
Several transformer-based large models have been recently developed. The most prominent ones are BERT \cite{devlin2018bert} and GPT \cite{radford2018improving} models. These models learn universal language representation from a large unlabeled dataset and distil the knowledge to a downstream application on the labeled data. The pre-trained BERT or GPT is fine-tuned on a specific downstream application. 
NLP tasks can be divided into two categories: \textbf{(1) Discriminative}  tasks summarize an input sequence or classify a sentence. The BERT model is widely used for these kinds of tasks.
 \textbf{(2) Generative} tasks use a GPT model to summarize the input sequence and generate new tokens.

\subsubsection{BERT} \label{subsubsec:BERT}  
\begin{figure}
    \centering
    \includegraphics[scale=0.46]{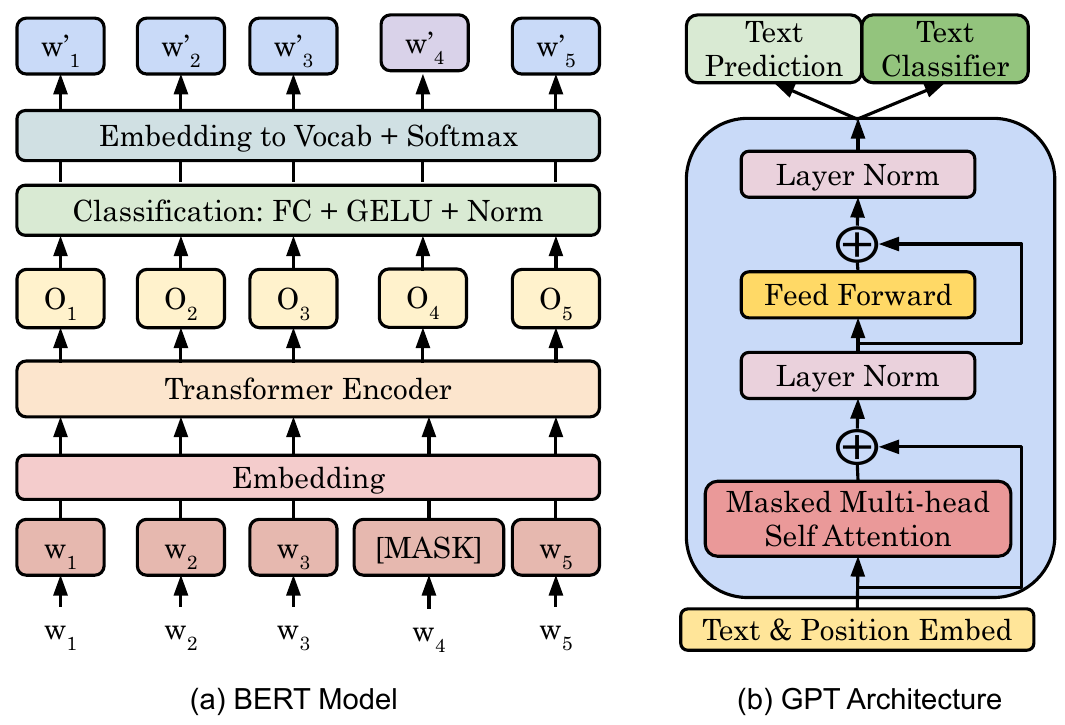}
    \captionsetup{justification=centering}
    \caption{(a) BERT and (b) GPT architectures}
    \label{fig:BERT}
\end{figure}

The early language models were designed to process text data sequentially in a unidirectional manner: from right to left or from left to right. By contrast, BERT predicts the missing data based on both the previous words and the following words in the input sequence; hence, the name bidirectional. The BERT model consists of only the encoder module of the original transformer. It masks 15\% of the words in input sequence data, as shown in Figure \ref{fig:BERT}(a). The hyperparameters of the BERT model are \textbf{(1)} the number of encoder layers (L), \textbf{(2)} hidden size (H), and \textbf{(3)} the number of attention heads (h). The BERT-base and BERT-large have the following hyperparameters: \{L = 12, H = 768, h = 12\}, and \{L = 24, H = 1024, h = 16\}, respectively.

\subsubsection{GPT} \label{subsubsec:GPT}

GPTs \cite{radford2018improving, radford2019language} are large language models (LLMs), which are pretrained in an unsupervised manner on diverse text data to perform predictive tasks. GPT retains only the decoder containing positional encoding, masked MHA, FFN, and normalization operation, as illustrated in Figure \ref{fig:BERT}(b). The variants of GPT include GPT-1, GPT-2, GPT-3 \cite{brown2020language}, etc. The GPT model is used in various real-world applications, such as ChatGPT \cite{schulman2022chatgpt}.

\subsection{Vision Transformer} 
\begin{figure}[H]
    \centering
\includegraphics[scale=0.2]{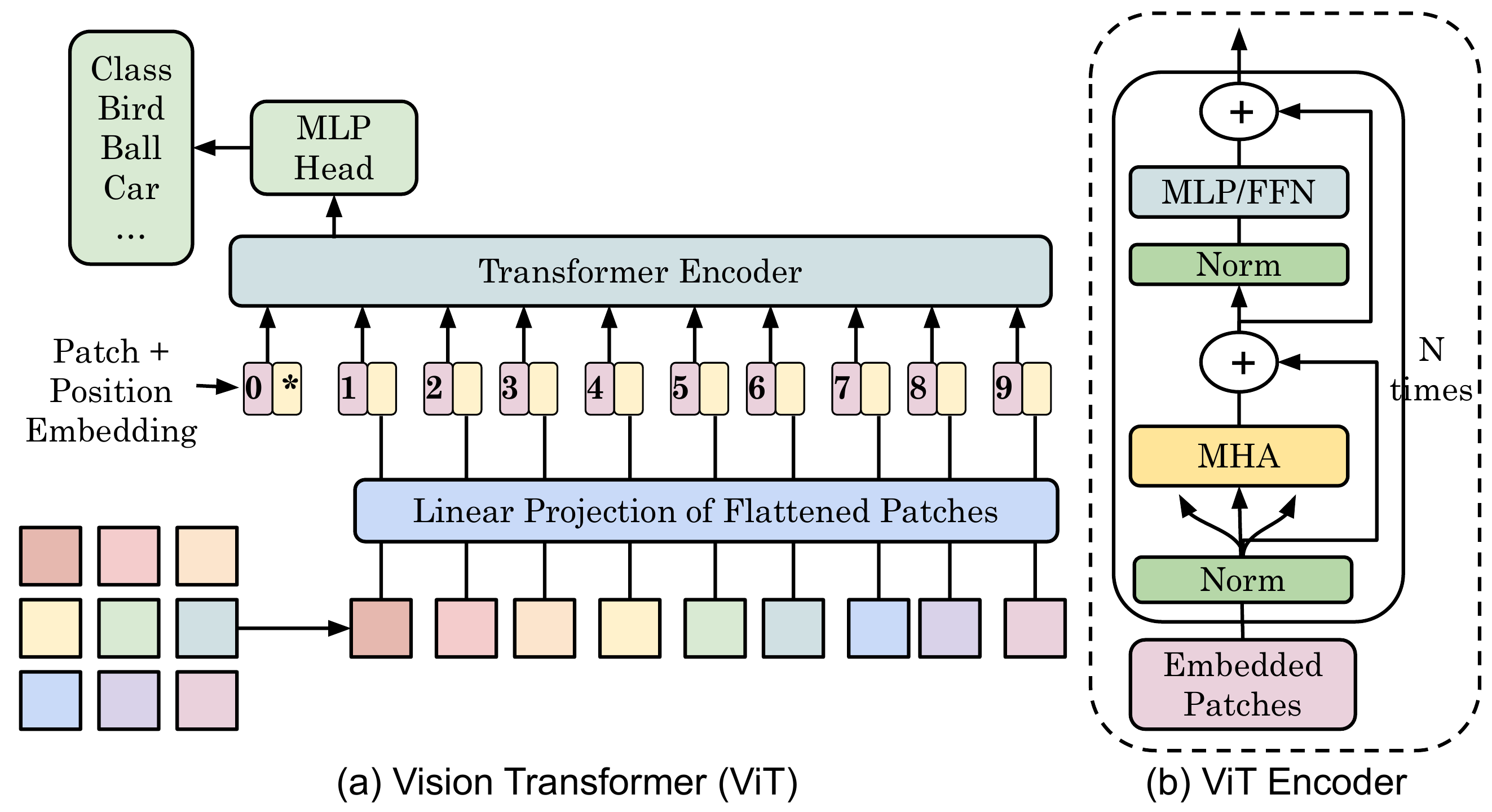}
    \captionsetup{justification=centering}
    \caption{(a) Overall architecture of ViT (b) ViT encoder}
    \label{fig:ViT_Overview}
\end{figure}

The vision transformer ViT  \cite{dosovitskiy2020image} has opened up a new area of research, focusing on using self-attention modules for computer vision tasks. 
Vision transformer models have many advantages over CNNs such as large receptive field, higher capacity to learn complex features, low inductive bias, etc. 
Unlike CNNs, which learn local representations through their spatial inductive bias, transformer models  learn  global representations through the use of the self-attention mechanism. They are also effective at modeling long-range interdependencies and can process multi-modal data such as images, videos, speech, and text. 
 The ViT model, depicted in Figure \ref{fig:ViT_Overview}(a), consists of three main modules: (1) Patch Embedding, (2) Position Embedding, and (3) Transformer Encoder.

\subsubsection{Patch and Position Embedding} \label{subsubsec:patch_embedding}
The input to a ViT is a 3D image of dimension (H$\times$W$\times$3), which is transformed into a flattened sequence of 2D patches. The ViT model splits the input image into several non-overlapping patches, each of size p$\times$p, to treat them as token embeddings. For instance, consider an image of dimension (H, W, IC), where H, W, and IC represent the the height of the input image, the width of the image, and the input channel size, respectively. The resolution of each 2D patch is (p, p), and the image is transformed into a vector of dimension (h*w, p*p*C), where H = h*p and W = h*p. The flattened projection is processed through a FC layer and passed to the next operations in the transformer. The position of each element plays an important role in better learning global information. Therefore, a 1D learnable position embedding is linearly added to the patch embeddings to preserve the spatial positional information \cite{dosovitskiy2020image}.

\subsubsection{Transformer Encoder} \label{subsubsec:Transformer_Encoder}

The vision transformer retains only the encoder module of the vanilla transformer, similar to BERT. The encoder extracts features from the input activation map. It establishes long-range dependency among the patches through the self-attention mechanism.  In the encoder module of ViT, the normalization operation is applied before MHA and FFN units, as illustrated in Figure \ref{fig:ViT_Overview}(b). 
The FFN module is a sequence of two Fully Connected layers whose output is added to the tensor before the second normalization layer through a residual connection. The final layer of ViT is an FC layer, which predicts output probabilities.

%\section{Vision Transformer} \label{sec:Vision_transformer}
%Challenges, motivation and overview
\section{Motivation and overview} \label{sec:motivation}

This section describes the motivation, necessity, and challenges faced in developing optimization methods for transformers. 

\subsection{Motivation for optimizing transformer models}
We now discuss the necessity of optimizing large-scale transformer models:

\subsubsection{Model size reduction} 
Large language models are highly demanding in terms of memory and computing resources, making them difficult to deploy in real-time applications. 
For example, BERT-base and BERT-large models have 110M and 340M parameters, respectively. Similarly, computer vision models have huge model size, e.g., the original ViT-base model consists of 86M trainable parameters \cite{chuanyangsavit}.  Techniques like SparseGPT \cite{frantar2023massive} can help in removing 100 billion parameters without any accuracy loss. Larger models also provide higher scope for compression. In other words, for a fixed target sparsity, larger models experience a much smaller accuracy drop than their smaller counterparts. For instance, the most extensive models from the OPT and BLOOM families can be pruned to 50\% sparsity with minimal increase in perplexity \cite{frantar2023massive}.
Therefore, model compression techniques can allow storing large models in limited storage capacity.
 
\subsubsection{Performance benefits} 
Model compression can improve hardware efficiency on several metrics such as latency, energy and power. The inference of a large-sized transformer requires a significant amount of computing time. A smaller model can be generally quickly loaded and executed, leading to low inference latency. 
For example, MobileBERT \cite{sun2020mobilebert} is a compressed version of BERT-base model and it runs 5.5$\times$ faster than BERT-base model on Pixel 4 mobile phone. Also, smaller models require less memory to store and run, which can benefit resource-constrained environments such as edge devices. Running smaller models requires less energy than running larger models, which can extend the battery life of mobile devices and reduce power consumption in data centers. 

\subsection{Challenges for optimizing transformer models}
Although important, optimization of transformer models presents several challenges.

\subsubsection{Need of Computing Resources} 
Developing and implementing transformer optimization techniques require significant computational resources, particularly during the finetuning phase. Finetuning a compressed or optimized model involves retraining the model on a smaller dataset, which can require several iterations of training and validation.

\subsubsection{Wider distribution of weights} 
Mao et al. \cite{mao2021tprune} illustrate the challenges in transformer pruning by comparing the weight distribution of the ResNet model on the CIFAR10 dataset with the transformer model on the WMT dataset. Their analysis revealed that the weight distribution of the transformer network is wider than that of the ResNet model, indicating that the weights of the transformer tend to be larger than those of a CNN model. This difference in weight distribution presents a significant challenge for pruning transformer models as the process requires careful consideration of the complex interdependencies among the weights. Therefore, pruning transformer requires more sophisticated techniques than CNN.

\subsubsection{Simplification prohibits generalization} 
ML models need to generalize well to new and unseen data. While simplification and compression lead to performance improvement on the target dataset, they can result in poor performance on a dataset from different domain or having different characteristics. This is because a compression technique may remove weights trained for generalization.

\subsubsection{Hardware-related challenges} 
Transformer models use hardware-unfriendly operations that hinder their efficiency and are difficult to implement on specialized hardware. Unlike CNNs, which rely on linear operations, transformer models employ a more complex architecture with many nonlinear operations, including attention mechanisms, softmax and multi-headed attention \cite{dong2023heatvit}.

\section{Knowledge distillation} \label{sec:Knowledge_Distillation}
Knowledge distillation (KD) \cite{hinton2015distilling} is a widely used model compression technique where the knowledge is transferred from a large pretrained teacher model to a small student model, so it can replicate or mimic the teacher model's behavior. 
KD methods have been effective in compressing large transformer models, such as DistilBERT \cite{sanh2019distilbert}, TinyBERT \cite{jiao2019tinybert} etc. The distilled models are smaller and faster and have comparable accuracy as the teacher model. Also, they can enhance the accuracy of the small networks on applications that need complex representations.

\begin{figure}[H]
    \centering    
    \includegraphics[scale=0.53]{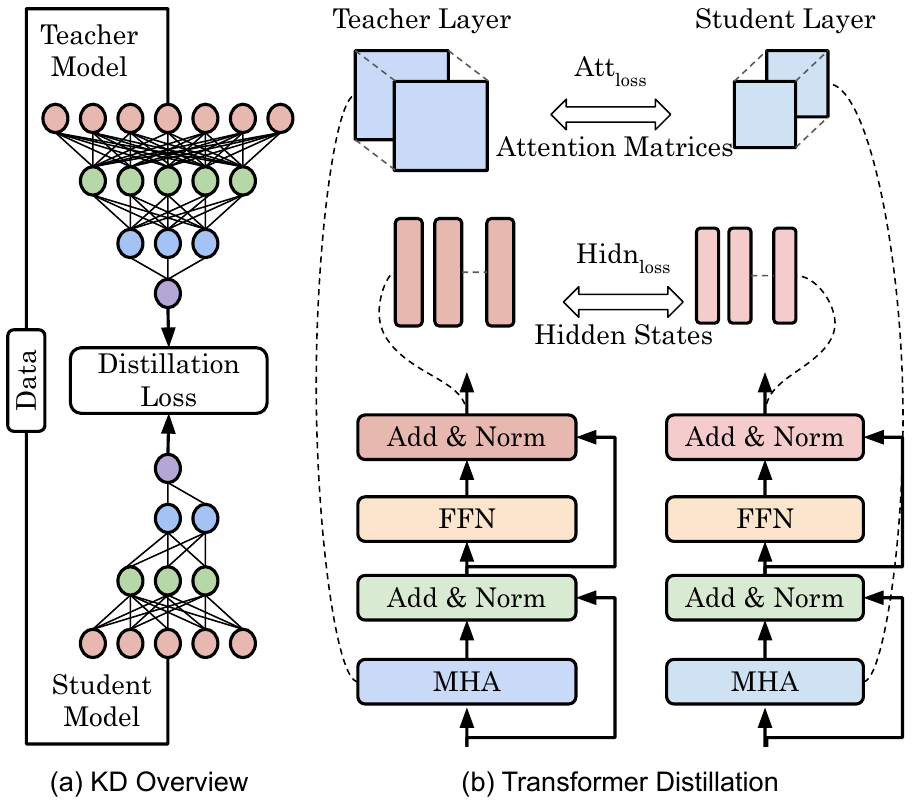}
    \captionsetup{justification=centering}
    \caption{(a) Overview of KD \cite{jiao2019tinybert} (b) transformer distillation }
    \label{fig:KD}
\end{figure}

\subsection{Overview of Knowledge Distillation Methods}
%Krishna - done-grammarly
Typically, the distillation methods utilize the teacher model's predictions to guide the student model's training. The process first creates a large neural network, and the task is to make a smaller transformer network approximate the function learned by the larger network.  
The student model is trained to predict both the correct output and the soft targets produced by the teacher model. The soft target here refers to the probabilities the teacher produces when the predictions are made for a given input. This is done by minimizing the distillation loss between the targets produced by the teacher model and the predictions produced by the student model. The overview of distillation method is depicted in Figure \ref{fig:KD}(a).

The distillation process usually employs a linear combination of two loss terms, with a hyperparameter controlling the balance between them.  
The hyperparameter controls the softness of the teacher's output probabilities, with higher values producing softer targets, making it easier for the student network to learn. The first term is usually the standard cross-entropy or any other loss function depending on the target task, and the second term measures the difference between the output probabilities of the student and the soft targets generated by the teacher network. In general, several types of loss functions exist to measure the difference between student and teacher models, such as Kullback–Leibler (KL)-divergence, mean Squared Error (MSE) and Cosine Similarity.

We now review KD methods that are used to compress the large-sized BERT and ViT models. Table \ref{tab:KDmethods} provides a classification of these methods. 

\begin{table}[htbp]
  \centering\footnotesize
  \caption{Classification of transformer KD methods}
    %\begin{tabular}{|l|p{6cm}|}
    \begin{tabular}{|l|C{5cm}|}
    \hline
    \multicolumn{2}{|c|}{Distillation loss} \\
    \hline
    KL divergence & \cite{adhikari2020exploring, sun2020mobilebert, liang2023homodistil, wang2020minilm, touvron2021training, yu2022unified} \\
    \hline
     MSE & \cite{jiao2019tinybert, hofstatter2020improving, hou2020dynabert, chen2022dearkd} \\
    \hline
    Cross-entropy & \cite{hou2020dynabert} \\
    \hline
    Cosine similarity & \cite{sanh2019distilbert} \\
    \hline
    \multicolumn{2}{|c|}{Based on task-awareness} \\
    \hline
    Task-specific & \cite{adhikari2020exploring, hou2020dynabert, touvron2021training, chen2022dearkd, jia2021efficient, yu2022unified} \\
    \hline
    Task-agnostic & \cite{sanh2019distilbert, sun2020mobilebert, liang2023homodistil, wang2020minilm, zafrir2021prune} \\
    \hline
    \multicolumn{2}{|c|}{Learning granularity} \\
    \hline
    Layer-wise & \cite{jiao2019tinybert, sun2020mobilebert, hou2020dynabert, liang2023homodistil, chen2022dearkd, jia2021efficient} \\
    \hline
    Output-wise & \cite{adhikari2020exploring, sanh2019distilbert, hofstatter2020improving, liang2023homodistil, touvron2021training, yu2022unified, zafrir2021prune} \\
    \hline
    Attention-wise & \cite{ji2021show} \\
    \hline
    \multicolumn{2}{|c|}{Network Type} \\
    \hline
    Transformer/BERT & \cite{adhikari2020exploring, sanh2019distilbert, jiao2019tinybert, hofstatter2020improving, sun2020mobilebert, hou2020dynabert, liang2023homodistil, wang2020minilm, zafrir2021prune} \\
    \hline
    Vision transformer & \cite{touvron2021training, jia2021efficient, yu2022unified} \\
    \hline
    \end{tabular}%
  \label{tab:KDmethods}%
\end{table}%

\subsection{Methods based on task-awareness}
The KD methods can be broadly divided into two categories based on the level of task-specificity of the knowledge transferred from the teacher model to the student model. They are summarized below.

\subsubsection{Task-agnostic KD}
 The task-agnostic KD refers to distilling ``generic'' knowledge, i.e.,  without considering any specific task, which can be useful for several downstream applications. Homotopic Distillation (HomoDistil) \cite{liang2023homodistil} is a task-agnostic distillation method which combines iterative pruning and layer-wise (attention-wise and hidden layer-wise) transfer learning. The student model is initialized from the teacher model and is iteratively pruned until the target width is reached. The iterative pruning method removes the least important parameters throughout the distillation process based on the importance of the parameters with respect to the final score.

\subsubsection{Task-specific KD}
 
Task-specific distillation transfers knowledge to a small model for the same downstream application. This distillation method is extremely useful and suitable for scenarios where we intend to get the best performance for the specific task, whereas task-agnostic distillation is suitable for transferring only the general knowledge and may not obtain the best performance on the target task. DeiT \cite{touvron2021training} is the first distillation method for ViT. The authors train a student transformer model to match hard labels provided by a pre-trained CNN teacher network on the target Imagenet dataset. The authors utilize only the final output of the teacher and student model while ignoring the  information of intermediate-layers  in both networks.

\subsection{Methods based on distillation granularity}
 The distillation granularity refers to the level at which information transfer happens between the teacher and student network.  As shown in Figure \ref{fig:KD}(b), the granularity can be network, layer or token. We now discuss them.

\subsubsection{Network-level Distillation} 
The network/model-level distillation transfers knowledge only at the model output level. In this method, the student network is trained to match the output of the teacher model by considering the training to minimize the loss between teacher and student models. This technique is also known as prediction-layer distillation, as the student model is trained to match the predictions.

DistilBERT \cite{sanh2019distilbert} is a pretraining method based on network-wise KD \cite{hinton2015distilling}. It generates a small general-purpose language model which can be finetuned on a wide range of applications. 
DistilBERT combines language modeling, distillation and cosine-distance losses. DistilBERT retains 97\% of the language understanding capabilities of BERT, while having 40\% lower model size and 60\% lower latency on the Intel Xeon E5-2690 CPU.
DistilGPT2 \cite{sanh2019distilbert} uses the same approach under the supervision of GPT2 and generates a compressed version of the GPT model. DistilGPT2 obtains similar performance as the GPT2 model with only 84M model parameters, as opposed to 124M parameters in the GPT2 model.

TinyBERT \cite{jiao2019tinybert} is designed using a mixture of task-agnostic and task-specific KD methods. It is a two-stage distillation method, where the first stage transfers general domain information from a large pretrained BERT model to obtain a small-sized general TinyBERT model. The general TinyBERT model acts as a teacher in the second stage and is further finetuned or distilled on the target dataset to obtain a task-specific TinyBERT model. TinyBERT with only four self-attention layers can match 96.8\% predictive performance of the teacher BERT-base network on GLUE benchmark while being 7.5$\times$ smaller. 
UVC \cite{yu2022unified} is a unified compression framework to achieve pruning, layer skipping, and KD in a single constrained optimization loop. Specifically, its prunes heads in the MHA unit and inner dimension in the FNN block. The original uncompressed ViT network provides the soft labels during the KD process.

\subsubsection{Layer-level distillation} 
The layer-level distillation refers to transferring knowledge at the level of individual layers. In this method, the student model is trained to produce similar outputs of selected layers as the teacher model. 
Hidden state-level transfer learning is a type of layer-level learning that aims to minimize the loss between the hidden states of teacher and student networks. The hidden state represents the output of MHA and FNN modules of the encoder or decoder.

\begin{figure}
    \centering
    \includegraphics[scale=0.55]{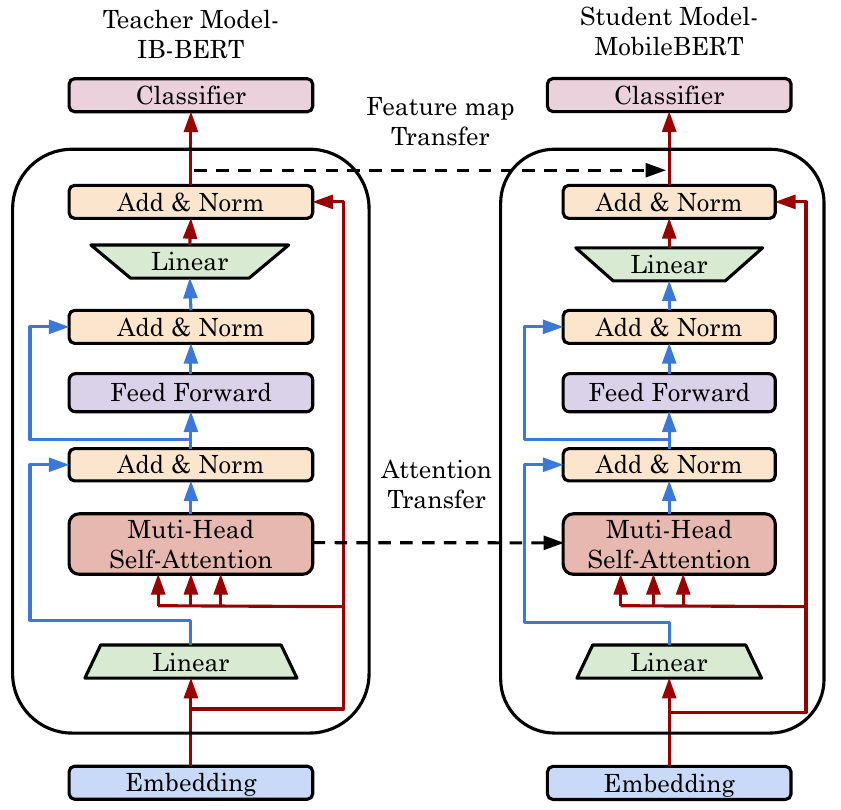}
    \captionsetup{justification=centering}
    \caption{KD approach used by Sun et al. \cite{sun2020mobilebert} to distill knowledge of IB-BERT to MobileBERT network}
    \label{fig:MobileBERT}
\end{figure}

Sun et al. \cite{sun2020mobilebert} propose a network called Inverted-Bottleneck BERT (IB-BERT). It enhances the original BERT model by adding the linear layers in each self-attention module, as shown in Figure \ref{fig:MobileBERT}. The IB-BERT model acts as a teacher model, and the knowledge is distilled to a smaller version, MobileBERT, progressively over multiple steps in a task-agnostic fashion. 
The knowledge from IB-BERT is transferred to MobileBERT in a layer-wise fashion, i.e., the attention level and hidden layer-wise independently, as depicted in Figure \ref{fig:MobileBERT}. 
The distilled MobileBERT model is 4.3$\times$ smaller than the BERT-base network and 5.5$\times$ faster on Pixel 4 mobile phone.

\begin{figure}
    \centering
    \includegraphics[scale=0.45]{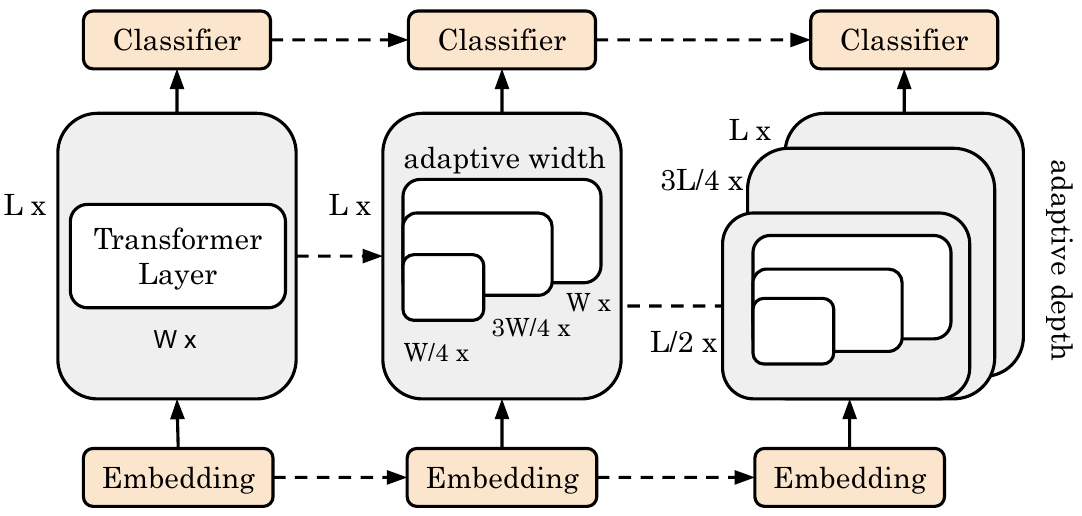}
    \captionsetup{justification=centering}
    \caption{KD approach used in DynaBERT technique \cite{hou2020dynabert} }
    \label{fig:DynaBERT}
\end{figure}

DynaBERT \cite{hou2020dynabert} first trains a width- and depth-adaptive teacher model. Then, based on this teacher model, it dynamically adjusts the width and depth of the student model to minimize the target hardware latency using KD. As illustrated in Figure \ref{fig:DynaBERT}, DynaBERT is a two-stage KD process. First, the knowledge is transferred from the large model to a width-adaptive subnetwork and then from this intermediate model to a depth-adaptive model. The distilled models achieve better language capabilities than BERT-base, RoBERTa and TinyBERT models with less latency on GPU and CPU devices. One observation from the adaptive distilled models is that the width direction is more robust to model compression than the depth direction.

 \subsubsection{Attention-based distillation}
The attention-based distillation trains the attention matrices of the student network from the teacher network, such that it transfers the linguistic information. The motivation for this method comes from BERT's capability to learn attention weights in such a way that it captures rich linguistic knowledge, which includes syntax and coreference information \cite{jiao2019tinybert}. Minilm \cite{wang2020minilm} is a deep self-attention distillation technique where the student is trained to mimic the self-attention behavior of the last layer of the BERT teacher. However, within the last self-attention layer, the task is to minimize the KL divergence between the QKV attention matrices. The attention matrix transfer learning process can be extremely useful for task-specific distillation scenarios.

\begin{table*}[htbp]
  \centering
  \caption{Classification of pruning technqiues based on saliency of network parameters: zero/first/second-order}
    \begin{tabular}{|p{1cm}|p{6cm}|p{4cm}|p{4cm}|}
    \hline
     & Approach &  Pros  & Cons \\
    \hline
    Zero-order & prunes weights based on local importance score (e.g., magnitude) of each weight in the model  \cite{zafrir2021prune}   & computationally efficient & Sub-optimal for large networks due to ignoring global importance of weights \\
    \hline
    First-order & Considers impact of each parameter on the model accuracy, e.g., weights moving away from zero are considered important  \cite{sanh2020movement}  & More accurate in high-sparsity regimes due to considering the gradient information & computationally expensive due to requiring gradient computation for every weight. \\
    \hline
    Second-order & uses the approximations of the loss curve to guide pruning. Considering loss curvature helps establish relationship between weights and loss function.   \cite{kurtic2022optimal} & Most accurate due to computing the Hessian matrix (or its approximations) & Very expensive as it requires hessian computation\\
    \hline
    \end{tabular}%
  \label{tab:order}%
\end{table*}%

\subsubsection{Embedding-layer distillation} 
In addition to model-level, attention-level and hidden states, the knowledge from the teacher embedding layer can be transferred to the student's equivalent layer to learn the embedding layer. 
Manifold learning-based distillation \cite{chen2020learning} methods  use  inter-sample information to support layers with mismatched dimensions. Hao et al. \cite{jia2021efficient} utilize patch-level information and develop a fine-grained manifold distillation method to transfer the patch-level manifold information between teacher and student ViTs. The tiny model is trained in such a way that it mimics the patch-level manifold space of the teacher model using three manifold matching loss terms.

Although useful, KD methods suffer from problems such as limited generalization, interpretability and overfitting. The distilled student model may not generalize well to new and unseen data when the student tries to only mimic the teacher and ignores the full distribution of the target data. 
KD can lead to overfitting if the student model is overtrained and tries to fit the teacher model too well.

\section{Pruning} \label{sec:Pruning}
Neural network pruning is a method to reduce the size and computation complexity by removing redundant weights and activations. The pruning algorithms force the weights/nodes/neurons/heads to be zeros as much as possible during inference run-time.  In this section, we classify methods based on saliency, sparsity pattern and transformer granularity.

\subsection{Overview of pruning techniques}
The general methodology of most pruning methods is to first train a neural network to achieve the best accuracy possible. The second step in this process includes identifying and removing the least important parameters based on magnitude or contribution to the overall model performance. The third step is to finetune the pruned model to recover the accuracy. The second and third steps are iteratively performed until there is an accuracy loss.

\subsection{Pruning taxonomy based on saliency quantification}
The pruning techniques can be divided into zero, first and second-order based on how they quantify the saliency of network parameters. Table \ref{tab:order} describes the three methodologies, along with their pros and cons.  

An example of a first-order technique is AxFormer  \cite{nagarajan2022axformer}. For large transformers, iteratively performing pruning and fine-tuning leads to overfitting the training data for the downstream applications. They solve this problem using a hierarchical greedy scheme that needs no additional fine-tuning.  They first find the baseline loss of the transformer by fine-tuning it on a downstream application. Then, the loss (say K) is computed by removing an element.  If K is below the lowest loss encountered so far, the element is pruned.  
To avoid overfitting, they prune an element only if it reduces the loss of at least half of the samples in the validation set. This ensures effective generalization.

Their pruning technique works hierarchically: it first looks at self-attention and FFN blocks, and only, if required, it analyzes their building blocks, such as neurons and attention heads. This approach prunes bulky blocks quickly and speeds up subsequent iterations. To further narrow down the search space, if a block (say, self-attention) is found to be of high significance, they exclude all the heads in that block from further consideration. 
For effective pruning, it is important to analyze the elements in the right order. Towards this, they note that the lower layers of BERT extract phrase-level and surface features; intermediate layers find syntactic features, and deeper layers focus on semantic features. Deeper layers are required only for capturing long-range dependency. The depth of analysis required by each task is different, e.g., local context is sufficient in sentiment extraction since sentiments change quickly. In fact, syntactic and semantic knowledge is usually not necessary. As such, they inspect from the last layer towards the first layer since the last layers are unimportant or harmful for sentiment analysis.

In the transformer, the use of soft attention facilitates end-to-end training. However, by accounting for only the top-N (say N=30) attention values, the transformer can focus on the most important phrases of the input.  They replace hard attention with soft attention in the layers where hard attention reduces the validation loss. Hard attention can sometimes enable better representation by focusing on just one input token. Hence, hard attention is especially useful for capturing phrase-level information in lower layers. Their technique leads to smaller, faster and more accurate models. Their technique can also further improve Q8BERT and DistllBERT models. Also, their models are relatively insensitive to the choice of random seed initialization. Finally, their technique has small latency since it only requires multiple iterations on a small validation set and no fine-tuning or retraining.

An example of the second-order pruning technique is oBERT \cite{kurtic2022optimal}, which approximates the Hessian function to measure the importance of model parameters. The pruned models attain 8.4$\times$ inference speedup with less than 1\% accuracy drop and 10$\times$ speedup with less than 2\% accuracy drop on Intel Xeon Platinum 8380 CPU platform.

\begin{table*}[htbp]
  \centering
  \caption{Classification of Pruning Technqiues based on the Sparsity Pattern}
    \begin{tabular}{|c|p{12cm}|c|}
    \hline
      &  Approach  &  Example  \\
    \hline
    Unstructured  &  It prunes the weight matrices of a model irregularly, resulting in unstructured weight/activation matrices. An example of it is element-wise pruning.  &  \cite{cheong2019transformers, gordon2020compressing, sanh2020movement}  \\
    \hline
     Semi-structured  & An example of semi-structured sparsity is N:M sparsity, where the weight matrix is divided into groups, each of size M, of which N elements are pruned. &   \cite{mishra2021accelerating, fang2022algorithm, holmes2021nxmtransformer,fang2022efficient}\\
    \hline
    \multirow{5}[10]{*}{Structured} & It prunes at component-level, e.g., neurons, channels, heads, columns, rows or entire layers, instead of individual weight parameters. This leads to more regular network that gain performance even on general-purpose hardware.  &    \\
\cline{2-3}          & 1. Row/Column:  remove redundant rows/columns in weight matrices  &  \cite{xia2022structured, zhu2021vision}   \\
\cline{2-3}          & 2. Head-wise: it is a row-wise structured pruning technique that removes the redundant heads in MHA &  \cite{michel2019sixteen, voita2019analyzing}  \\
\cline{2-3}          & 3. Layer-wise:  It prunes individual layers of a network & \cite{fan2019reducing}   \\
\cline{2-3}          & 4. Block Pruning: it first groups a weight matrix into a 1D (Figure \ref{fig:Pruning_types}c) or 2D (Figure \ref{fig:Pruning_types}d) blocks and prunes the entire block &  \cite{lagunas2021block}  \\
\hline    \end{tabular}%
  \label{tab:structuredPruning}%
\end{table*}%

\begin{figure*}
    \centering
    \includegraphics[scale=0.55]{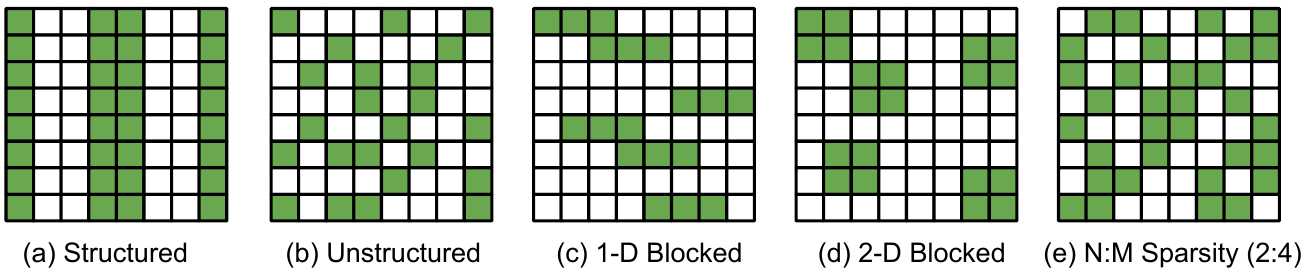}
    \captionsetup{justification=centering}
    \caption{Types of sparsity}
    \label{fig:Pruning_types}
\end{figure*}

\subsection{Classification based on the matrix sparsity pattern}
A neural network can be pruned at different levels, resulting in different sparsity patterns. The methods are classified into unstructured, semi-structured and structured methods. The techniques are described in Table \ref{tab:structuredPruning} and illustrated in Figure \ref{fig:Pruning_types}. We summarize a few transformer-specific works based on this classification below:

\subsubsection{Unstructured Pruning}

The irregular pruning methods result in a significant reduction of model parameters due to the lowest level of pruning granularity. However, unstructured sparse patterns require specialized hardware architectures and sparse libraries to take advantage of the significantly compressed model. Although 97\% of the network parameters can be pruned, it is hard to obtain substantial inference speedup on many hardware platforms \cite{sanh2020movement}. Gordon et al. \cite{gordon2020compressing} apply magnitude-based pruning \cite{han2015deep} to compress BERT, where weights close to zero are pruned. The authors observe that low pruning levels (30-40\%) do not affect pretraining loss, while medium pruning levels hinder useful pretraining information from being transferred to downstream applications. Additionally, the pretraining loss depends on the downstream application in the case of high pruning levels.

Gradual magnitude pruning (GMP) \cite{zhu2017prune} is a process of gradually pruning the weight parameters with low magnitude during the training process. Sparse*BERT \cite{campos2022sparse} applies GMP on LLMs and shows how pruned models can transfer between domains and applications. They show that models pruned on a particular large-scale dataset and applications on the general domain language can be utilized on new domains and small-scale datasets without requiring significant hyperparameter tuning. They can obtain similar accuracy as unpruned LLMs.

Prune-OFA \cite{zafrir2021prune} creates unstructured sparse pre-trained BERT models that can be fine-tuned on the target downstream applications at high sparsity ratios. This method consists of a teacher preparation step for initializing the student model and a student pruning step which is fine-tuned for the downstream task through a knowledge distillation approach. Prune-OFA also introduces a pattern lock to prevent the zeros in the model from being updated while fine-tuning the network.

PLATON \cite{zhang2022platon} is another example of unstructured pruning method. It captures the uncertainty of importance scores based on the absolute difference between the importance score at the current pruning iteration and the moving average of the previous iterations. This method retains weights with low important score values and high uncertainty. 
On a wide range of transformer models, such as BERT-base \cite{devlin2018bert} and ViT-B16 \cite{dosovitskiy2020image}, PLATON can compress the model by up to 90\% while increasing the accuracy by 1.2 percentage point.

\subsubsection{Semi-structured  Sparsity}
This type of sparsity pattern is more efficient than unstructured pruning and has been implemented in commercial hardware, e.g., the tensor core in Nvidia A100 GPU can accelerate the 2:4 sparsity pattern, illustrated in Figure \ref{fig:Pruning_types}(e), by a factor of 2 \cite{mishra2021accelerating}. N$\times$MTransformer \cite{holmes2021nxmtransformer} models N:M sparsity as a constrained optimization problem and optimizes the downstream tasks while considering the hardware constraints. The authors use the Alternating Direction Method of Multipliers (ADMM), a popular technique for non-convex optimization problems with multi-objective constraints. N$\times$MTransformer prunes Q, K, and V matrices, attention output and fully connected layers in the Transformer, and the sparsified model is 1.7 points more accurate than the SOTA N:M sparse language models.

Fang et al. \cite{fang2022algorithm} propose a network-hardware co-design framework to generate a series of N:M (3:4, 2:4, 1:4) sparse transformer models for deployment on a diverse set of FPGA platforms and a dedicated hardware architecture to support this specialized sparse implementation. The set of N:M sparse transformers are generated using inherited dynamic pruning (IDP), resulting in 6.7 percentage point increase in accuracy. %  

Chen et al. \cite{chen2021chasing} propose three sparse vision transformer exploration methods to obtain compressed models. The first method, Sparse Vision Transformer Exploration (SViTE), dynamically extracts sparse subnetworks and explores sparse connectivity during the training process. Structured Sparse Vision Transformer Exploration (S$^{2}$ViTE) structurally prunes and grows the attention heads as structured sparse models are more hardware-friendly. The Sparse Vision Transformer Co-Exploration (SViTE+) co-explores data and architecture sparsity and determines the most important patch embeddings. The end-to-end exploratory methods improve the accuracy of DeiT-small by 0.28\% while compressing at least 50\% weights.

\subsubsection{Structured pruning}

Structured pruning methods prune at the granularity of entire layers/filters/channels/heads, leading to a sparse matrix with structured pattern.  WDPruning \cite{yu2022width} is a structured pruning technique to reduce the width of FC and MHA layers and the depth of the overall network simultaneously. The width of the weight matrices is pruned using a set of learnable parameters, which are used to dynamically adjust the width of the matrices. On the other hand, the depth of the model is pruned by shallow classifiers based on the intermediate data of the self-attention blocks. The pruning results on DeiT-base \cite{touvron2021training} shows that the throughput can be improved of 15\% for an accuracy drop of 1\%.

\subsection{Classification based on Pruning granularity}

In this subsection, we focus on pruning the trained weights of a transformer model. Based on the algorithm and hardware requirement, a transformer can be pruned at different granularity levels, such as element-, layer-, head-, line-wise.

\subsubsection{Element-wise pruning}
The element-wise pruning method is analogous to zero-order, which picks the individual element in a transformer as the pruning granularity, resulting in an irregular sparse matrix. The importance of each weight can be measured based on different criteria such as magnitude, output activation values, or scores calculated by other functions. Transformer.zip \cite{cheong2019transformers} performs iterative magnitude pruning \cite{han2015deep}, which prunes all the parameters below a certain threshold in each pruning iteration.

\subsubsection{Row/Column Pruning}

Row/Column is a line-wise structured pruning technique to remove redundant rows/columns in the weight matrices of a transformer. The row pruning refers to removing individual attention heads, while column pruning removes output features. Both techniques prune less important parts of the self-attention unit while maintaining the regular structure of the model. CoFi \cite{xia2022structured} learns the pruning mask of all operators in a Transformer: FFN layers (Z$_{FFN}$), FFN intermediate dimensions (Z$_{int}$), MHA layers (Z$_{MHA}$), Attention heads (Z$_{head}$), Hidden dimensions (Z$_{hidn}$). This framework achieves 10x speedup and close to 95\% sparsity across several datasets while preserving 90\% of the accuracy of the transformer. VTP \cite{zhu2021vision} target output feature of the linear projections, i.e., FC layers, in a ViT model by learning the sparsity mask of each output feature based on L$_{1}$-norm.

TPrune \cite{mao2021tprune} is a combined row and column-wise transformer pruning technique for resource-constrained environments. This method divides the weight matrix into several sub-blocks with the same shape and then utilizes the row and column-wise lasso penalty. The row-wise and column-wise l2-regularizer terms are added to the loss function, and the model is trained to learn the structured sparse representations. The regularizer here is the square root of the sum of squares of weights along a dimension. This way, the least important rows or columns are automatically pruned, as gradient descent aims to minimize the combined loss function. The individual structurally pruned subblocks are concatenated to form the final weight matrix. The pruned transformers achieve 1.16–1.92$\times$ speedup for the same model accuracy on mobile devices.

UP-ViT \cite{yu2021unified} is a unified framework to structurally prune all the important dimensions of ViT blocks, such as MHA, FFN, normalization layers, and convolution channels in ViT variants. The importance of each channel is calculated by first dividing the ViT model into several individual uncorrelated components and evaluating the performance difference after removing each channel in every component. The authors apply the UP method on several SOTA ViT models, such as DeiT, PVT and achieve acceptable accuracy performance tradeoffs on compressed models.

\subsubsection{Block Pruning}

Lagunas et al. \cite{lagunas2021block} compute the importance of each block in the attention layers based on its contribution to the overall model performance and prune the least important blocks. HMC-Tran \cite{huang2021hmc} is a tensor-core aware pruning (TCP) to exploit sparsity in a coarse-grained manner using block pruning technique (Figure \ref{fig:Pruning_types}(d)). The authors first divide the weight matrix into p$\times$q blocks, say 16$\times$16, and prune the entire block whose l2-norm is less than a predefined value \textit{prec}. TCP attains a speedup of 3.68$\times$ with 92\% sparsity on BERT-base model on V100 Tensor core GPU, while the baseline SVD achieves only 3.56$\times$ speedup.

\subsubsection{Head Pruning}

Head pruning is a row-wise structured model compression method that removes the redundant heads in the multi-head self-attention module. Michel et al. \cite{michel2019sixteen} show that a few layers in a transformer can be reduced to as low as a single head. The authors use a first-order proxy method to determine the importance of each head and prune them iteratively. The experiments on Vanilla Transformer and BERT show that the models can be compressed up to 20–40\% without any quality loss. Voita et al. \cite{voita2019analyzing} first analyze the intrinsic properties and determine the importance of each head to draw a conclusion that specific heads take specific roles. The authors then develop a gating mechanism to prune half of all the heads with less than 0.25 BLEU loss.

\subsubsection{Layer-wise Pruning} 

Layer-wise pruning is a structured pruning technique that uses individual layers as the pruning granularity to reduce the depth of the overall transformer network. LayerDrop \cite{fan2019reducing} selects a sub-network from the original Transformer model by learning the retention rate for each layer during training, and only the layers with high impact are preserved during the inference runtime.

\subsubsection{FFN Pruning} Pruning redundant weights in an FFN layer is extremely important as this layer account for close to 2/3rd of the total parameters in a Transformer model (excluding the embedding parameters). Ganesh et al. \cite{ganesh2021compressing} showed that MHA and FFN layers take almost similar time on GPUs even though the former layer account for 1/3rd of the parameters, while FFNs become a bottleneck on CPUs. VTP \cite{zhu2021vision} prunes channels in such a way that it focuses more on the FFN unit than MHA weights.

\subsection{Quantitative comparison of pruning techniques}
\begin{figure}[ht]
    \centering
    \subfloat[{\small DeiT-Base}]{{\includegraphics[scale=0.3]{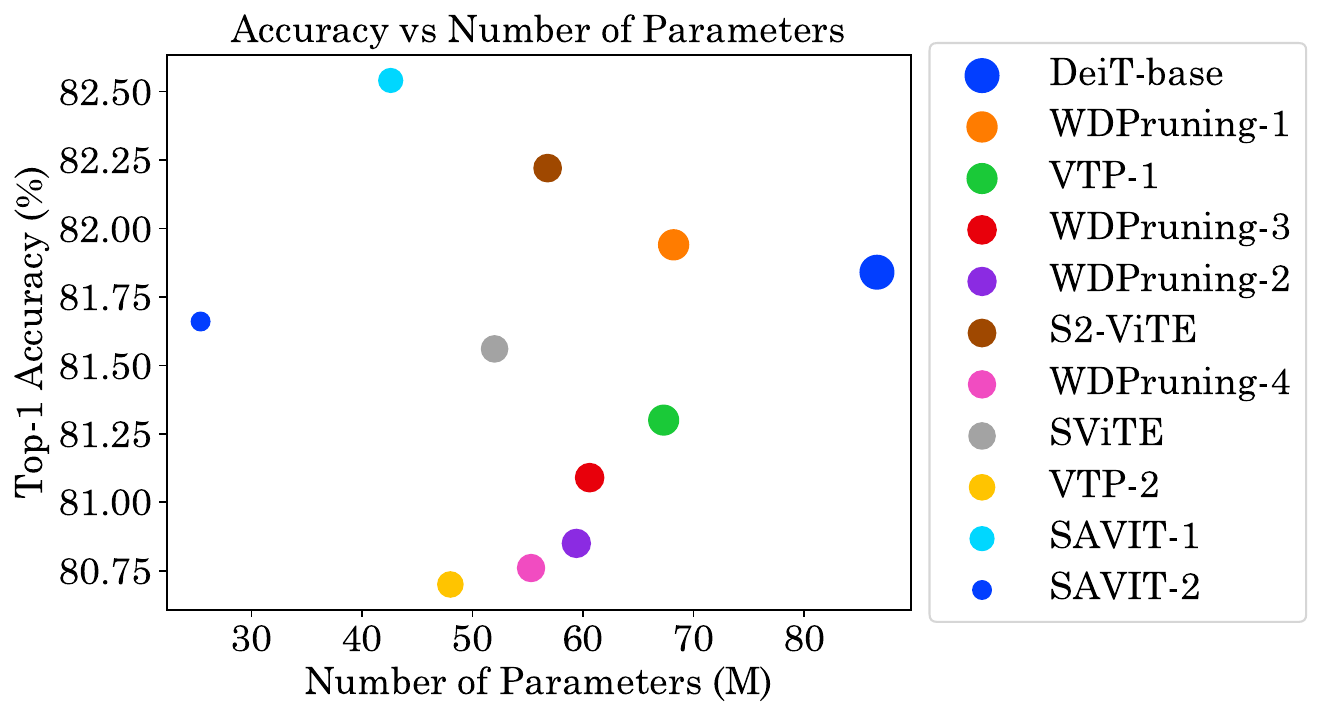}}}
    \hskip -3.5ex
    \qquad
    \subfloat[\small DeiT-Small]{{\includegraphics[scale=0.3]{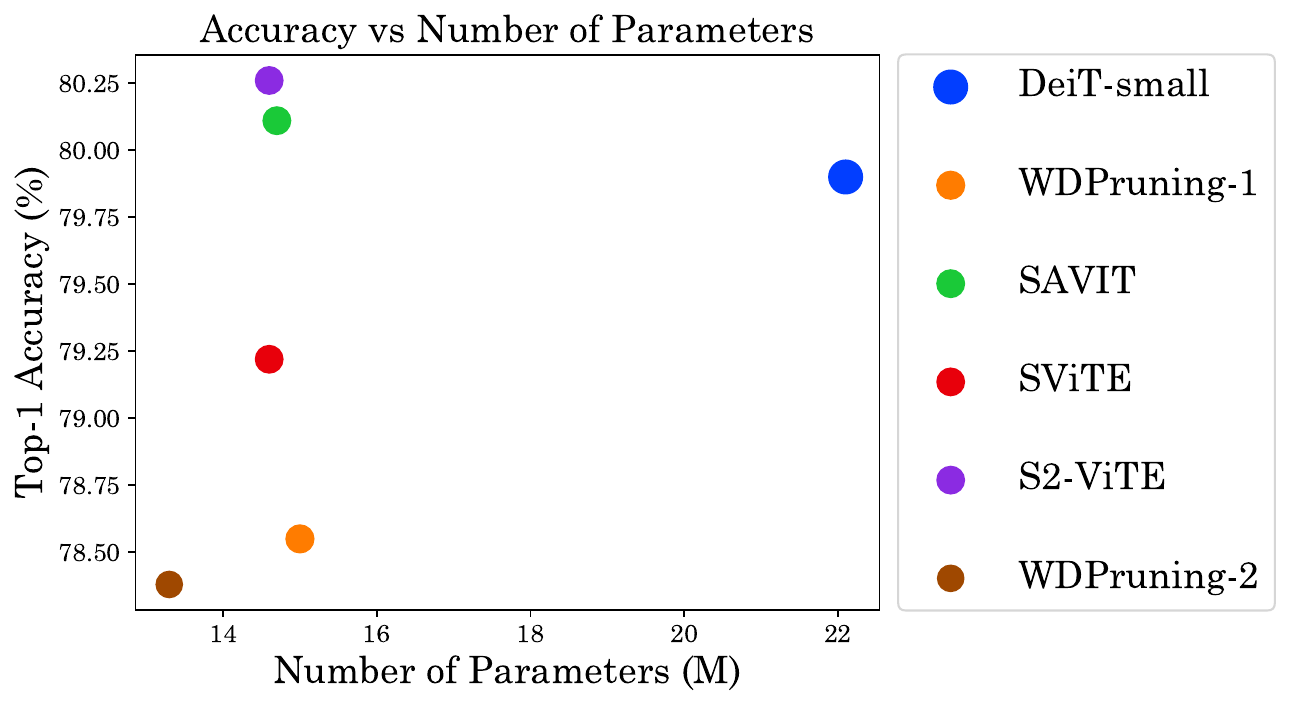}}}
    \hskip -3.5ex
    \qquad
    \subfloat[\small DeiT-Tiny]{{\includegraphics[scale=0.3]{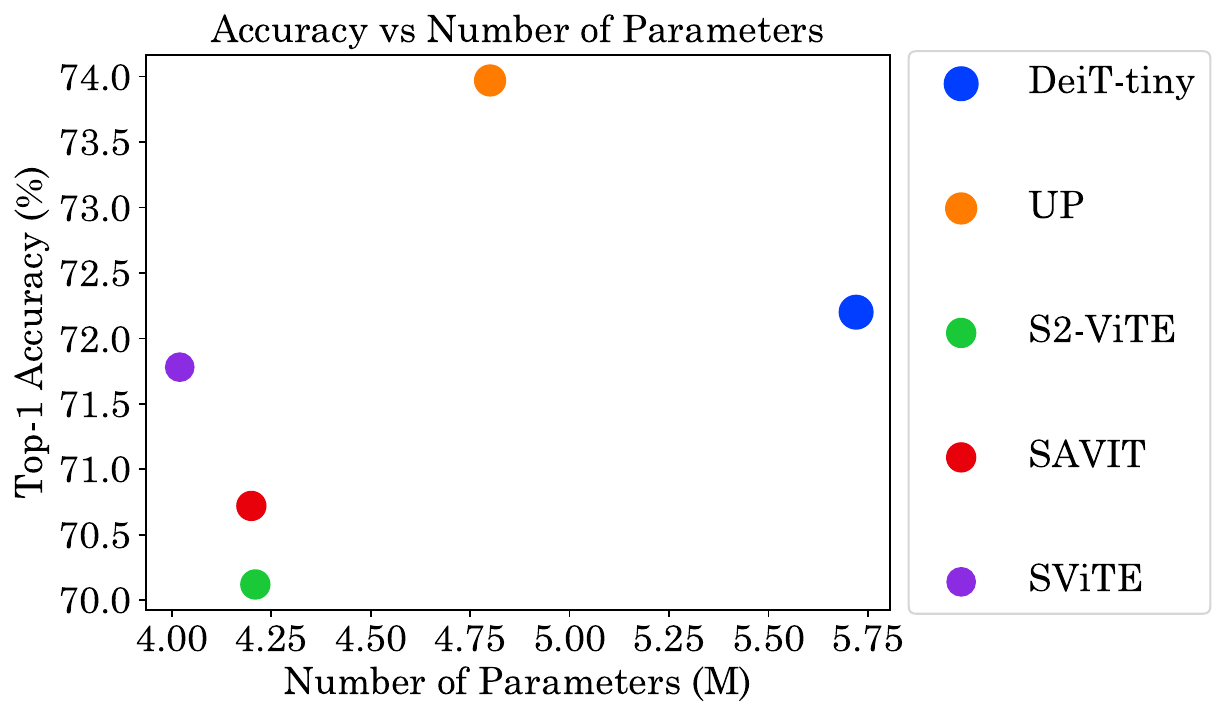}}}
    \caption{Comparison of pruning technqiues: SAVIT \cite{chuanyangsavit}, VTP \cite{zhu2021vision}, WDPruning \cite{yu2022width}, SViTE \cite{chen2021chasing}, UP \cite{yu2021unified} on (a) DeiT-base, (b) DeiT-small (c) and DeiT-tiny networks \cite{touvron2021training}}
    \qquad
    \label{fig:pruning_compare}
\end{figure}

Figures \ref{fig:pruning_compare}(a), \ref{fig:pruning_compare}(b), \ref{fig:pruning_compare}(c) compare the accuracy vs number of parameters of pruning methods on DeiT-base, DeiT-small and DeiT-tiny networks \cite{touvron2021training}, respectively. We obtain the accuracy numbers from the corresponding papers. The plots show that certain pruning techniques can reduce the number of parameters while improving the accuracy of the DeiT model, while a few methods remove parameters with some compromise in accuracy. For example, WDPruning \cite{yu2022width}, S$^{2}$ViTE \cite{chen2021chasing}, SAViT \cite{chuanyangsavit} methods increase the accuracy of the compressed model with less number of model parameters on DeiT-base model. On the other hand, other methods, such as VTP \cite{zhu2021vision}, comes with a drop in accuracy. Therefore, the accuracy of the pruned models depend on the pruning method and finetuning pipeline.

\subsection{Token/Patch Pruning}

In the previous subsection, we discussed pruning methods pertaining to the weights of a transformer model, while this subsection focuses on token/patch pruning, which is analogous to activation pruning in the CNN model. Token pruning \cite{kim2021learned} reduces the computation complexity of a transformer by removing redundant tokens or words from the input vocabulary, whereas patch pruning removes less important patches in the embedding of a ViT model. This pruning can be applied at different stages of a transformer model, such as at embedding, intermediate and final classification layers.

Learned Token Pruning (LTP) \cite{tang2022patch} adaptively prunes less effective tokens as the sequence passes through different layers of the network. The tokens below a certain threshold, learned during the training process, are pruned in every layer, allowing the length of the pruned sequence to vary with respect to the input sequence. Luo et al. \cite{luo2022attention} proposed a token pruning method for vision transformers by using the attention score as a natural indicator to determine the importance and prune tokens. An attention-based pruning module is inserted between the self-attention layers. The integrated weight parameters are fused with MHA to estimate the importance of each token and prune the tokens in the layer accordingly.

Yang et al. \cite{yang2022dtatrans} dynamically monitors the tolerance of tokens and adapts the precision of Q/K/V vectors. They note that the noise tolerance of a token depends on its importance. They sort the tokens based on their importance scores. There is negligible impact on BERT accuracy when adding Gaussian noise (equivalent to quantization) in the tokens with very low importance. However, doing this to highly important tokens has a high impact on accuracy. They further note that pruning 19\% least important tokens causes only a 0.2\% drop in accuracy, whereas pruning 24\% tokens degrade accuracy by 5\%. Hence, the pruning ratio needs to be carefully controlled, and accuracy loss due to pruning needs to be compensated.

Their technique divides the tokens into three types: high-precision (e.g., top 15\% most important), low-precision (next 70\%) and pruned (last 15\%). They are stored with 8b, 4b and 0b, respectively. They regard pruning as 0-bit quantization, which unifies both these techniques. The exact ratio of tokens of each type is decided based on Bayesian optimization. The pruned tokens are consolidated in a single representative token (RToken), obtained by weighting the tokens based on their importance score and then summing up these values. This RToken (which can be 4b or 8b) is concatenated with 8b and 4b tokens and fed to the transformer. At the end of the transformer block, these pruned tokens are updated and concatenated with the output non-zero-bit tokens. This output is fed to the next transformer block. Processing this RToken adds only minor overhead but avoids the accuracy loss due to completely pruning unimportant tokens. This approach allows more aggressive pruning for the same accuracy loss.

Their hardware accelerator uses a variable-speed systolic array \cite{song2020drq} to support 4b and 8b matrix multiplication. Due to the dataflow constraints of SA, the PEs (processing elements) performing 4b*4b and 8b*4b operations have to stall for one cycle and two cycles, respectively. Due to this, the PEs remain under-utilized. To deal with this issue, they group similar-precision tokens together and place low-precision tokens in the front. This reduces stall cycles since the cases of $Q$ and $K^T$ having different precisions is reduced.

\subsubsection{Uniform vs Non-uniform Token Pruning} The uniform token pruning methods use a single pruning configuration for all the tokens throughout the network for a given dataset. Nevertheless, the input sequence can vary with respect to different tasks and datasets. Therefore, applying the same pruning percentage can potentially under-prune short sequence or over-prune long sequence \cite{kim2021learned}. The non-uniform token pruning techniques adapt the pruning percentage based on the characteristics of the input sequence. SpAtten \cite{wang2021spatten} is an example of a non-uniform token pruning method that assigns the pruning rate proportional to the input sequence length.

\subsubsection{Static vs. Image-Adaptive Patch Pruning}
Static token pruning methods \cite{chen2021chasing, tang2022patch} prune the number of input tokens by a fixed ratio for different images. They neglect the fact that each image's information varies in region size and location. The image-adaptive token pruning methods \cite{pan2021ia} remove the surplus tokens based on the image characteristics to attain a per-image adaptive pruning rate. Therefore, the latter methods can achieve a higher overall model compression ratio than the former method. AS-ViT \cite{liu2022adaptive} is an adaptive sparse token pruning method which uses a set of learnable thresholds and MHA to prune the redundant tokens. The attention weights of the self-attention unit evaluate the token significance with a few additional operations, and the learnable parameters are inserted within the ViT model, distinguishing important tokens from uninformative ones. The learnable threshold parameters are optimized in such a way that they can balance accuracy and model complexity, thereby generating different sparse combinations for different input sequences.

HeatViT \cite{dong2023heatvit} is an image-adaptive token pruning method for efficient and accurate ViT inference.  
The authors designed a token selector consisting of an attention-based multi-head token classifier and a token packager to classify tokens accurately and consolidate non-informative tokens. The pruning rate is enhanced by carefully analyzing the inherent computational patterns in ViTs, as opposed to static pruning. The hardware efficiency is further improved by employing 8-bit fixed-point quantization.

\subsubsection{Quantitative comparison of token pruning techniques}
\begin{figure}[h]
    \centering
    \includegraphics[scale=0.43]{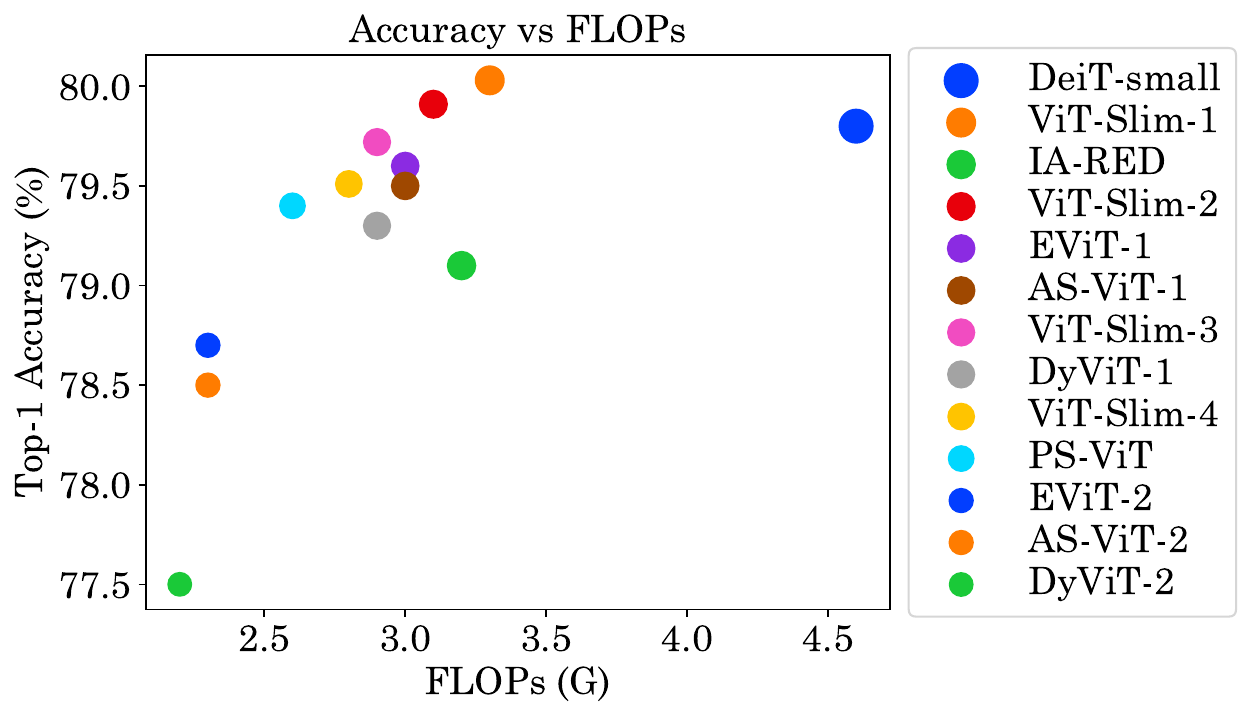}
    \captionsetup{justification=centering}
    \caption{Comparison of token pruning methods: PS-ViT \cite{tang2022patch}, DyViT \cite{rao2021dynamicvit}, EViT \cite{xu2022evo}, AS-ViT \cite{liu2022adaptive}, ViT-Slim \cite{chavan2022vision}, IA-RED \cite{pan2021ia}. The baseline is DeiT-small \cite{touvron2021training} model.}
    \label{fig:token_pruning_compare}
\end{figure}

In Figure \ref{fig:token_pruning_compare}, we compare the accuracy and floating point operations (FLOPs) of various token pruned models. As pruning tokens does not alter the network/weight structure, the number of parameters remains the same as the baseline DeiT-small model. However, the number of multiplications and additions are reduced due to token/activation pruning. The size of each circle in the figure corresponds to the relative FLOPs of the token pruned model with respect to the baseline DeiT-small model. All the pruning methods reduce the total number of FLOPs. A few techniques like ViT-Slim \cite{chavan2022vision} improves the baseline DeiT-small accuracy and reduces the FLOP count. DyViT \cite{rao2021dynamicvit} method achieves best compression with respect to FLOPs but comes with a drop in accuracy. 

\subsection{Post Training Pruning (PTP)}

The conventional pruning algorithms require fine-tuning the pruned network and/or jointly learning the pruning configurations. This is required to retain the accuracy lost due to trimming weights, thereby requiring a significant amount of retraining time. Post-Training Pruning (PTP) does not require any additional retraining and maintains the same baseline accuracy. 
For instance, Kwon et al. \cite{kwon2022fast} propose a PTP method for transformers based on Fisher information matrix as a criteria to identify redundant heads/filters. The framework requires only 3 minutes on a single GPU to remove heads in MHA and filters in FFN layers. It achieves a 1.56x speedup on inference latency for BERT with less than 1\% accuracy loss. The PTP methods can be divided into static and dynamic, which are explained below.

\subsubsection{Static PTP}
The static PTP methods identify and prune the least important parameters in the model irrespective of the input token sequence. The model is pruned only once and and is used for inference for all input sequence lengths. Frantar et al. \cite{frantar2023massive} propose a static PTP  method to compress giant language models, such as GPT, and show that it is possible to remove 50\% of the weight parameters without significantly compromising model perplexity. They develop SparseGPT, a one-shot and layer-wise solver based on closed form equations by approximating sparse regression solver and is efficient enough to produce a sparse model only in a few hours of GPU time. The proposed method achieve 60\% unstructured sparsity on OPT175B \cite{zhang2022opt} and BLOOM-176B \cite{scao2022bloom} models  with minimal accuracy loss under 4.5 hours. SparseGPT can be further extended to N:M sparsity (2:4 and 4:8) with some additional accuracy loss compared to the unoptimized model.

\subsubsection{Dynamic PTP}
Adaptive inference or dynamic PTP refers to the ability of a pretrained transformer to dynamically reduce and adjust the layer length during inference based on the input sequence/token, without requiring any additional finetuning. The intuition behind this technique is that each input sample is different in terms of complexity and using a fixed-size model for all input samples may not be computationally optimal. Hence, adaptive inference methods adaptively skip part of the layer computations according to the input sample to obtain the best performance. EBERT \cite{liu2021ebert} is an example of such method which dynamically prunes the redundant heads in the MHA unit and structured computations (output channels) in the FFN unit for each input sample at inference time. The authors employ two predictor networks (two feed-forward, one batch norm and ReLU layer), one for MHA and one for FFN unit.   
The predictor network generates a \{1, 0\} mask, equal to size of number of heads in MHA and number of ouput channels in FFN layer. The BERT model and the randomly initialized predictor network are jointly trained. The goal is to learn the predictor network to determine the most important components in the BERT model.

\subsection{Hardware-aware pruning}\label{sec:hardwareAwarePruning}
To realize the full performance benefit of pruning, there is a need to customize pruning to different hardware platforms.

Fan et al. \cite{fan2022adaptable} execute the BERT-large model on CPU and GPU platforms and show the percentage latency of attention layers, linear layers and remaining computations. For an input sequence length ($L$) of 256, linear layers dominate the execution time, whereas, for $L$=1024 and 2048, the attention layers become dominant. Hence, both these layers need to be accelerated. They classify the sparsity patterns into five basic patterns: random, low-rank, block-wise, sliding window and butterfly (BF). Of these, butterfly pattern is the only one which (1) allows structured data accesses (2) simultaneously benefits both global and local context (3) benefits both attention and FFN layers. Butterfly matrices are universal representations of structured matrices having a simple recursive pattern. They can approximate linear transformations. The low-rank sparsity requires sequentially reading the rows and columns, which leads to poor hardware efficiency. The sliding-window pattern only studies local context, and hence, it needs low-rank sparsity to compensate for the accuracy loss. Block-wise sparsity engines require additional algorithmic transformations. 
 
\begin{figure} [htbp]\centering
    \includegraphics[scale=0.43]{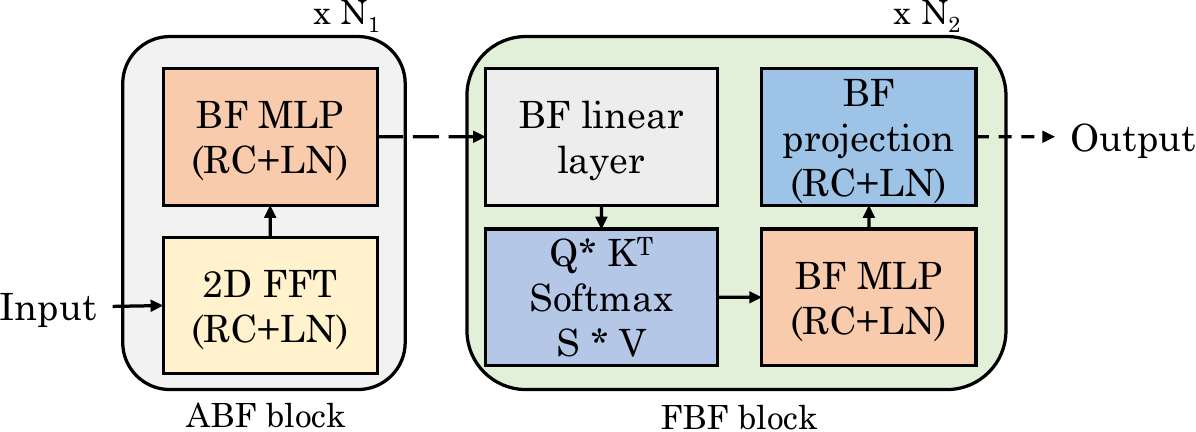}       
    \caption{Accelerator proposed by Fan et al. \cite{fan2022adaptable} (RC= residual connection, LN=layer-normalization, BF = butterfly) }
    \label{fig:fabnet}
\end{figure}

They propose two types of building blocks: ABF-block and FBF-block. The ABF-block has attention as the backbone and compresses all the linear layers using butterfly factorization. It has three BF linear layers (BFLLs) for producing Q, K and V matrices. Then, there is an MHA layer and another BFLL for extracting relationships between tokens. Finally, there is a BF FFN consisting of two BFLLs. The FBF-block replaces attention with a 2D FFT (fast Fourier transform) layer and hence, lower parameter and computation count. The mixing of input tokens by FFT enables subsequent BF FFN to compute longer sequences. By virtue of FFT, this block uses a unified BF pattern, leading to better hardware efficiency. However, the use of FFT degrades accuracy. Their proposed design uses $N_1$ FBF blocks and $N_2$ ABF blocks to address these tradeoffs, as shown in Figure \ref{fig:fabnet}.

Their accelerator has multiple BF engines, which can be programmed at runtime to accelerate either FFT or BF linear layers. It uses multiplexers and demultiplexers to choose the correct input and provide the correct output. This allows the reuse of add/subtract/multiply units. The butterfly pattern requires different inputs at different stages. Hence, both row-major and column-major patterns lead to bank conflict. They propose a custom data layout that shifts down the first element of every column by a certain number of rows. This layout removes bank conflicts in reading data in the first two stages of the BF pattern. They use double-buffering to overlap memory access with computation.
Since FFT computation involves complex data, they concatenate the lower (higher) portions of two input buffers to create first (second) ping-pong storage.  Their algorithmic optimizations reduce the model size and FLOPs with no loss in accuracy.

Zhang et al. \cite{zhang2021algorithm} note that  even with structured pruning, the shape of pruned weight matrix may differ in different encoders of various heads in an MHA.  They propose compressing the transformer model in a weight-shape-aware manner so that the weight matrices of Q, K, V, O, FFN1 and FFN2 layers are of similar shapes. This improves the utilization of FPGA buffers and MAC (multiply-accumulate) array. 
Their compression technique first finds the weight importance based on the \enquote{winning ticket hypothesis} methodology. It shows that the sub-model, created by removing the lowest-magnitude weights and training from the original initialization, can achieve similar accuracy as the original unoptimized model. They use LayerNorm to find the importance of every weight column. One by one, in every encoder and decoder, they attach LayerNorm to any MatMul (matrix multiplication) containing weight. Then, they train the transformer till the convergence of $\gamma$ factors in the newly attached LayerNorms. The final value of $\gamma$ shows the importance of each weight column. The scaling factors $\gamma$ of all the LayerNorms are stored in a new model.
In LayerNorm, the scaling elements are the same for different rows but different for every row element. Thus, using LayerNorm with MatMul, $\gamma$ and $\beta$ show the scaling factor for each column. Since $\gamma$ is more dominant than $\beta$, $\gamma$ is taken as the column-importance factor. Then, weight columns with $\gamma$ values below a threshold are pruned.

Then, a two-stage pruning strategy is used. (1) Coarse-grain pruning removes weights with the same ratio throughout the transformer model. Pruned weight matrices are of similar shape. (2) Fine-grain pruning, which removes redundant weights without accuracy loss. In both stages, pruning and training are performed alternately. While performing MatMul, they compute the output in a column-wise manner. A single-size element-wise vector multiplication and addition substitute the MAC of various sizes. Modern FPGAs have high bitwidth (e.g., 27b * 18b) DSPs, which provide no extra advantage for INT8 computations. They use double-MAC technique $(((a <<) + b)*c)$ which accomplishes two multiplications in a single operation. In their dataflow, the weight term is used as the shared term $c$. Column-wise computation naturally avoids multiplications with zero operands. Their technique compresses the transformer by 95 times and achieves high throughput on FPGA.

\begin{figure} [htbp]\centering
    \includegraphics[scale=0.43]{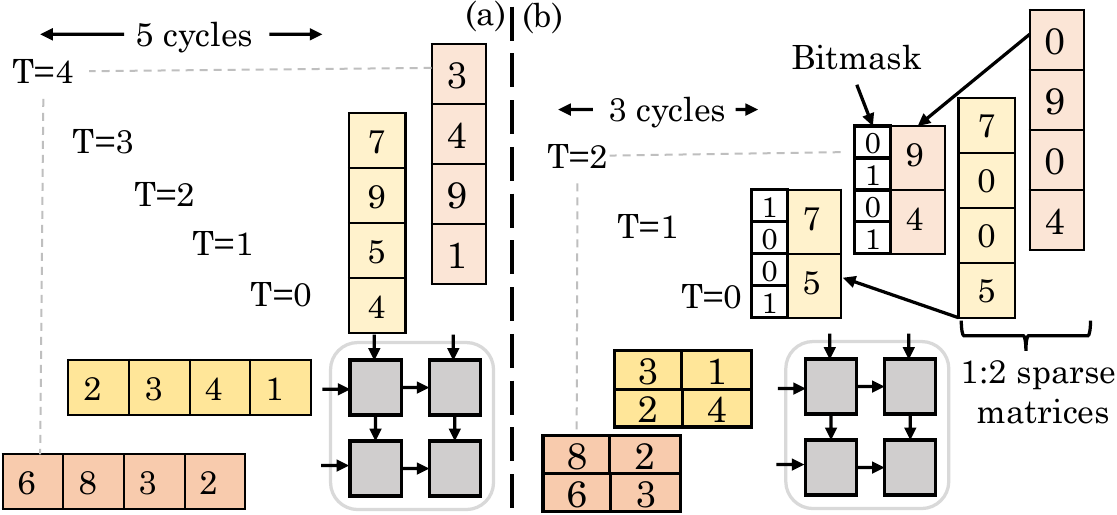}       
    \caption{Multiplication of (a) dense (b) sparse matrices using systolic array \cite{fang2022efficient}  }
    \label{fig:SystolicDenseSparse}
\end{figure}

Fang et al. \cite{fang2022efficient} present a technique to deal with networks having N:M sparsity. A transformer having N:M sparsity requires both sparse-dense and dense-dense MatMul. They present a unified MatMul engine for accelerating both these MatMul. It has $H$ planes of $K*K$ SA. Each PE has a MAC, MUXes, registers and non-zero (NZ) value selection unit. The MAC unit multiplies two 16b numbers and accumulates with a 32b partial sum. NZ value selection unit is activated only for sparse-dense MatMul. Based on the input bitmask, it loads only NZ weights and the corresponding activations. Figure \ref{fig:SystolicDenseSparse} shows dense-dense and sparse-dense MatMul for a 1:2 sparsity pattern. Clearly, for sparse-dense MatMul, trivial computations are avoided, improving hardware efficiency and performance.

They note that for a fixed sparsity ratio (i.e., fixed (M-N)/N), setting N to 1 provides comparable accuracy as N=2, and hence, they set N=1. The accuracy loss remains negligible even with 75\% sparsity, and therefore, they set N:M value to 1:4. They set $H=4$ and $K=16$ and thus, their accelerator has 1024 MAC units. Compared to Lu et al. \cite{lu2020hardware}, which uses 4096 MAC units, their design has comparable inference latency and much higher throughput per MAC unit. On using N:M of 1:8, the latency reduces further.

Li et al. \cite{li2022slice} note that transformers that use composite sparse attention, such as Longformer, are inefficient on GPUs since different sparse patterns have different locality patterns. Previous works store the sparse matrices using either coarse-grain format (e.g., block coordinate format) or fine-grain format  e.g., coordinate (COO) or compressed sparse row (CSR), is inefficient for all sparse patterns. Hence, composite sparse attention takes three-fourths of the execution time on Longformer. They propose novel GPU kernels for accelerating such networks. Based on the spatial locality, they divide sparse patterns into two categories. (1) coarse-grain: all types of blocked patterns, local and dilated. (2) fine-grain: random, global and selected. For these categories, they use BSR (block sparse row) and CSR formats, respectively.

While feeding the input, they create metadata for these formats and load them to the GPU. (1) BSR metadata is produced based on the window size and remains fixed for the dataset. (2) CSR metadata is produced based on global indexes of crucial token locations. This metadata needs to be revised for each iteration. Then, coarse- and fine-grain kernels are used to execute corresponding patterns. SDDMM (sampled dense-dense matrix multiplication) and SpMM (sparse-dense matrix multiplication) are processed parallelly in two streams using coarse- and fine-grain kernels. For fine-grain kernels, they adapt Sputnik kernels and use CUTLASS kernels, which are more efficient than Sputnik kernels. They fuse scale and mask operations with sparse softmax. They propose two coarse-grain kernels that use BSR for (a) SDDMM and (b) SpMM.

(a) It assigns every row block in the output BSR matrix to a single thread-block. One thread-block handles the entire blocked GEMM. For C = I1*I2, every thread-block computes the NZ blocks in a row by reading the blocks from I1 and I2. They decompose blocked GEMM into hierarchical tiled GEMMs at thread-block, warp and thread levels.
(b) This kernel uses a blocked 1D tiling method. It is similar to (a), except that the output row block is not fully processed by one thread-block. Rather separate thread-blocks are used for 1D tiles of the output matrix. Similar to (a), further levels of tiling are also used. 
  
They propose a sparse softmax kernel for computing the outputs of both coarse and fine-grain kernels. Since softmax reads all the row elements, presence of overlapped coarse/fine-grain patterns in the same row leads to inaccuracies. Hence, they invalidate  the overlapping regions. Then, a row-level softmax is done following scheme (a) above.    
The output row block has NZ elements from both coarse- and fine-grain patterns. They sequentially scan the row using BSR metadata and CSR metadata to process NZ elements in coarse- and fine-grain patterns, respectively. Then, the results of these scans are combined to get overall results.  Their technique improves the performance of Longformer and QDS-Transformer models on state-of-art GPUs.  They achieve higher performance than techniques using either coarse- and fine-grain methods.

\subsection{Storage formats for sparse matrices}\label{sec:storageFormats}
Some researchers have proposed specific formats for storing different sparse matrices.

\begin{figure} [htbp]\centering
    \includegraphics[scale=0.43]{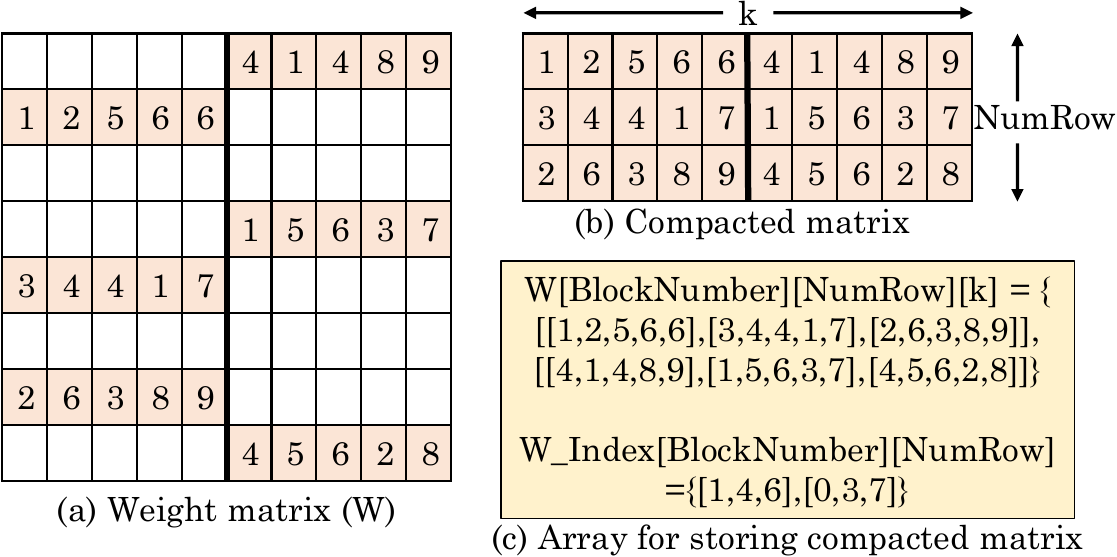}       
    \caption{(a) Sparse matrix resulting from BBP (b) compacted matrix (c) CBR format \cite{qi2021accommodating}    }
    \label{fig:SparseMatrixQi}
\end{figure}

Qi et al. \cite{qi2021accommodating} propose techniques for optimizing the transformer model. They compare \enquote{block-balanced pruning} (BBP) \cite{li2020efficient} with \enquote{block-wise pruning} (BW) \cite{narang2017block}. BBP prunes at row/column granularity in every block of a matrix (refer Figure \ref{fig:SparseMatrixQi}(a)), whereas BW prunes at the granularity of a block. On applying these to the transformer, BBP provides superior accuracy than BW at nearly all sparsity ratios. BBP is a fine-grained pruning scheme which retains more crucial information. Hence, they choose it as their compression scheme. They propose a \enquote{compressed block row} (CBR) format for storing matrices resulting from BBP. Based on the compacted matrix, CBR uses two arrays, as shown in Figure \ref{fig:SparseMatrixQi}(b)-(c). (1) A 3D array to save non-zero elements. (2) block indices of non-zero sub-rows.   Compared to COO and CSR formats, their CBR format needs lower memory. This is because it stores only the non-zero row-index and not the column index of each non-zero element.  Their pruning technique reduces latency on an FPGA.
 
\begin{figure} [htbp]\centering
    \includegraphics[scale=0.43]{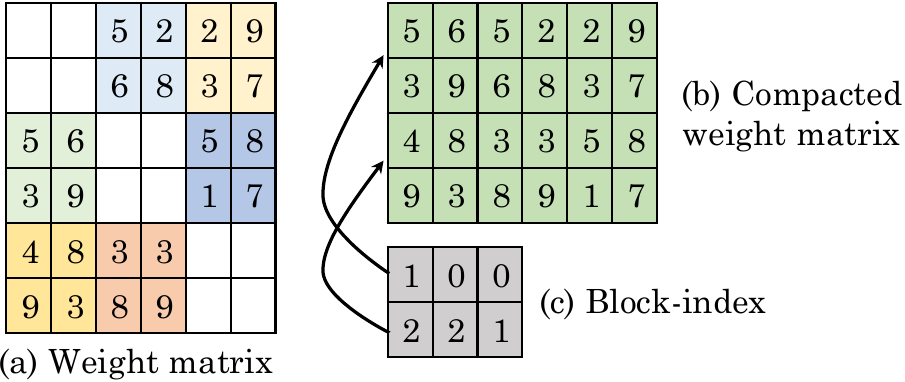}       
    \caption{(a) Resultant matrix from column-balanced block-pruning (b) compacted matrix and (c) block-index \cite{peng2021accelerating}}
    \label{fig:ColumnBalanced}
\end{figure}

Peng et al. \cite{peng2021accelerating} present a ``column balanced block-wise pruning'' (CBBWP) scheme for bringing the best of block-wise and bank-balanced pruning. It prunes blocks with low L2 norm in each column so that every column is left with the same number of blocks. Figure \ref{fig:ColumnBalanced} shows their proposed storage format. It needs only one index pointer for every block, saving memory. CBBWP enables parallelism within and across the blocks. Their FPGA accelerator optimizes MatMul with the CBBWP scheme.

\section{Quantization} \label{sec:Quantization}

Quantization refers to the process of reducing the precision of the model parameters (weights and activations). This method reduces the precision of these numbers to lower bit widths, such as 16-bit or 8-bit integers. This section provides a brief overview of transformer quantization methods, taxonomy, and implications on hardware performance. 

\subsection{Overview of quantization}
There are multiple ways to quantize a network while preserving the model performance. There general quantization methods can be classified into: static vs dynamic, uniform vs mixed precision, Post Training Quantization (PTQ) vs Quantization-aware Training (QAT). Table \ref{tab:quantization_overview} provides high-level definitions of these different methods and serves as a background for rest of the transformer-specific methods.

Table \ref{tab:PTQ_QAT} provides classification of quantization methods based on the methodology used (PTQ vs QAT) and bitwidth assignment within the network (unform vs mixed precision).

\begin{table}[htbp]
  \centering
  \caption{Classification of Quantization methods based on PTQ vs QAT and Uniform vs Mixed}
    \begin{tabular}{|c|p{1.5cm}|p{1.5cm}|}
    \hline
     Method &  PTQ &  QAT \\
    \hline
    Uniform &  \cite{yuan2022ptq4vit, zadeh2020gobo} &  \cite{zafrir2019q8bert, wang2022deep, mishra2021accelerating}  \\
    \hline
     Mixed Precision & \cite{liu2021post, li2022psaq, frumkin2022cpt, dettmers2022llm, zadeh2022mokey} & \cite{li2022accelerating} \\
    \hline
\end{tabular}%
  \label{tab:PTQ_QAT}%
\end{table}%

\begin{table*}[htbp]\centering
  \caption{Comparison between different categories of quantization techniques }
    \begin{tabular}{|p{8cm}|p{9cm}|}
    \hline      
    \textbf{Static:} The statistics of the pre-trained model, such as value ranges in a layer, are collected during offline phase over a calibration dataset. These values remain constant during the inference.  &  
	\textbf{Dynamic:} It dynamically calculates the quantization parameters of activation tensors during model execution. Hence, this process is expensive as quantization is performed for every new input sample     \\
    \hline
    \textbf{Uniform:} It quantizes all the layers with the same bitwidth \cite{rockint8}.  It is simple to implement but the performance is sub-optimal. 
    & 
    \textbf{Mixed-precision:} It assigns different bitwidth to different tensors or layers \cite{wang2019haq} within a network. It attains better accuracy but finding the optimal bitwidth for each layer/tensor is a combinatorial process . For example, Li et al. \cite{li2022accelerating} quantize attention weights to 8 bits and FFN weights to 16 bits. \\
    \hline
    \textbf{PTQ:} It quantizes a pretrained model without requiring additional fine-tuning steps. It can either use a calibration dataset or can be performed without any such dataset \cite{nagel2019data} in a data-free manner.  It avoids need of fine-tuning, but degrades accuracy for low precisions. & 
    \textbf{QAT:} It finetunes the quantized model by simulating the effect of quantization function by using Straight-Through Estimator (STE) \cite{bengio2013estimating} to approximate gradients through the non-differentiable quantization function in backpropagation. This makes the model robust to quantization noise but requires additional expensive training pipeline \\
    \hline
    \end{tabular}%
  \label{tab:quantization_overview}%
\end{table*}%

\subsection{General quantization procedures}
\subsubsection{Quantization function}

The most basic quantization function is linear quantization \cite{krishnamoorthi2018quantizing}, which represents the floating-point range using a fixed number of discrete levels. The range is divided into equally spaced intervals evenly distributed around a central value. For instance, the range is uniformly distributed between -128 and 127, with zero at the center for 8-bit signed quantization. This method is useful for data that have a symmetrical distribution, such as weight tensors in neural networks \cite{xiao2022smoothquant}. The quantization process is represented in Equation \ref{Eq:symmetric_quant_1}.

\begin{equation}\label{Eq:symmetric_quant_1}
    Q(r) =  \lceil \frac{r}{S} \rfloor + Z; S = \frac{r_{max} - r_{min}}{2^{b-1}-1}
\end{equation}

where $\mathbf{r}$ is the floating-point value and Q(r) is the corresponding quantized representation of r, $\mathbf{S}$ and $\mathbf{Z}$ represent the scale parameter and zero-point respectively. r$_{max}$ and r$_{min}$ denote the maximum and minimum values of the floating point range. The rounding function is given as $\lceil\cdot\rfloor$, and $b$ constitutes the quantization bit width. Besides linear quantization, other weight/activation quantization functions include Dorefa \cite{zhou2016dorefa}, PACT \cite{choi2018pact}, QIT \cite{jung2019learning}, LSQ \cite{esser2019learned}, etc. Q8BERT \cite{zafrir2019q8bert} is a QAT method which uses the linear quantization scheme \cite{krishnamoorthi2018quantizing} in the forward pass and Straight-Through Estimator (STE) in the backward pass.

Li et al. \cite{li2022accelerating} apply \enquote{symmetric linear quantization} on the ELBERT model. Then, they compute clipping range $W_{clip}$ as $sign(W)* min(|W|, 2^i - 2^{-d})$. Here $i$ and $d$ are the number of integer and fraction bits, respectively, in $W_{clip}$.  After this, the quantized weight is computed as $\lfloor 2^d W_{clip} +0.5\rfloor \times 2^{-d}$. They perform 8-bit quantization for weights and activations of MHA and 16-bit quantization for those of FFN. This allows storing all the weights on-chip. This quantization strategy incurs less than 1\% accuracy loss. They use LUTs (look-up tables) to compute log and multiplications in the entropy calculations of the early-exit strategy. The rest of the design and optimizations are similar to that of Lu et al. \cite{lu2020hardware}.
Implementation of their technique on FPGA incurs lower latency and energy consumption over a GPU implementation.

\subsubsection{Matching Full Precision Model}

In this subsection, we discuss qauntization methods which quantize the model in such a way that the low-precision model tries to match the full-precision model. This is similar to the knowledge distillation method, where the quantized model learns the patterns from floating point network so that the overall loss between both the versions are minimized.

Liu et al. \cite{liu2021post} formulate PTQ as an optimization problem to find optimal quantization intervals using Pearson correlation coefficient and ranking loss. The quantization scheme for MHA and FFN modules is learned differently within the transformer. The precision of the MHA module is learned by an attention map ranking loss, and the precision of the FNN unit is learned using the cosine similarity. The main aim of these learned methods is to maximize the similarity between the outputs of the full-precision and quantized network.

PSAQ-ViT \cite{li2022patch} is a data-free quantization framework for CV transformers, which does not require the calibration dataset to reduce the precision of a trained transformer model. The authors utilize the properties of the self-attention module and analyze the general difference in its processing of Gaussian noise and real images to generate realistic samples to estimate the quantization parameters. This framework uses a relative value metric to optimize the Gaussian noise to approximate the real images and then calibrate the quantization parameters. The experiments on benchmark models demonstrate the effectiveness of PSAQ-ViT, outperforming real-data-driven methods.

PSAQ-ViT-V2 \cite{li2022psaq} also utilizes a student-teacher learning methodology, where the goal is to minimize the KL divergence between the pretrained full precision model and the quantized network. CPT-V \cite{frumkin2022cpt} is based on contrastive loss, which learns the data representation that is not variant to changes in certain attributes.
The contrastive loss also minimizes the distance between the quantized and full precision predictions in a self-supervised approach for a given mini-batch. It requires only 1000 calibration images to determine the best quantization scales for each layer in the ViT model.

Yuan et al. \cite{yuan2022ptq4vit} observe that the activations after softmax and GELU operations differ from the traditional Gaussian distribution. Also, the commonly used metrics, such as MSE and cosine distance, are not efficient  in finding optimal quantization scale parameters. Therefore, the authors propose PTQ4ViT by employing twin-uniform to minimize the quantization error and Hessian-guided method to learn the scales of FC layers. The scaling factors are determined in such a way that the distance between the output before and after quantization is minimized. The experiments on ViT, DeiT, and Swin show that PTQ4ViT obtains near-lossless accuracy on the ImageNet dataset.

\subsubsection{Dealing with outliers}

Dettmers et al. \cite{dettmers2022llm} note that the outliers in activation matrices in a few layers break the quantization of LLMs. They quantize outliers to FP16 and other activations to 8 bits to resolve this issue, thereby improving accuracy but bringing challenges in the implementation.

The GOBO technique \cite{zadeh2020gobo} quantizes model weights and embeddings. In every layer, \ApproxSign99.9\% weights follow Gaussian distribution, their technique divides weights into two groups: Gaussian and outliers (0.1\%). They calculate the mean and standard deviation of the layer's weights. Then, if the probability of a weight belonging to this distribution is below a threshold, it is considered an outlier. The outliers are stored as it is (FP32). Not storing outliers leads to large accuracy loss, but storing more than 0.1\% outliers provides a marginal improvement in accuracy.
The Gaussian group weights are quantized to just eight FP32 centroids (with $<$1\% accuracy loss) or sixteen FP32 centroids (with no loss). Then, only a 3b index to the dictionary is required for each Gaussian group weight. They reduce the number of multiplications by expressing ax+bx as (a+b)x. Their technique allows memory compression by 10$\times$. Their technique is faster and more energy efficient than the Tensor core-like design.

Mokey technique \cite{zadeh2022mokey} quantizes both weights and activations.  In the BERT-Large model, activations contribute more than 50\% of the memory footprint for token lengths above 512. Unlike weights, activations cannot be quantized in the offline stage, and they are also spread more widely than weights. Their quantization approach works in three steps (1) They generate a dictionary using ``agglomerative clustering'', which progressively merges nearby clusters to bring the cluster count to the desired value. It leads to more accurate models than K-means clustering. Their technique creates a bell-shaped pattern with a mean of zero and a standard deviation of one. On this pattern, they repeatedly apply agglomerative clustering to quantize it to 16b FX dictionary of centroids, called ``golden dictionary'' (GD). The same GD is used for all the models. GD is symmetric around zero, so it requires storing only half the entries.   (2) Their technique profiles the model to collect activation samples. Using them and the pre-known weight tensors, they adjust the GD to fit every tensor. Specifically, if the target tensor has mean and standard deviation as $m$ and $s$, they compute $GD*s +m$.

For every input tensor, their technique creates two dictionaries for each layer: a Gaussian group of values close to the mean, which covers $>$98\% weights and $>$95\% activations; and an outlier group of the remaining values. Although individual activations depend on the input, their layerwise distribution does not change much. (3) The weight tensors are encoded as indexes to their dictionaries. During inference, their technique outputs FX16 activations, which are converted into indices to corresponding dictionaries before storing them in memory.  All weights and activations are quantized to 16-entry dictionaries of FX16 centroids, which require only 4-bit indices. 

\begin{figure} [htbp]\centering
    \includegraphics[scale=0.43]{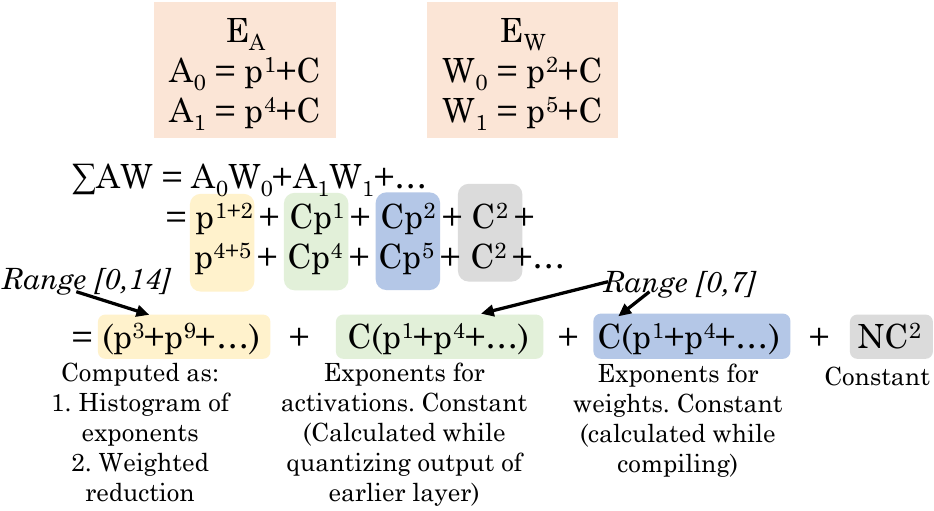}       
    \caption{Computation of output activations in Mokey technique \cite{zadeh2022mokey} }
    \label{fig:Mokey}
\end{figure}

For further simplification, they fit an exponential curve to model the GD since the near-mean values of a Gaussian distribution follow an exponential function of the form $a^{Q}+b$, where $Q$ is an integer. Then, instead of computing 256 possible products, they need to compute only 15 exponent sums based on the property $a^q*a^p = a^{q+p}$. This is shown in Figure \ref{fig:Mokey}. An actual MAC operation is required in the rare case when either weight or activation is an outlier.  Their technique outperforms previous quantization techniques.

\subsubsection{Combining pruning and quantization}
Joint-way compression  applies two or more compression methods, such as pruning and quantization, to gain additional savings. Most CNN model compression techniques, including pruning and quantization, are usually a two-step finetuning process. First, the model is pruned and retrained, followed by quantizing to low precision and finetuning. However, training the compressed model twice requires high computation time. Wang et al. \cite{wang2022deep} propose a technique where quantization and pruning steps are carried out in the same finetuning step on downstream tasks using the pretrained models. The activations are quantized using PACT \cite{choi2018pact} technique, while the weights are quantized using statistics-aware weight binning (SAWB) quantizer \cite{choi2019accurate}, which minimizes the quantization error utilizing first and second order moments. Their technique achieves 16x speedup over the unoptimized transformer models across language and vision tasks by employing 4-bit quantization and 50\% weight pruning. Mishra et al. \cite{mishra2021accelerating} first perform 2:4 sparsity-aware finetuning on a pretrained transformer, followed by 8-bit PTQ of the pruned weights. The resultant vanilla transformer and BERT models are 8x smaller than the original model, with no loss in accuracy.

\begin{figure*} [htbp]\centering
    \includegraphics[scale=0.45]{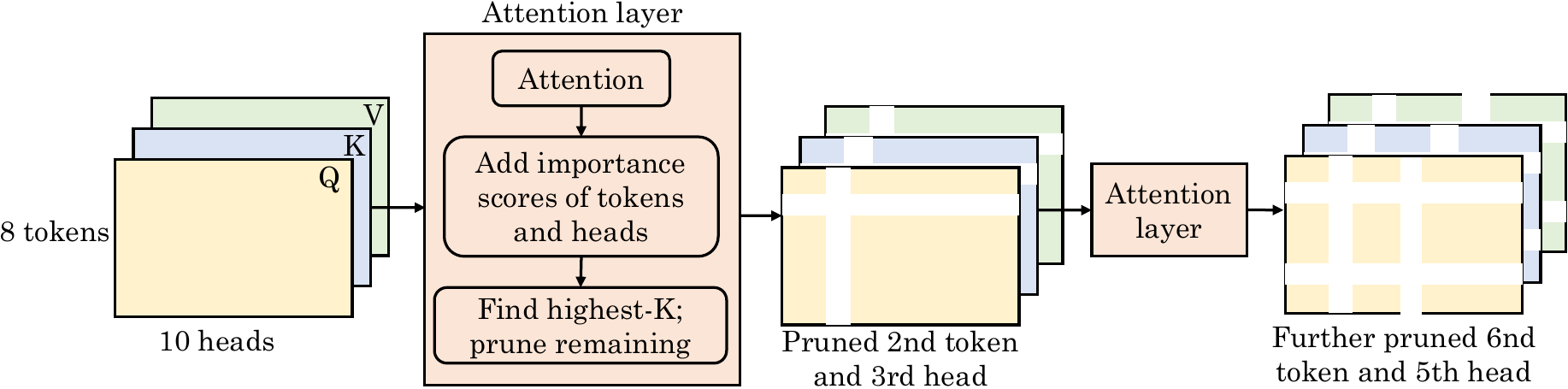}       
    \caption{Cascaded pruning in SpAtten technique  \cite{wang2021spatten}}
    \label{fig:CascadePruning}
\end{figure*}

The authors of SpAtten technique \cite{wang2021spatten} note that natural languages have many redundancies due to adverbs, articles, prepositions, etc. They assess token importance and then prune unimportant tokens based on attention probability values. More tokens are pruned for longer sentences since they generally have higher redundancies. Token pruning optimizes both attention and FC layers.
They further note that the correlations tracked by some of the heads in MHA are redundant. They assess the impact of each head on the output and then prune unimportant heads. Both pruning operations are done in a cascaded manner: A pruned element is removed from all the subsequent layers, and thus, deeper layers need to process fewer tokens/heads. This is shown in Figure \ref{fig:CascadePruning}. Token and head pruning decrease sentence and feature lengths, respectively. While conventional activation pruning decides based on activation magnitude, their technique decides based on attention probabilities accumulated over layers.

MNNFast and $A^3$ \cite{ham20203} do not reduce DRAM accesses, and hence, they are useful for compute-bound discriminative models (e.g. BERT) only and not only for memory-bound generative models (e.g. GPT-2). Whereas SpAtten reduces memory accesses and, thus, optimizes both types of models. $A^3$ prunes QKV vectors of a token in one head, and MNNFast prunes V vector; hence, they bring down computation of attention layers, but not FFN layers. In SpAtten, once a token is pruned, those computations in subsequent layers are also avoided, and thus, it optimizes FFN layers also.

They propose progressive quantization of attention inputs to further reduce memory accesses. They observe that when a few tokens are dominant, the quantization error is low, and just MSB (most significant bit) is sufficient, whereas, for a smooth distribution, both MSB and LSB (least signifcant bit) are required. Hence, they perform more aggressive quantization when some attention probabilities are dominant. They first bring MSBs of attention inputs and compute attention probabilities. If the highest probability is below a threshold, it indicates a flat distribution. Then, LSBs are also fetched, and probabilities are recomputed. Effectively, this technique uses more bits for harder inputs. Overall, they reduce memory accesses by performing extra computations, which benefits memory-bound models (e.g., GPT-2). This technique is not used for compute-bound models (e.g., BERT). Since attention layers frequently use softmax, which can reduce quantization errors, the quantization technique has negligible impact on accuracy.

They further propose a hardware accelerator. To support pruning, they design a highly parallel top-K engine, which finds K most influential heads or tokens in $O(n)$ time, whereas a sorting engine would take $O(n \log n)$ time. The engine supports the splitting and concatenation of LSBs and MSBs. SpAtten is evaluated across 30 well-known benchmarks, such as GLUE, SQuAD, and Wikitext-2, on BERT and GPT-2 networks. The co-designed pruned Transformer and accelerator reduce the memory access by 10.0× with negligible accuracy loss and achieve significant speedup and energy savings on a wide range of platforms, including Raspberry Pi ARM CPU, TITAN Xp GPU, Xeon CPU, and Nano GPU.

\begin{table}[htbp]
  \centering \footnotesize
  \caption{Quantization granularity }
    \begin{tabular}{|p{2cm}|p{6cm}|}
    \hline
      Granularity      & Remark   \\
    \hline
    Per-Tensor & Quantizes each tensor in the network differently. It accounts for different distributions of weights and activations.   \\
    \hline
    Per-Channel & It quantizes each channel of a tensor differently. It is more accurate than per-tensor, but requires more metadata and quantization steps during inference.    \\
    \hline
    Per-Head   & Each head in the MHA has different quantization scale. In language/vision tasks, some words/regions are more important, leading to different patterns of self-attention applied to different parts of the input.    \\
    \hline
     
   Per-Token/Patch \cite{yao2022zeroquant} &  It quantizes each element in the token/patch vector with different quantization scales\\

     \hline	

    Weight-group \cite{shen2020q,bondarenko2021understanding} &  Each group within a single weight matrix has different scale value. Especially useful when the embedding has a complex or non-uniform weight distribution.   \\
   \hline    
    \end{tabular}%
  \label{tab:Quantization_granularity}
\end{table}

\begin{figure*}
    \centering
    \includegraphics[scale=0.45]{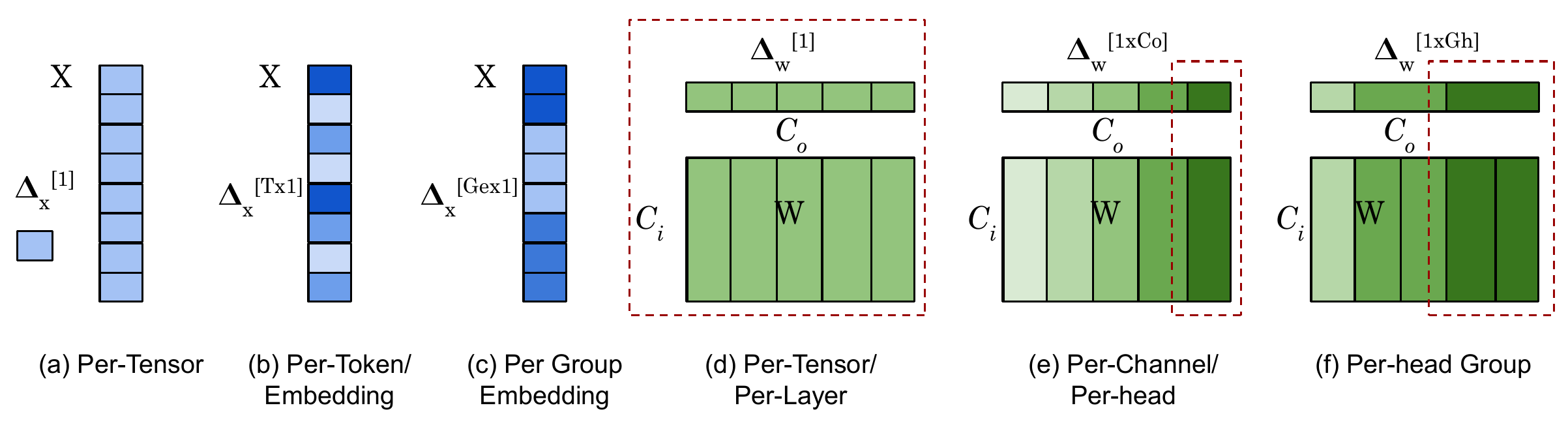}
    \captionsetup{justification=centering}    
    \caption{Quantization techniques. $\Delta$: Quantization parameter, C\textsubscript{i}: Input Channel Dimension, C\textsubscript{o}: Output Channel Dimension, T: Token Dimension, Ge: Number of Embedding Groups, Gh: Number of Head Groups}
    \label{fig:Quantization_Schemes}
\end{figure*}

\subsection{Classification  based on  granularity of quantization}

A transformer model can be quantized across different layers and levels of granularity, such as head-wise, channel-wise, embedding-wise, etc. The main challenge with transformer quantization is the difference in the range of parameters between different MHA and FFN layers, and therefore using similar scale parameters can lead to results in strong outliers. Also, previous works have shown that accuracy of a ViT model degrades by quantizing the layer norm and softmax operators \cite{liu2021post}. Table \ref{tab:Quantization_granularity} summarizes the methods based on their granularity and Figure \ref{fig:Quantization_Schemes} illustrates these methods.

\subsubsection{Per-Tensor Quantization}

Per-tensor quantization is a process of quantizing different tensors (weight and activation tensors) with different quantization scale parameters instead of possessing the same scale for the layer. This scheme is extremely beneficial to attain good accuracy as weight and activation tensors can have varying distributions.

\subsubsection{Per-Channel Quantization}

Per-channel quantization is a commonly used method for CNNs, where each channel of a weight filter/activation map is quantized to different precision or scale. Per-channel transformer quantization involves independently quantizing each dimension in each weight tensor matrix with different scale values. This method is more accurate than the per-tensor quantization as it considers each channel's statistical properties, giving more fine-grained control over each channel. However, this technique can be computationally more expensive than per-tensor as it requires storing more quantization parameters and quantization steps during inference.

\subsubsection{Per-Head Quantization}

Per-head quantization is a precise technique that is particularly useful in the case of a model with different attention head distributions, where each individual head in an MHA module is quantized with a different scale or precision. This process is extremely helpful for Transformer models or text/image inputs with varied data distribution. In language tasks, a few words/tokens in a sentence exhibit more importance than words to better understand the overall meaning of a sentence, leading to different patterns of self-attention applied to different parts of the input. In vision applications, a few regions in the input image may have more variability in color or texture than other set of pixels, leading to different feature distributions and requiring the model to learn features using a varied weight distribution.

\subsubsection{Per-token/Per-Embedding/Per-Patch Quantization}

The per-token and per-patch quantization methods refer to quantizing the sequence vector for language task and patch sequence for vision application, respectively, with different quantization scale values. This kind of fine-grained optimization for the specific characteristic in the token/patch can lead to better accuracy, although the number of additional quantization parameters increases with an increase in the embedding size.

ZeroQuant \cite{yao2022zeroquant} is based on multiple contributions in hardware-friendly quantization grouping scheme, layer quantization learning method and optimizing the inference backend to achieve speedup on different devices. This method applies dynamic group-wise quantization for the weight matrix and token-wise quantization for the activation tensor. The optimized inference backend reduces the cost of quantization and dequantization operations to achieve speedups on INT8-supporting tensor core GPUs. ZeroQuant proposes a knowledge distillation for Mixed Precision Quantization, where the Transformer is quantized layer-by-layer to Int4/Int8 precision. The method achieves optimal accuracy and speedups for BERT and GPT-3 (350M parameters) but cannot maintain accuracy for a large-scale GPT model with 175B parameters.

\subsubsection{Group Quantization}
Group quantization is a process of partitioning the weight and activation parameters within a layer into several groups and quantizing them with different precision or scale. Per-embedding group quantization \cite{bondarenko2021understanding} technique divides the embedding weights into groups, and each group is quantized differently using a different set of quantization parameters. This type of group quantization is particularly useful when the embedding has a complex or non-uniform weight distribution. It allows the quantization parameters to be tailored to the specific characteristics of each group. This technique can enhance the transformer model's accuracy by allowing the quantization parameters to be customized for each group of embedding weights. Q-BERT \cite{shen2020q} divides the layer parameters into groups, typically 128, and quantizes based on the importance of each group using second-order Hessian approximation in the range of 4 to 16. Q-BERT \cite{shen2020q} is a group-wise, mixed-precision quantization scheme that uses the second-order Hessian information \cite{dong2019hawq} to evaluate the sensitivity of the different tensors on the overall accuracy. Their technique achieves 3-bit and 8-bit weight and activation bitwidth, respectively.

\subsection{Classification based on resultant bit-width or data-type}

The easiest and most common quantization technique to preserve accuracy is to use integer arithmetic for linear operations while retaining floating-point precision for non-linear operations. However, this mixed precision approach requires the support of custom hardware with a large area for processing floating-point precision into integer-only hardware, thus introducing overhead for interaction between integer and FP precision. Several PTQ methods chose to leave softmax and activation functions in floating point precision as quantizing to low precision may lead to a signifcant drop in model accuracy.

\subsubsection{Integer-only Quantization}
 
Integer-only quantization methods propose efficient methods to quantize every layer, tensor, and operation, both linear and non-linear, to integer precision without affecting the accuracy. The int-only quantization methods eliminate the need for quantization and dequantization steps and enable entire network inference in a uniform integer domain.

I-BERT \cite{kim2021bert} quantizes all the layers in a BERT model, GELU, Softmax, and LayerNorm operations to Int8 precision and achieves 2.4-4x speedup over FP32 BERT model on Nvidia T4 GPU. I-ViT \cite{li2022vit} utilizes Dyadic quantization \cite{yao2021hawq} for int-only quantization of linear matrix multiplication. FQ-ViT \cite{lin2022fq} is a fully quantized model incorporating a log2 quantization and an integer softmax operation. Specifically, the authors introduce Log-Int-Softmax (LIS) to quantize the attention maps to 4 bits and replace the multiplication with a bitshift operator. Rock et al. \cite{rockint8} quantize all tensors and operations, including LayerNorm, SoftMax and GELU, of BERT to uniform Int8 precision. Their technique leads to only a minor drop in accuracy.

\subsubsection{Binarization/Ternarization}

Binarization and ternarization are extreme forms of quantization, where the model parameters are represented using only two and three values, respectively, thereby significantly reducing memory consumption. While binarization requires only \{1, -1\} values, ternary weights are either \{-1, 0, 1\}, and hence the memory reqiurement decreases by 32$\times$ and 16$\times$. respectively. These techniques often go hand-in-hand with PTQ or QAT methods and can be very well utilized for low-power devices such as MCU. Although binarization or ternarization can be very effective in significantly reducing the storage space, they can also impact the model accuracy due to extreme forms of low-precision implementation. While the floating point representation offers the highest level of accuracy, extreme quantization is the most efficient for low-power computational platforms.

\begin{figure}[h]
    \centering
    \includegraphics[scale=0.55]{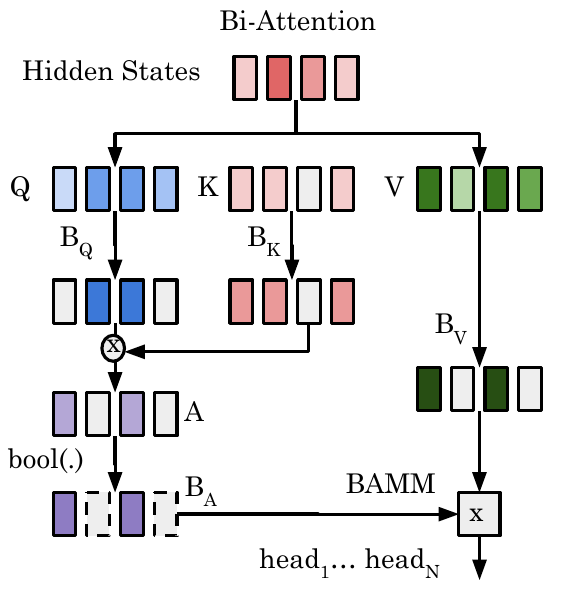}
    \captionsetup{justification=centering}
    \caption{Bi-attention \cite{qin2022bibert}  }
    \label{fig:bi_attention}
\end{figure}

Although binarization methods have achieved acceptable accuracy on Convolutional architectures \cite{liu2018bi, qin2020forward}, they cannot be directly applied and generalized on transformer models. For instance, BiBERT \cite{qin2022bibert} showed that directly binarizing the model parameters of BERT can cause a drop of 20 points in GLUE dataset \cite{wang2018glue}. BiBERT proposes a bi-attention module, depicted in Figure \ref{fig:bi_attention}, where the attention activations are binarized using a binarization function. The bi-attention module also replaces the softmax function with a bool function that binarizes the attention values to \{0, 1\}. The ignorance of softmax can cause quantization errors and damage the attention activations as the model is trained with softmax operation in the MHA modules throughout the network. The authors also propose Bitwise-Affine Matrix Multiplication (BAMM), which used to support the computation between the binarized attention score (B$_{A}$) and binarized vector (B$_{V}$) during inference.

Binarized Transformer (BiT) \cite{liu2022bit} is a multi-stage distillation-based binarization technique, where the authors first distill the information from a full precision model to a medium quantized model (lower precision but not binarized network). The intermediate reduced precision network then acts as a teacher model for the binarized student transformer network. This two-step process ensures good transferability between the large-sized full precision model and the binarized student model, which has been shown on the BERT model in the GLUE benchmark dataset. The authors of XTC \cite{wu2022extreme} first conducted a study to estimate the importance of hyperparameters and training strategies from several previous works. They finetune many existing compressed BERT models and conclude that longer training epochs and smaller learning rate values can aid quantization compression. Based on the observations, they propose methods to quantize BERT while achieving acceptable model performance and 50x size reduction. BinaryBERT \cite{bai2020binarybert} proposes a ternary weight splitting method, where a ternary model is chosen as a proxy, and the goal is to bridge the gap between a binary and full precision model. Binary Ensemble BERT (BEBERT) \cite{tian2022bebert} is a technique to overcome the limitations of previous binarization methods, which integrates several Binary BERT models in an ensemble fashion using AdaBoost. It incurs only a minimal accuracy loss of 0.3\% over the full-precision BERT model.

\subsubsection{Emerging Numerical Formats}

The emerging numerical formats such as Bfloat16 \cite{kalamkar2019study}, Microsoft Floating Point (MSFP) \cite{darvish2020pushing}, Tensorfloat32 \cite{choquette2021nvidia}, and FP8 \cite{van2023fp8, kuzmin2022fp8} provide better computational efficiency and accuracy for several transformer workloads. Bfloat16 is a 16-bit floating-point format developed by Google, consisting of 8 exponent and 7 mantissa bits. This precision offers significant speedups for BERT models while maintaining comparable accuracy to using FP32. MSFP is a floating-point datatype for efficient cloud inference. MSFP shows promising results on the BERT model as it incurs 3$\times$ lower cost compared to Bfloat16 with less than 1\% drop in accuracy. TensorFloat32 (TF32) is a precision format developed by Nvidia, which provides an easy path to accelerate FP32 on A100 and H100 GPUs. This format uses 1 sign bit, 8-bit exponent (same as FP32 precision) and 10-bit mantissa (same as FP16 format). Finally, FP8 is an 8-bit floating-point format that is mainly used for low-power inference, which allows a larger dynamic range at the expense of precision \cite{kuzmin2022fp8}. The exponent and mantissa bits for FP8 can be set dynamically. For example, FP8 can take any of the following two formats: (1-bit sign, 5-bit mantissa, 2-bit exponent) or (1-bit sign, 4-bit mantissa, 3-bit exponent). All these formats discussed here offer several benefits over the traditional floating-point precision, enabling faster and more efficient inference with reduced memory bandwidth.

\begin{comment}
\begin{table}[htbp]
  \centering
  \caption{Classification of Quantization methods based Precision}
    \begin{tabular}{|c|p{1.5cm}|p{1.5cm}|}
    \hline
     Precision &  Weight &  Activation\\
    \hline
    	Binary/Ternary & \cite{} &   \cite{} \\ \hline
	4-bit &  \cite{liu2021post} &  \cite{liu2021post}  \\ \hline
	6-bit &  \cite{liu2021post} &  \cite{liu2021post}  \\ \hline
     8-bit& \cite{zafrir2019q8bert, li2022accelerating, liu2021post} &   \cite{zafrir2019q8bert, li2022accelerating, liu2021post} \\
    \hline
    16-bit & \cite{li2022accelerating} &   \cite{li2022accelerating} \\
\hline
	
\end{tabular}%
  \label{tab:precision_classification}%
\end{table}%
\end{comment}

Table \ref{tab:precision_classification} provides classification of quantization methods based on the bitwidth to which weights or activations are quantized. It can be observed from the table that majority of the methods use 8 bits for transformer quantization as this level provides the least accuracy loss while reducing the precision.

\begin{table}[htbp]
  \centering
  \caption{Classification of Quantization methods based Precision}
    \begin{tabular}{|c|p{4cm}|}
    \hline
     Precision &  References\\
    \hline
    	Binary/Ternary & \cite{liu2022bit, wu2022extreme, bai2020binarybert, tian2022bebert, qin2022bibert} \\ \hline
	3-bit & \cite{frumkin2022cpt, zadeh2020gobo} \\ \hline
     4-bit & \cite{liu2021post, li2022psaq, frumkin2022cpt, yuan2022ptq4vit, zadeh2020gobo, wang2022deep} \\ \hline
	6-bit & \cite{liu2021post, yuan2022ptq4vit} \\ \hline
     8-bit & \cite{zafrir2019q8bert, li2022accelerating, liu2021post, li2022psaq, frumkin2022cpt, yuan2022ptq4vit, dettmers2022llm, wang2022deep, mishra2021accelerating} \\ \hline
    16-bit & \cite{li2022accelerating}\\ \hline
    %32-bit & \cite{}\\ \hline	
\end{tabular}
  \label{tab:precision_classification}
\end{table}

\begin{figure}[ht]
    \centering
    \subfloat[DeiT-Base]{{\includegraphics[scale=0.3]{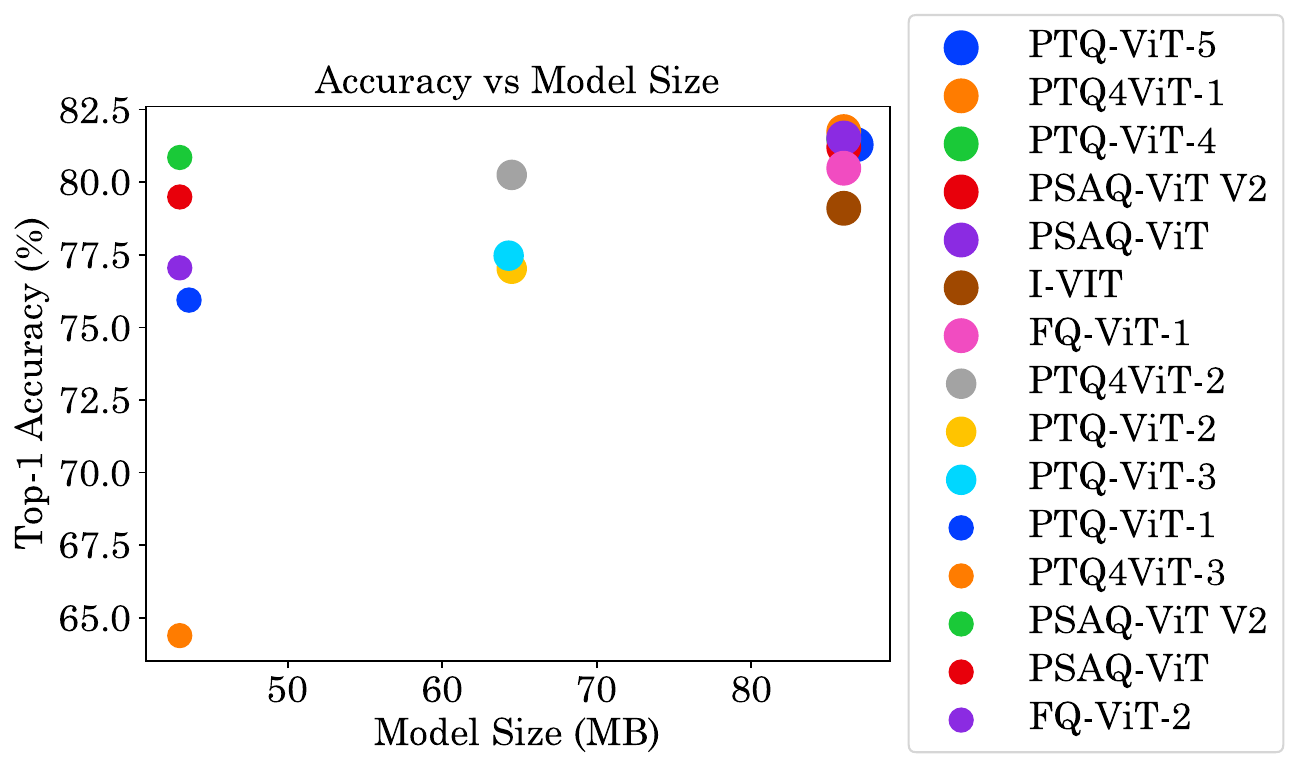}}}
    \hskip -3.5ex
    \qquad
    \subfloat[DeiT-Small]{{\includegraphics[scale=0.3]{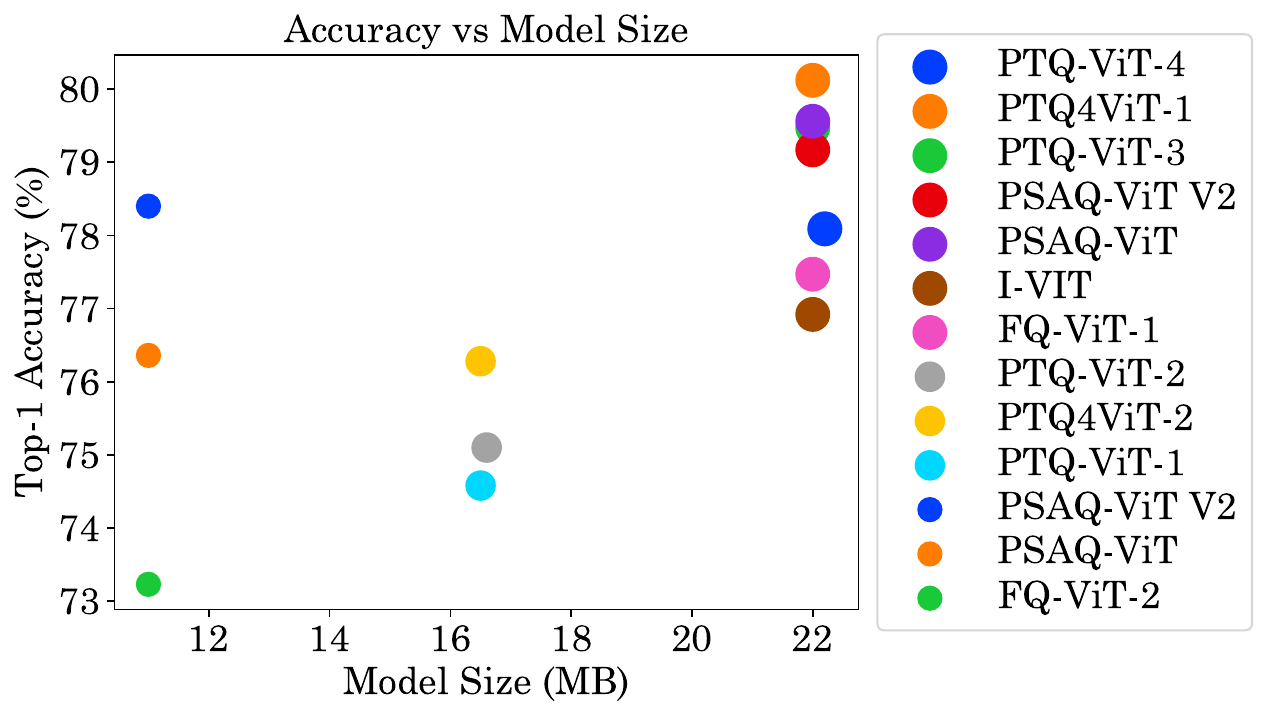}}}
    \hskip -3.5ex
    \qquad
    \subfloat[DeiT-Tiny]{{\includegraphics[scale=0.3]{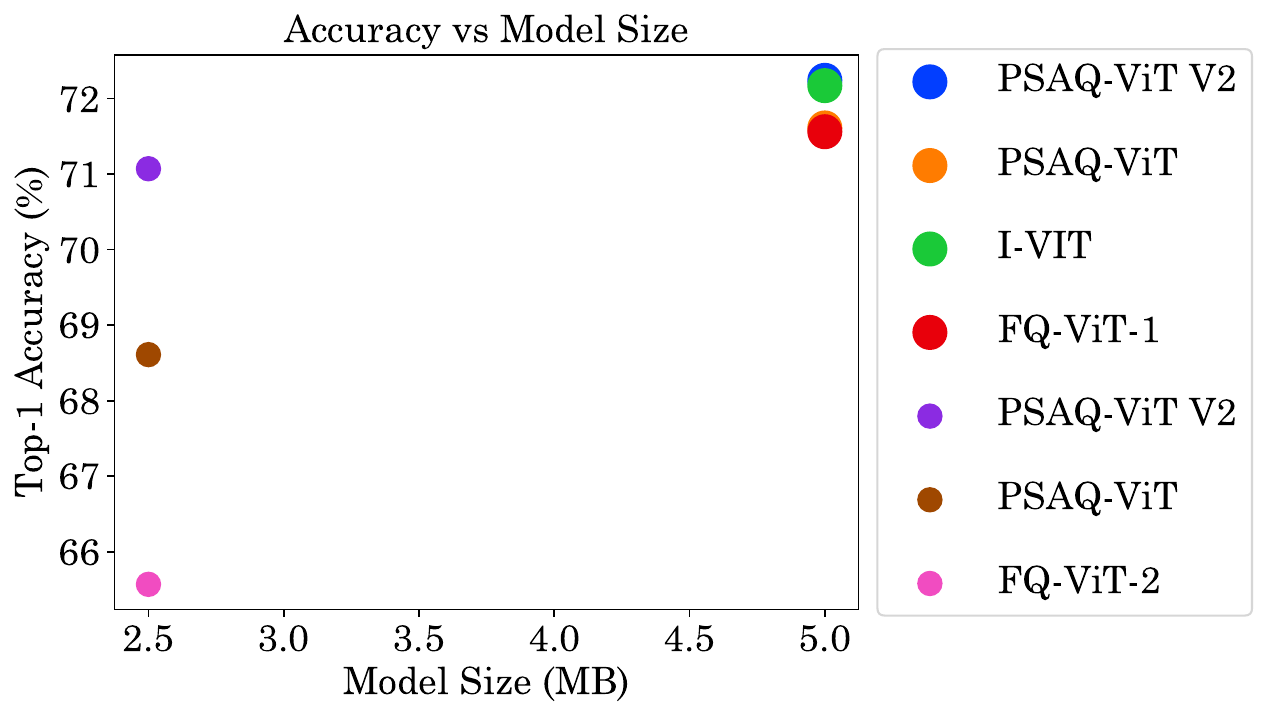}}}
    \hskip -3.5ex
    \qquad
   \caption{Comparing accuracy and model size of DeiT-base, -small and -tiny models on the following quantization technqiues: PTQ-ViT \cite{liu2021post}, 
PSAQ-ViT \cite{li2022patch}, 
PSAQ-ViT V2	 \cite{li2022psaq}, 
FQ-ViT \cite{lin2022fq}, 
PTQ4ViT \cite{yuan2022ptq4vit} and 
I-VIT \cite{li2022vit}. 
}
    \label{fig:quantization_compare}
\end{figure}

\subsection{Quantitative comparison of quantization techniques}
In Figures \ref{fig:quantization_compare}(a), \ref{fig:quantization_compare}(b) and \ref{fig:quantization_compare}(c), we compare the model size and accuracies of several quantization methods on DeiT-base, tiny and small backbone configurations, respectively. The methods we compare on DeiT are PTQ-ViT \cite{liu2021post}, PSAQ-ViT \cite{li2022patch}, PSAQ-ViT V2 \cite{li2022psaq}, FQ-ViT \cite{lin2022fq}, PTQ4ViT \cite{yuan2022ptq4vit} and I-VIT \cite{li2022vit}. 

The model size of the quantized model after quantizing the weight and activation parameters to same bitwidth using different quantization methods the remain same. This is because they differ only in the quantization function. However, the accuracy of a quantization method depends on the quantization process in terms of how well the knowledge is transferred from full precision model to quantized model or how well the method handles the outliers. For example, PSAQ-ViT V2 transfers the knowledge better than PSAQ-ViT as the teacher-student model in the former method is more efficient than the patch similarity in the later method. It transfers the attention distribution from high precision weights to low precision ones well. Ultimatley, the model performance of quantized models depend on the method used.

\section{Efficient transformer design} \label{sec:MCU_Mobile}
The complexity and computational demands of transformer models hinder their deployment in resource-constrained environments such as embedded systems. Therefore, researchers have developed lightweight transformer models, which still achieve high accuracy. These models can bring the benefits of deep learning to a wider range of tasks and devices. This section reviews the design principles of few such techniques. Table \ref{tab:LightweightNetwork} provides a classfication of these methods.

\begin{figure*}
    \centering
    \includegraphics[scale=0.4]{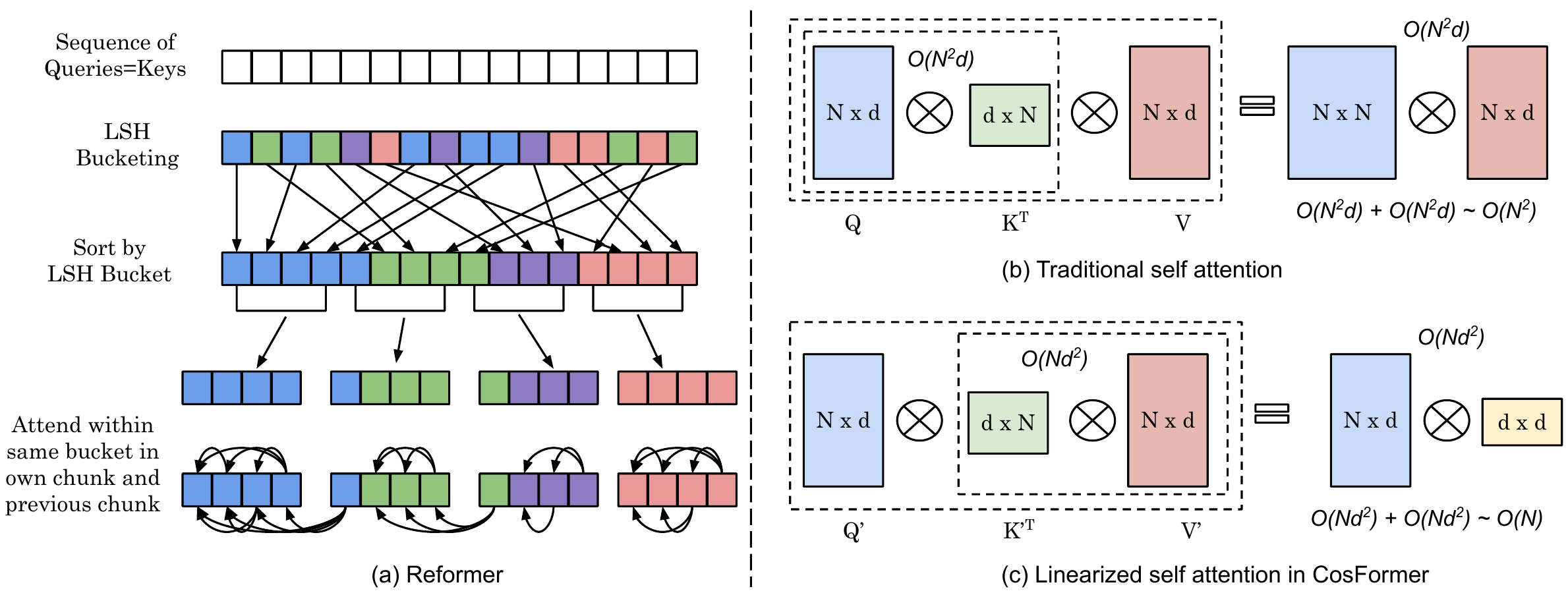}
    \captionsetup{justification=centering}    
    \caption{(a) Locality-sensitive hashing method in Reformer \cite{kitaev2020reformer}.  (b) Self-attention operation in the classical transformer with O(N$^2$) time complexity (c) The execution methodology of attention with O(N) complexity in CosFormer \cite{qin2022cosformer}}
    \label{fig:Reformer_cosformer}
\end{figure*}

\begin{table}[htbp]
  \centering \footnotesize
  \caption{Classification of lightweight network designs}
    \begin{tabular}{|p{1.4cm}|p{6cm}|}
    \hline 
Reducing complexity of attention from quadratic to linear & Applying attention across channel dimension instead of spatial dimension \cite{maaz2022edgenext}, using element-wise operation in attention computation   \cite{mehta2022separable}, replacing softmax attention with   ReLU attention \cite{choromanski2020rethinking} or  linear function \cite{qin2022cosformer} or low-rank factorization of attention \cite{wang2020linformer} \\\hline
Reducing number of parameters & Reducing number of heads  \cite{wang2020linformer}, depthwise convolution \cite{maaz2022edgenext}, group convolution and parameter-free layers \cite{lu2022tformer} \\\hline
Hybrid convolution-attention networks & replacing 3$\times$3 CONV in the bottleneck  residual block with MHA   \cite{srinivas2021bottleneck}, introducing local self-attention into convolution \cite{yang2022lite},  replacing MBConv with MobileViT block  in MobileNetV2   \cite{mehta2021mobilevit} \\\hline
Fusing local and global features & \cite{mehta2021mobilevit,chen2022mobile,yang2022lite} \\\hline
Domains & NLP \cite{qin2022cosformer,wang2020linformer,choromanski2020rethinking}, CV \cite{mehta2021mobilevit,mehta2022separable,chen2022mobile,yang2022lite,lu2022tformer,li2022efficientformer,maaz2022edgenext,srinivas2021bottleneck} \\\hline
 \end{tabular}%
  \label{tab:LightweightNetwork}%
\end{table}

\subsection{Methods for NLP}

To retain long-term/global information, the self-attention operation updates each token's representation by attending to all other tokens in the sequence. However, this requires a quadratic computation cost  with respect to the token sequence length for that layer. In addition, transformers perform batch-wise matrix multiplication  to learn  global representations in MHA. These operations incur high overhead. The goal of efficient transformer variants is to avoid such costly operations and replace quadratic-complexity MHA operations with linear-complexity operations.

\textbf{Reformer:} Kitaev et al. \cite{kitaev2020reformer} utilizes locality-sensitive hashing (LSH) method to lower the time complexity of traditional self-attention from $\mathcal{O}(N^2)$ to $\mathcal{O}(N logN)$, where N is the sequence length. The general idea of LSH is to reduce the dimensionality of the data while preserving the similarity between the data points of high dimension. LSH attention involves using locality-sensitive hashing to efficiently find the nearest neighbors among the keys. This is achieved by assigning each vector to a hash using a hashing scheme that ensures nearby vectors get the same hash with high probability, and then employing random projections to create multiple hashes for each vector. The tokens within each chunk are attended among themselves, resulting in $\mathcal{O}(N logN)$ complexity. The entire schematic of LSH in Reformer is summarized in Figure \ref{fig:Reformer_cosformer}(a). The authors also use the concept of reversible residual layers, where the activations are stored only once during training, instead of L times, where L is the number of layers in the model. The resultant transformer model is efficient in terms of both latency and memory due to these two technqiues.

\textbf{Linformer:} Wang et al. propose Linformer \cite{wang2020linformer}, which decomposes the dot-product attention operation into small chunks of attention multiplications through linear projections. Therefore, the quadratic self-attention can be executed using a low-rank factorization of the original attention. The number of heads is reduced compared to the baseline transformer, and the decrease in the number of heads in MHA is compensated with long input sequences. While the transformer inference latency increases with an increase in sequence length, the latency of the Linformer remains relatively flat. Thus, for long input sequences, Linformer provides significant speedup over  transformer.

\textbf{Performers:} The authors of Performers \cite{choromanski2020rethinking} introduce a new approach called FAVOR+ to estimate the softmax self-attention for models that have longer sequence lengths. The Performers method require linear space and time complexity. Therefore, this method is more efficient in terms of both latency and memory compared to quadratic $\mathcal{O}(N^2)$ transformer \cite{vaswani2017attention} and $\mathcal{O}(N logN)$ Reformer \cite{kitaev2020reformer} attention methods. 
The authors utilize positive random features to generate an estimate of the softmax function with a positive feature map, which is essential for ensuring stable training. The Performer model outperforms other models in terms of speed, memory usage, and performance. The authors further demonstrate that it is not necessary to approximate softmax to obtain good results. Instead, they use ReLU attention to achieve superior performance when training from scratch.

\begin{figure*}
    \centering
    \includegraphics[scale=0.4]{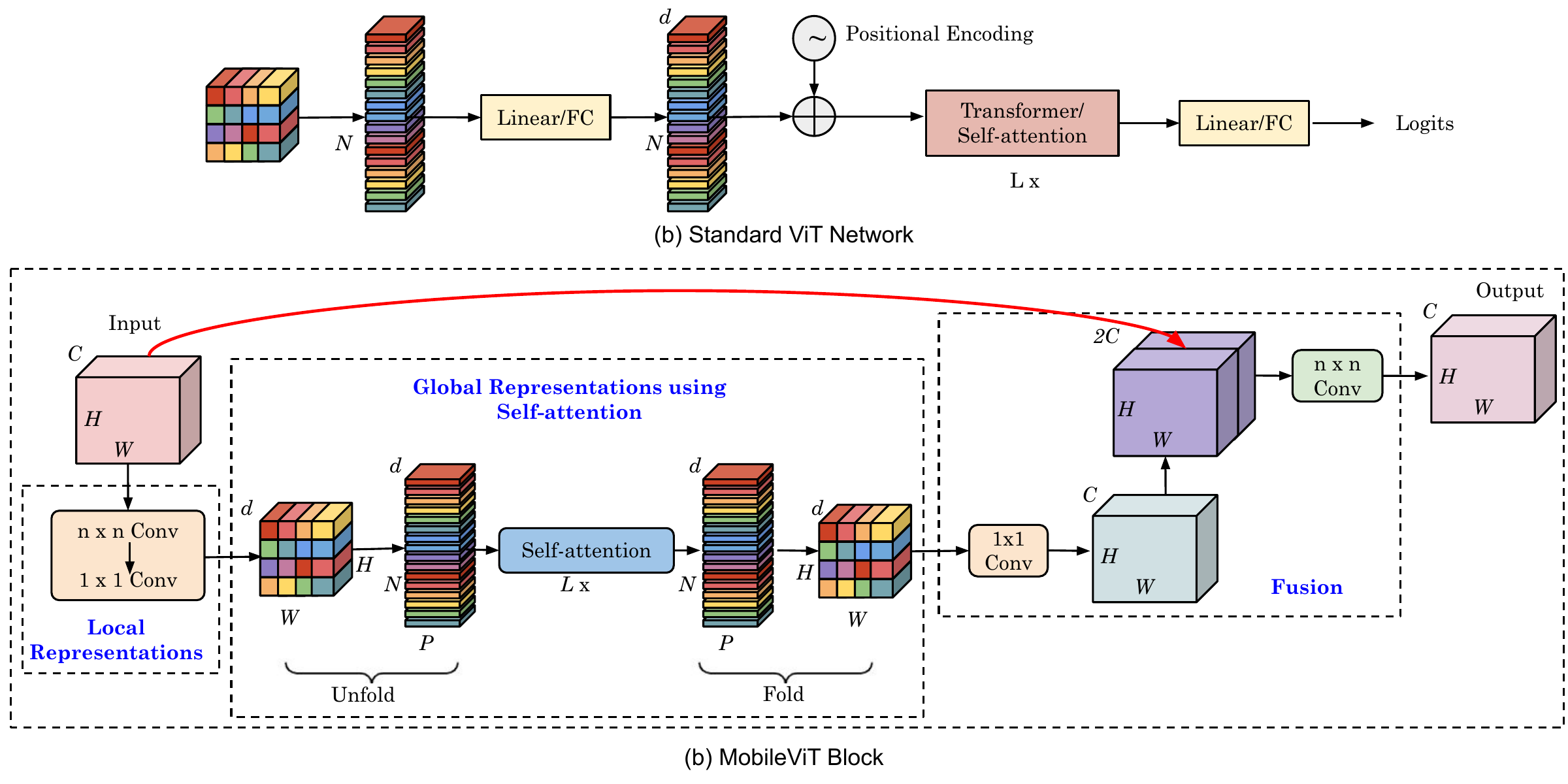}
    \captionsetup{justification=centering}    
    \caption{(a) Standard ViT model proposed by Dosovitskiy et al. \cite{dosovitskiy2020image}. It first flattens the input image into several patches. This resultant patches are processed using a standard FC layer, followed by series of self-attention modules. (b) MobileVit Block \cite{mehta2021mobilevit}. This block replaces a few MBConv modules in MobileNetV2. The input is a feature map from the previous layer. The local features are extracted using an n $\times$ n Convolution layer and global information is modeled using a transformer. The local and global features are fused together using residual connection.}
    \label{fig:MobilevitV1_Vit}
\end{figure*}

\textbf{CosFormer:} Qin et al. \cite{qin2022cosformer} propose Cosformer, which replaces the quadratic softmax attention operation with a linear function. 
The features are passed through ReLU to enforce non-negative property before computing the similarity scores to avoid aggregating negatively-correlated contextual information. The attention weights are re-weighted using the cosine function to enhance local correlations, which usually contain relevant token information for NLP tasks. 
The quadratic attention operation can be achieved in linear complexity using the matrix product property as follows: 
$\left( \phi\left( Q \right) \phi \left( K \right)^{T} \right) V = \phi\left( Q \right)\left( \phi \left( K \right)^{T} V\right)$. Figure \ref{fig:Reformer_cosformer}(b) depicts the multiplication in traditional transformer, where Query (Q) and Key (K) are first multiplied resulting in an attention matrix, followed by multiplication with Value (Q). Figure \ref{fig:Reformer_cosformer}(c) depicts the linear complexity attention in CosFormer, where first the Key and Value matrices are multiplied, followed by attention-Key dot product. The time complexity in both the cases are depicted in Figures \ref{fig:Reformer_cosformer}(b) and (c).

\subsection{Methods for Computer Vision}

While efficient transformer design methods for NLP tasks focused predominantly on optimizing   attention, efficient models for vision applications aim to curate lightweight models that can run efficiently on resource-constrained devices.  
In this subsection, we focus on such efficient variants in computer vision field.

\textbf{MobileViT:} MobileViT \cite{mehta2021mobilevit} combines the strengths of standard convolution and attention mechanism and presents a new approach for better local-global context fusion. The authors replace the traditional Mobile Inverted Residual Convolution (MBConv) layer in the upper stages of the MobileNetV2 model \cite{sandler2018mobilenetv2} with the MobileViT block to obtain better global representation. Using a self-attention mechanism, the MobileViT block substitutes the local processing in convolutions with global processing. The local representations and the global information are concatenated to generate enhanced local-global representations. The hybrid unit helps to learn better representations with fewer parameters and a simple training procedure. This modified approach surpasses lightweight CNNs with a similar parameter budget on the ImageNet dataset. Although MobileViT reduces the number of parameters and increases accuracy, the MHA module is still a performance bottleneck as it requires quadratic complexity. The MHA module is directly inherited from the original vanilla transformer. The sequence of operations and details of each module is summarized in Figure \ref{fig:MobilevitV1_Vit}.

\begin{figure}[htbp]
    \centering
    \includegraphics[scale=0.48]{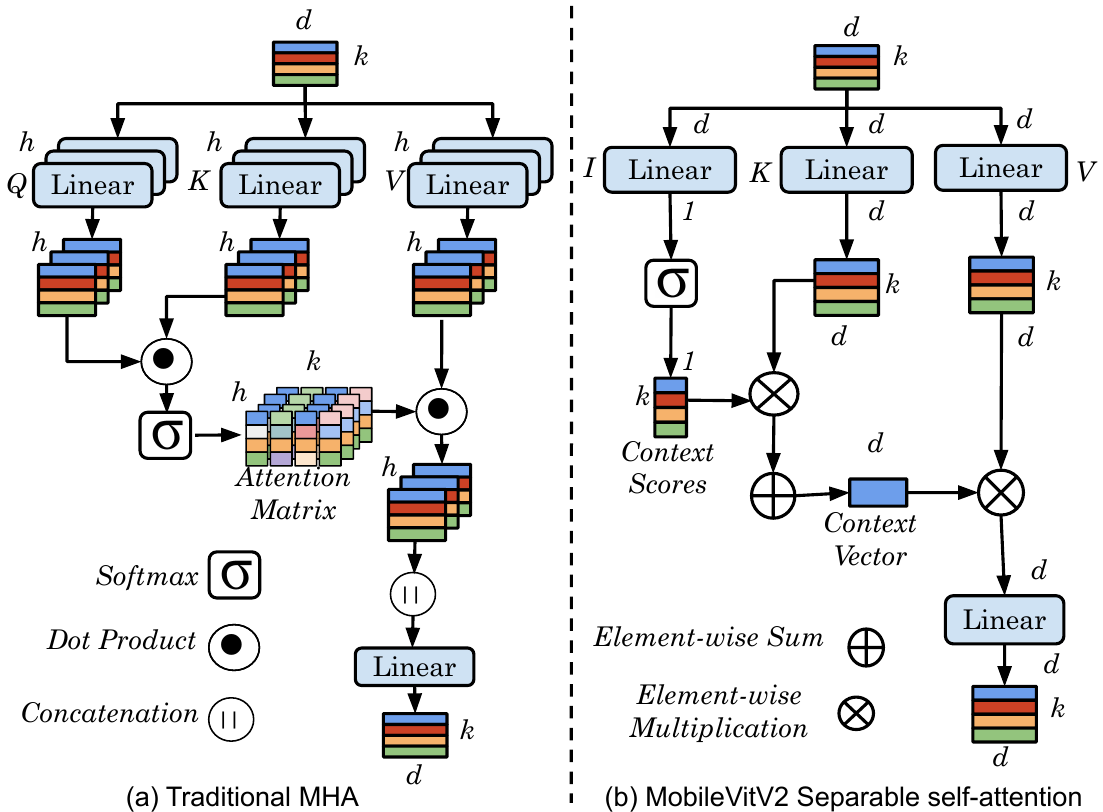}
    \captionsetup{justification=centering}
    \caption{(a) Traditional self-attention in MHA. (b) Separable self-attention in MobileViTV2 \cite{mehta2022separable}, which replaces the expensive batch-wise matrix multiplication with element-wise linear operations.}
    \label{fig:MobileViTv2}
\end{figure}

\begin{figure}[htbp]
    \centering
    \includegraphics[scale=0.5]{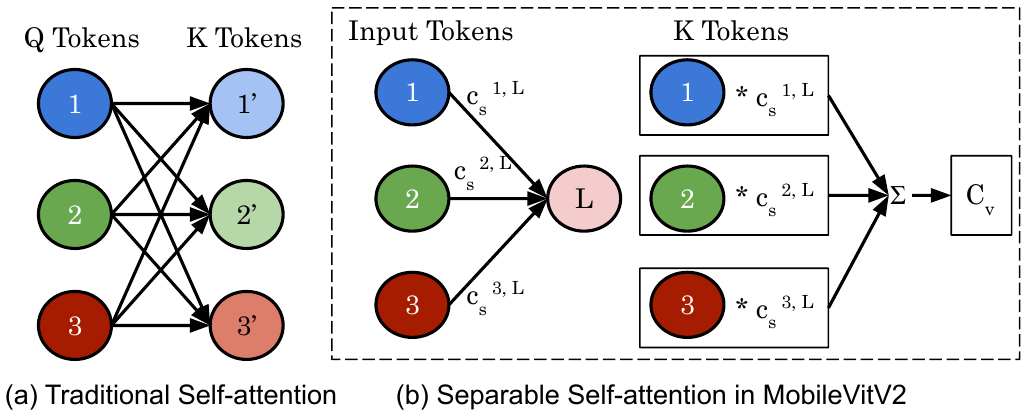}
    \captionsetup{justification=centering}
    \caption{(a) Computation in the traditional self-attention Each token in Query attends to all other tokens in Key, resulting in O(N$^2$) complexity. (b) Comparison of separable attention \cite{mehta2022separable} with self-attention \cite{vaswani2017attention}. In this enhanced module, the input is processed using three branches Input (I), Key (K) and Value (V). The input I maps each token in input X to a scalar through an FC layer of Weights W$_I$. This weights serve as Latent node L. The separable attention computes context scores only with respect to this latent token L, resulting in O(N) complexity. The context scores c$_s$ are used compute context vector c$_v$, which is equivalent of attention matrix in self-attention.}
    \label{fig:separable_attention}
\end{figure}

\textbf{MobileViTV2:} MobileViTV2 \cite{mehta2022separable}  enhances the MobileViT network by introducing a separable self-attention layer. This enhanced separable module uses an element-wise operation for self-attention computation, thereby bringing down the MHA complexity from quadratic  to linear, with respect to the patch sequence length. The MobileViTV2 model replaces the expensive batch-wise matrix multiplication with a context-aware element-wise operator, as depicted in Figure \ref{fig:MobileViTv2}. The separable self-attention layer is explained in detail in Figure \ref{fig:separable_attention}, which requires only $\mathcal{O}(N)$ complexity. Although MobileViTV2 has higher number of parameters than MobileViTV1, its latency is lower. The MobileViTV2 network achieves 75.6\% accuracy on the ImageNet dataset, performing better than MobileViT by 1\% and running 3.2$\times$ faster on iPhone12.

\textbf{Mobile-former:} The Mobile-former network \cite{chen2022mobile} employs parallel \underline{Mobile}NetV2 and Trans\underline{former} modules with two-way bridges (Mobile$\rightarrow$Former and Mobile$\leftarrow$Former). 
It seeks to achieve a bidirectional fusion of local and global feature representations at each level in the network.  
The input to the MobileNetV2 is the input image, while the transformer module takes a few learnable tokens as input, which are used to encode the global features of the input image. The mobile block is an MBConv block from MobileNetV2 \cite{sandler2018mobilenetv2}, where the ReLU is replaced by dynamic ReLU \cite{chen2020dynamic}. Mobile$\rightarrow$Former employs lightweight cross attention to fuse the local representations with global features. The Former block is a standard transformer module that consists of MHA and FFN. The Mobile$\leftarrow$Former unit bridges from global to local features. The Mobile-Former network attains an accuracy of 77.9\% with 294M FLOPs and exceeds MobileNetV3 accuracy by 1.3 percentage point on the ImageNet dataset. The network also outperforms DETR \cite{carion2020end} by 1.1 AP on the object detection task.

\begin{figure}
    \centering
    \includegraphics[scale=0.4]{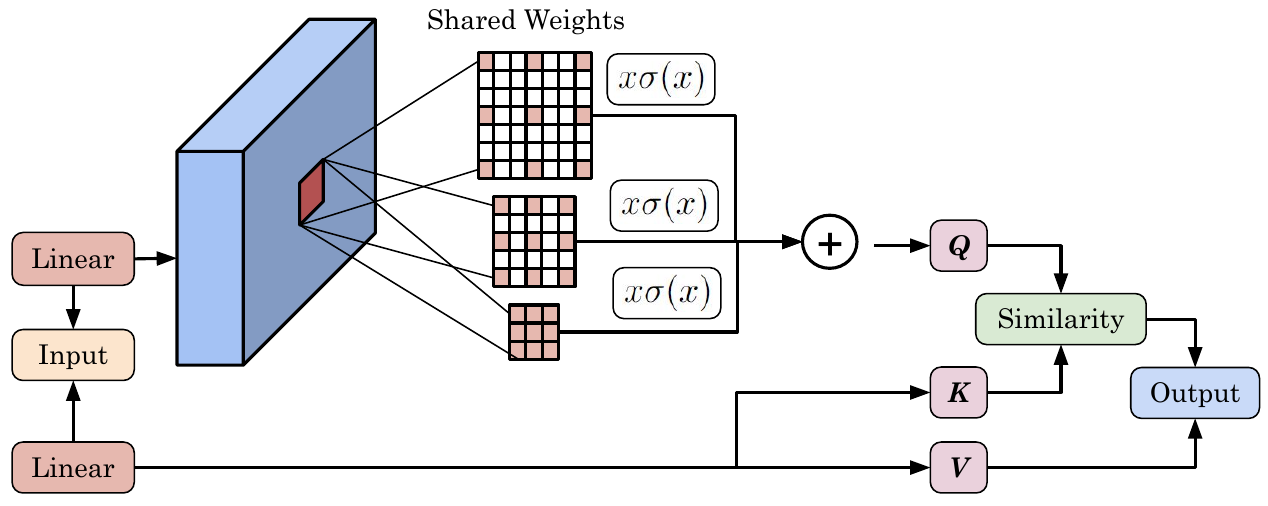}
    \captionsetup{justification=centering}
    \caption{Recursive Atrous Self-Attention (RASA)}
    \label{fig:RASA}
\end{figure}

\textbf{LVT:} Lite Vision Transformer \cite{yang2022lite} propose two new layers, viz., Convolutional Self-Attention (CSA)  and Recursive Atrous Self-Attention (RASA) in the transformer architecture. CSA is used in the first stage of LVT whereas RASA is used in last three stages of LVT. They note that CONV layer is more effective in extracting low-level features. They introduce local self-attention into a 3*3 CONV. Compared to transformers, this leads to enhanced low-level features, improving the generalization capabilities. RASA uses multi-scale context with a single kernel for computing similarity between Q and K, as shown in Figure \ref{fig:RASA}. RASA processes features recursively using ``atrous
self-attention'' as the activation function. This allows RASA to improve model depth without increasing parameter-count.

\textbf{TFormer:} TFormer \cite{lu2022tformer} network proposes techniques for making CV transformer transmission-friendly. It employs many \textit{parameter-free} on-the-fly operations along with traditional attention multiplication. Specifically, the authors of TFormer proposed a hybrid layer and partially connected feed-forward network (PCS-FFN), which replace the MHA module and FFN unit, respectively. The hybrid layer comprises only parameter-free layers, such as max and average pooling, whereas the PCS-FFN layer is based on group convolution to reduce the model parameters. These ideas reduce the size of the trained TFormer model that is transmitted from cloud to the inference hardware.
The TFormer models outperform DeiT \cite{touvron2021training} and PVT \cite{wang2021pyramid} in terms of the number of parameters and model accuracy on a wide range of tasks.

\begin{figure}[H]
    \centering
    \includegraphics[scale=0.48]{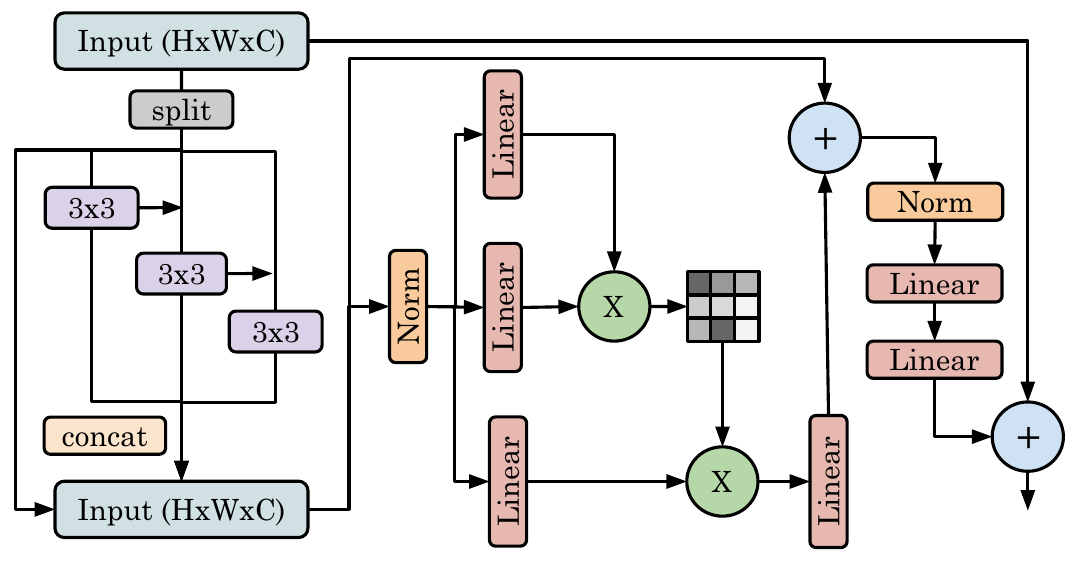}
    \captionsetup{justification=centering}
    \caption{EdgeNeXt \cite{maaz2022edgenext}}
    \label{fig:EdgeNeXt}
\end{figure}

\textbf{EdgeNeXt:} EdgeNeXt \cite{maaz2022edgenext} is a lightweight architecture with hybrid convolution and attention modules. 
The authors propose a \enquote{split depth-wise transpose attention} (STDA) unit to split the input tensor into multiple channel groups. This module utilizes N$\times$N depthwise convolution and pointwise operations for spatial mixing and channel mixing, respectively. 
EdgeNeXt uses cross-covariance attention that applies attention operation across the channel dimension instead of the spatial dimension to reduce the complexity from quadratic to linear. The network also employs adaptive kernel sizes, where small kernel sizes are used in the early stages, and large sizes are used in the latter part of the Transformer model. Figure \ref{fig:EdgeNeXt} illustrates the proposed EdgeNeXt module, and the network attains 71.2\% accuracy on the ImageNet dataset with only 1.3M parameters.

\begin{figure}[H]
    \centering
    \includegraphics[scale=0.48]{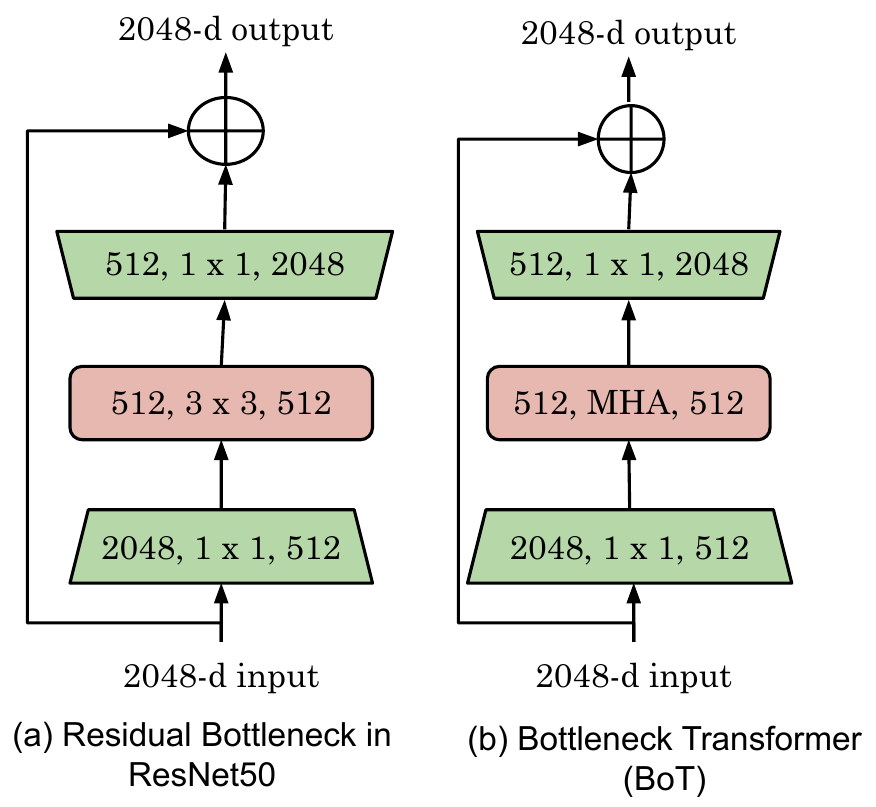}
    \captionsetup{justification=centering}
    \caption{(a) Bottleneck Module in the traditional ResNet Family \cite{he2016deep}. (b) BoTNet \cite{srinivas2021bottleneck} module, which replaces the 3$\times$3 Convolution with MHA}
    \label{fig:BoTNet}
\end{figure}

\textbf{BoTNet:} BoTNet network \cite{srinivas2021bottleneck} seeks to improve the global representation of feature maps.  Bottleneck Transformer (BoT) module is obtained simply by replacing the 3$\times$3 convolution in the bottleneck residual block with MHA. Figure \ref{fig:BoTNet}(a) illustrates the bottleneck module in the ResNet50 network \cite{he2016deep}, while Figure \ref{fig:BoTNet}(b) depicts the enhanced BoT module, which is obatined by replacing the middle convolution with a self-attention unit. The overall BoTNet model is constructed by replacing the last three bottleneck modules of a standard ResNet50 model with the BoT module without changing any other hyperparameters. BoTNet achieves 84.7\% accuracy on the ImageNet dataset and is 1.64 times faster than EfficientNet on the TPUV3 accelerator.

\subsection{Quantitative comparison of lightweight CV transformer techniques}

\begin{figure}[H]
    \centering
    \includegraphics[scale=0.41]{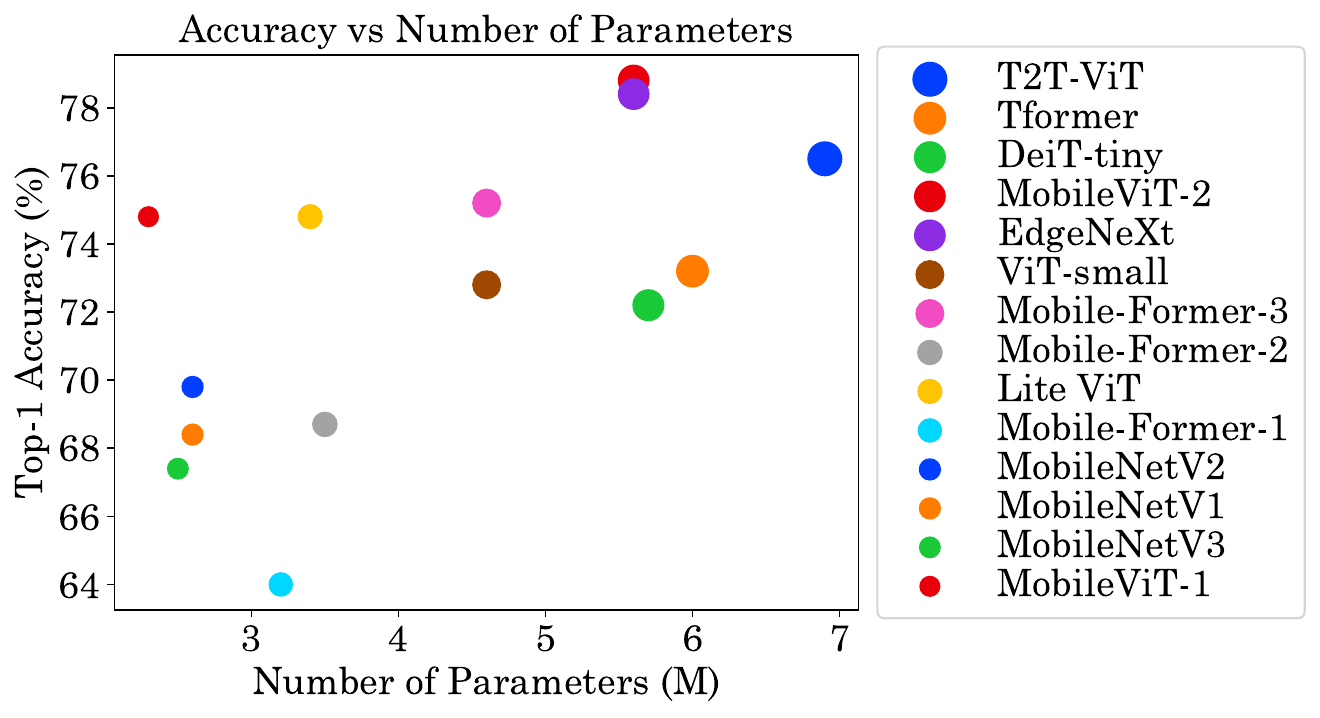}
    \captionsetup{justification=centering}
    \caption{Comparison of ImageNet top-1 accuracy and number of parameters of lightweight vision transformer models: MobileViT \cite{mehta2021mobilevit}, MobileViTv2 \cite{mehta2022separable}, Mobile-former \cite{chen2022mobile}, LVT \cite{yang2022lite}, TFormer \cite{lu2022tformer}, EdgeNeXt \cite{maaz2022edgenext}}
    \label{fig:lightweight_MHSA_compare}
\end{figure}

In Figure \ref{fig:lightweight_MHSA_compare}, we compare the ImageNet top-1 accuracy and parameter-count of vision transformer models summarized in this section with the baseline transformer models, viz.,  DeiT-tiny \cite{touvron2021training}, ViT-small \cite{dosovitskiy2020image}, T2T-ViT \cite{yuan2021tokens} and CNN models, viz., MobileNetV1 \cite{howard2017mobilenets}, MobileNetV2 \cite{sandler2018mobilenetv2}, MobileNetV3 \cite{howard2019searching}. A positive correlation (not necessarily linear) exists between the number of parameters and accuracy. Larger models generally achieve higher accuracy scores, although there are diminishing returns as the model gets larger.

This indicates that increasing the number of model parameters can improve the model's ability to learn and generalize to new data. There is a wide variation in accuracy scores across different models, even when they have a similar number of model parameters. This suggests that other factors, such as the model architecture and training process, are crucial in determining model performance. Also, adding more parameters beyond a certain point may not significantly improve accuracy. For instance, T2T-ViT \cite{yuan2021tokens} has the highest number of parameters among the compared models, though the accuracy is less than that of MobileViTV2. Some models, such as MobileViT,  achieve high accuracy with relatively small model sizes, indicating that they are more efficient at learning from the available data.

\section{Neural architecture search} \label{sec:NAS}

Neural architecture search (NAS) is a rapidly evolving research field which automates the end-to-end manual design process of a neural network for a given task and dataset. Hardware-aware NAS (HW-NAS) is a class of NAS that focuses on automatically searching accurate and hardware-efficient models \cite{chitty2022neural}. 
The searched transformer networks through NAS and HW-NAS often outperform manually designed models in terms of accuracy and compute performance on the target hardware \cite{chitty2022neural_transformer}. 

This section first provides a brief overview of NAS techniques targeting the transformer family and then summarizes a few works which utilize NAS for model compression. Table \ref{tab:NAStable} presents a classification of transformer NAS methods.

\begin{table}[htbp]
  \centering \footnotesize
  \caption{Classification of transformer NAS methods}
    \begin{tabular}{|p{2.7cm}|p{5cm}|}
    \hline
    \multicolumn{2}{|c|}{Search method} \\
    \hline
    Reinforcement Learning (RL)    & \cite{liu2021uninet} \\
    \hline
    One-shot/Differentiable & \cite{su2021vitas, gong2021nasvit, zhang2022vision, zhao2021memory, shi2021efficient, you2022shiftaddnas, zhao2021automatic, latifi2022efficient} \\
    \hline
    Evolutionary & \cite{so2019evolved, so2021searching, benmeziane2022real} \\
    \hline
    Once-for-all Search & \cite{chen2021autoformer, liao2021searching, chen2021searching, chen2021glit, yang2022searching, wang2020hat, yin2021autotinybert} \\
    \hline
    KD & \cite{wang2022lighthubert, zhang2022autodistill, kim2022revisiting} \\
    \hline
    Accuracy Predictor & \cite{luo2021lightspeech} \\
    \hline
    Training-free search & \cite{zhou2022training, javaheripi2022litetransformersearch} \\
    \hline
    \multicolumn{2}{|c|}{Search Parameter} \\
    \hline
    Head Number & \cite{chen2021autoformer, su2021vitas, zhou2022training, liao2021searching, chen2021searching, ni2022nasformer, shi2021efficient, luo2021lightspeech} \\
    \hline
    QKV Dimension & \cite{chen2021autoformer, zhou2022training, chen2021searching, ni2022nasformer} \\
    \hline
    FFN Dimension & \cite{chen2021autoformer, su2021vitas, zhou2022training, liao2021searching, chen2022auto, chen2021searching, chen2021glit, zhu2021autotrans, you2022shiftaddnas} \\
    \hline
    Embedding Dimension & \cite{chen2021autoformer, su2021vitas, zhou2022training, chen2021searching, you2022shiftaddnas} \\
    \hline
    Network Depth & \cite{chen2021autoformer, zhou2022training, chen2021searching, gong2021nasvit, zhang2022vision, zhu2021autotrans} \\
    \hline
    Kernel \& Channel size & \cite{chen2021glit, gong2021nasvit, yang2022searching, liu2021uninet, so2021searching, shi2021efficient, luo2021lightspeech} \\
    \hline
    \multicolumn{2}{|c|}{Network Type} \\
    \hline
    Transformer (enc-dec) & \cite{so2019evolved, wang2020hat, you2022shiftaddnas, luo2021lightspeech, javaheripi2022litetransformersearch} \\
    \hline
    BERT  & \cite{chen2020adabert, xu2021bert, gao2022autobert, zhao2021automatic} \\
    \hline
    Computer Vision  & \cite{chen2021autoformer, su2021vitas, zhou2022training, liao2021searching, chen2022auto, chen2021searching, ni2022nasformer, chen2021glit, gong2021nasvit, you2022shiftaddnas, benmeziane2022real} \\
    \hline
    \end{tabular}%
  \label{tab:NAStable}%
\end{table}

\subsection{Overview of Neural Architecture Search}

\begin{figure*}
    \centering
    \includegraphics[scale=0.43]{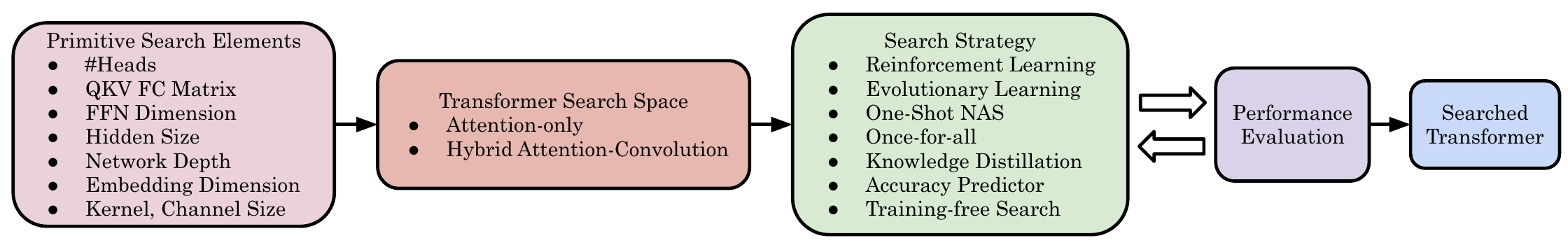}
    \captionsetup{justification=centering}    
    \caption{Overview of transformer Neural Architecture Search Methodology. The main step in the NAS process includes building primitive search elements, search space followed by applying a search methdology to search for an efficient transformer model }
    \label{fig:NAS_flow}
\end{figure*}

A NAS method typically consists of three components: \textbf{(1)} search space, \textbf{(2)} search strategy and \textbf{(3)} evaluation phase. The first step is to efficiently curate the search space consisting of all possible architectures. The transformer search space typically consists of the architectural hyperparameters, such as Q-K-V FC dimensions, the number of heads in the MHA unit and the inner dimensions of the FC layer in the FFN/MLP module at each level of the network. The depth of the transformer model, i.e., the number of encoder or decoder layers, is also considered in the search space. The vision transformer model includes the patch size  and patch embedding size in the search space, while the hybrid attention-convolution search space considers the kernel and filter size of the convolution operation. The search strategy automatically discovers top-performing architectures from the predefined search space for a dataset. The search algorithm outlines a methodology to find an optimal model from a pool of all possible networks. The evaluation phase is the most critical step, which evaluates the performance of the predicted architecture. It compares different neural architectures predicted by the search algorithm to properly guide the search process in the direction of finding an optimal model. The flow of a transformer NAS algorithm is illustarted in Figure \ref{fig:NAS_flow}. 

Since a model searched for one hardware platform may not be optimal for other platforms, HW-NAS methods \cite{chitty2022neural} include the performance metrics of the underlying hardware platform in the search method as a multi-objective optimization function.  For instance, Hardware-aware Transformers (HAT) \cite{wang2020hat} showed that the NLP transformer specialized for GPU runs faster on GPU than CPU and vice versa for similar validation accuracy.

\subsection{Classification based on Transformer Search Space}

The transformer search spaces can be classified into two types based on the operations present in the primitive element set.

\subsubsection{Attention-only Search Space}

The search elements in the self-attention-only search space include hyperparameters of the transformer attention module, such as number of heads, FFN dimension, and QKV FC matrix sizes. HAT \cite{wang2020hat} and AutoFormer \cite{chen2021autoformer} are examples of such search methods based on the vanilla transformer and ViT, respectively.

\subsubsection{Hybrid Attention-Convolution Search Space}

The hybrid attention-convolution search space consists of attention and convolution parameters within the transformer backbone network. Although convolution operations are primarily used in vision applications, they are also utilized in a few NLP applications, such as text classification. TextNAS \cite{wang2020textnas}, GLiT \cite{chen2021glit} and BurgerFormer \cite{yang2022searching} are examples of hybrid search space, which includes the kernel and channel size of the convolution along with the self-attention parameters.

\subsection{Classification based on Search Method}

The search algorithm is designed to find the best-performing architecture from the predefined set of primitive operations without significant human intervention. The search strategy has evolved greatly over the last few years, which includes reinforcement learning (RL) \cite{zoph2016neural}, One-shot/Differentiable \cite{liu2018darts}, Evolutionary \cite{chen2019renas}, Once-for-all \cite{cai2019once}, Random search \cite{li2020random}, low/zero cost proxy search \cite{mellor2021neural}, Bayesian Optimization \cite{white2021bananas}. We now describe the search methodology and summarize a few transformer-centric search methods.

\subsubsection{Reinforcement Learning Search (RL-NAS)}

The pioneering NAS method based on RL \cite{zoph2016neural} consists of an RNN model as a controller which interacts with the environment of all possible neural architectures. In each search iteration, the controller predicts the best architecture that is likely to produce good model performance, i.e., accuracy, and the predicted model is trained end-to-end. 
UniNet \cite{liu2021uninet} is an RL-based method to search for the optimal combination of convolution, MHA and MLP-mixer layer \cite{tolstikhin2021mlp}, along with their depth and channel dimension throughout the transformer backbone network. As-ViT \cite{chen2022auto} utilizes the RL-NAS method to find the base ViT architecture, which is further scaled up to meet the accuracy and computation budget requirement. Compiler-aware architecture search \cite{niu2020real} is an RL-NAS method to search for a BERT model network that achieves good accuracy and has low latency on mobile CPU and GPU platforms.

\subsubsection{One-shot/Differentiable Search}

The one-shot search methods reduce the computation time by creating a supernetwork of all possible neural architectures based on a predefined search. This approach uses weight-sharing to reuse the same weights for multiple architecture combinations, allowing the search algorithm to train and evaluate using a single big network instead of individual small models.  
DARTS \cite{liu2018darts} is a one-shot method that uses a learnable architectural parameter for each operation in the search space, formulating the search process in a differentiable manner. The final searched architecture is obtained by sampling the best operation at each level of the network. DARTSformer \cite{zhao2021memory} addresses the problem of applying DARTS directly on the transformer search space. They show that the memory consumption of the transformer supernetwork increases with hidden size. The authors combine DARTS with \enquote{reversible networks} \cite{gomez2017reversible} to search without running out of memory. This is achieved by reconstructing the input of a reversible network layer from its output during backpropagation, requiring only the output of the last layer to be stored. This reduces the memory burden on the supernetwork and allows for higher hidden sizes and more candidate choices. The searched network for machine translation performs better than the vanilla transformer.

Planer \cite{latifi2022efficient} is a differentiable search method \cite{wu2019fbnet}, which takes a transformer model and a target latency value to produce a sparsely-activated optimized network that meets the latency budget. The search space includes different combinations of FFN layer, number of heads and mixture-of-expert (MoE) layers. MoE layers consist of multiple expert layers, where each path predicts a specific subset of inputs. The output prediction accuracy is enhanced by combining the outputs from all the paths. The transformer models searched using Planer achieve 2$\times$ speedup over vanilla transformer on a GPU while maintaining the baseline accuracy.

You Only Compress Once BERT (YOCO-BERT) \cite{zhang2021you} first constructs a search space of 10$^{13}$ architectures, featuring all combinations of a BERT model. The optimal model for a given performance constraint is then searched using a novel stochastic nature gradient optimization method. ViT-Slim \cite{chavan2022vision} is another one-shot framework to search for an efficient architecture over three important modules - MHA, FFN and patching mechanism.  

\subsubsection{Evolutionary Learning Search} 

The evolutionary learning search algorithms \cite{chen2019renas} use the principle of natural evolution, such as selection and mutation, to find optimal neural architectures. The genetic algorithm (GA) in evolutionary learning is an iterative process of evaluating selected individuals according to a fitness function and generating a new set of architectures using the characteristics of best-performing models from the previous generation. Initially, a population is randomly generated by sampling different architectures from a large pool of networks in the search space. Each individual is a specific neural architecture trained on the target task to determine fitness. The weaker networks have less chance of surviving in the current generation as they compete with candidates of a higher fitness function. The next generation of top-k networks is obtained by mutation or crossover of top individual models in the current generation of networks. Although this search methodology is very effective, it requires a large amount of computation time and resources.

Evolved Transformer (ET) \cite{so2019evolved} and Primer \cite{so2021searching} are examples of evolutionary learning methods to find optimal encoder-decoder and decoder-only self-attention networks, respectively. ET allocates more computing resources to promising models and finds efficient cell/graph architecture stacked multiple times to form the encoder and decoder units. Primer resolves the hurdles of the search method in Evolved Transformer and uses a small proxy dataset to search for optimal models, which are then transferred to the large target dataset. The searched ET and primer models achieve superior validation metrics than the vanilla transformers on language tasks.

Real-time Style Transfer \cite{benmeziane2022real} is a hardware-aware CV transformer search method. The submodules of the ViT backbone for style transfer are searched using evolutionary search. The searched network is at least 2.1$\times$ faster than the baseline model for style transfer on Xiaomi Redmi 10 mobile and Raspberry Pi 3 embedded devices.

\subsubsection{Once-for-all search}

Once-For-All (OFA) \cite{cai2019once} is a two-step method which combines the one-shot approach and evolutionary learning search process. The OFA method first trains a supernetwork of maximum dimensions along all the hyperparameters, i.e., MHA heads, FNN dimension, kernel and filter size. During the second step of the evolutionary search process, the predicted models are sampled from the pretrained supernetwork and validated to obtain validation metrics without additional finetuning. This search method has the advantage of avoiding training of every sampled network as weights for the sampled models are retained from the trained supernetwork. Several transformer methods rely on the OFA technique by training a large transformer supernetwork and applying evolutionary search for the optimal number of heads and FFN dimension.

HAT \cite{wang2020hat} is an HW-NAS method, where the search space includes key transformer hyperparameters, namely, the number of heads, MLP dimension, and the number of encoder/decoder blocks. The search space is elastic in such a way that the encoder/decoder module at each stage can choose a different set of hyperparameters. The search methodology follows the once-for-all technique, where a supernetwork of the highest dimension is initially trained, and different submodels are sampled to perform the evolutionary search. The search process incorporates the target latency for different devices, viz., Intel Xeon CPU, Raspberry Pi ARM CPU, and Nvidia TITAN Xp GPU platforms. The searched model for Raspberry Pi-4 embedded device on the WMT’14 translation task achieves a 3$\times$ speedup with 3.7$\times$ fewer trainable parameters over the baseline vanilla transformer. The searched subtransformers reveal that GPU platforms prefer shallow model with wide layers, while ARM CPU picks deep model with thin layers to obtain optimal hardware performance.

Several vision transformer search methods, such as AutoFormer \cite{chen2021autoformer}, ViT-ResNAS \cite{liao2021searching}, NASformer \cite{ni2022nasformer}, BurgerFormer \cite{yang2022searching}, etc., also rely on OFA technique to search for an optimal model. The OFA method first trains a large Supernetwork of maximum dimensions and samples different-sized models from the supernetwork to specialize for different performance metrics, thereby significantly reducing the search cost. The supernetwork comprises maximum dimensions, such as QKV, number of heads, embedding dimensions etc. The objective function of evolutionary search considers different performance related metrics in different ViT methods, such as accuracy, model size and latency. The searched CV transformer models outperform SOTA CNN and manually designed self-attention-based models, such as DeiT \cite{touvron2021training}, in terms of accuracy and number of parameters.

\subsubsection{NAS using Knowledge Distillation}

Knowledge distillation methodology can accelerate the NAS process by transferring the acquired knowledge  from a large teacher network to the student model. This allows the NAS algorithm to evaluate the performance of the predicted subnetwork using the teacher model. LightHuBERT \cite{wang2022lighthubert} first trains a once-for-all BERT supernetwork of maximum dimensions from scratch using the loss function of the pre-training distillation. During the search process, the subnetworks with different sizes of embedding dimension, number of heads, FFN inner dimension and network depth are sampled and evaluated using the teacher model. The search process culminates when a network with desired performance is reached. 
On several speech recognition tasks, the searched LightHuBERT achieves comparable predictive performance as the baseline teacher model HuBERT \cite{hsu2021hubert}, even though it has 29\% fewer model parameters.
AutoDistill \cite{zhang2022autodistill} is a distillation-based search method that utilizes multi-objective Bayesian Optimization (BO)  to learn a small model by considering several objectives, constraints and hardware performance. The authors include layer-wise, progressive knowledge transfer, and a model pre-training distillation into the search process.

\subsubsection{Accuracy predictor based NAS}

The search methods based on accuracy predictors utilize an ML model as a surrogate model to predict the accuracy of a predicted network during the search process. The auxiliary model is previously trained on pre-collected samples of architecture-accuracy pairs. 
LightSpeech \cite{luo2021lightspeech} is a search method to find the optimal text-to-speech model based on the accuracy predictor. The search space is a hybrid attention-convolution backbone consisting of the number of heads and kernel size in the convolution operation. The searched transformer is 6.5$\times$ faster than the baseline model on Xeon CPU E5-2690 v4 with similar voice quality metrics.

\subsubsection{Training-free Neural Architecture Search}

The training-free NAS methods rely on a set of performance evaluation strategies based on the model architecture or gradient information for quickly estimating the model accuracy. During the search iteration, the estimated accuracy is used without training the model, thereby significantly reducing the search time. For instance, Abdelfattah et al. \cite{abdelfattah2021zero} devise a zero-cost proxy method, where they assign a score to the neural architecture at initialization which is indicative of the validation accuracy of just a single minibatch of data. LiteTransformerSearch  \cite{javaheripi2022litetransformersearch} is a training-free search method to find optimal transformer architectures for resource-constrained hardware platforms. The authors establish a strong relationship between the validation accuracy and the number of model parameters of the decoder layer in the transformer, thereby substituting the decoder parameter count as a proxy during the search process. The authors integrated the zero-cost proxy metric into the evolutionary search algorithm, where the accuracy is obtained by the surrogate model and latency is computed from the target hardware platform. 
The searched transformers attain a speedup of 1.3$\times$ and 1.5$\times$ on Intel Core i7 CPUs and Nvidia Titan Xp GPUs, respectively, over the baseline transformers while achieving similar perplexity. 
TF-TAS \cite{zhou2022training} is a zero-cost proxy method, which evaluates different configurations of the CV transformer model at a low cost. The evaluation is based on the synaptic diversity and synaptic saliency of MHA and MLP units, respectively. The search time to find the optimal transformer is just 0.5 days, while the supernetwork training-based search requires 24 GPU days.

\subsection{Application of NAS for Model Compression}

The algorithmic development of NAS methods led to applying the search strategies to automatically solve combinatorial  problems. In this subsection, we review use of NAS  methods for model compression, mixed-precision quantization and other use cases.

\subsubsection{NAS for Automated Pruning}

The NAS-based pruning methods apply the search methodology to automatically remove the redundant parameters, providing an alternative to manual pruning algorithms. Several BERT-centric pruning methods, such as AdaBERT \cite{chen2020adabert} and NAS-BERT \cite{xu2021bert}, compress the large model into a small model on downstream tasks. AdaBERT \cite{chen2020adabert}, LightHuBERT \cite{wang2022lighthubert}, AE-BERT \cite{huang2022automatic} are examples of BERT compression methods using NAS principles. NAS-BERT \cite{xu2021bert} is a task-independent compression strategy to reduce the size of a BERT model, while AdaBERT \cite{chen2020adabert} compresses a pretrained BERT in a task-dependent manner utilizing differentiable search \cite{liu2018darts}. AE-BERT \cite{huang2022automatic} is an automated compression methodology to find an submodel for a target pruning ratio from a pretrained BERT.
 
\subsubsection{NAS for Mixed-Precision Quantization Search}

The challenge in mixed-precision quantization is assigning the optimal bit-width of each layer such that model size is reduced without accuracy loss. If an \enquote{L}-layer transformer can be quantized to one of the \enquote{b} possible bitwidth values, there exists b$^{L}$ different quantized configurations. It is practically impossible to finetune every combination to find the optimal mixed-precision quantized model. Therefore, the problem of mixed-precision quantization can be reformulated as a NAS problem, thereby utilizing the search principles of NAS algorithms. AQ-BERT \cite{zhao2021automatic} is a mixed-precision quantization search method to assign different bit-width/precision to different encoder layers of a pretrained BERT model and different precisions to different sub-groups within a layer. The automatically searched and compressed BERT networks show their effectiveness on standard GLUE benchmarks and commodity hardware in terms of accuracy and latency.
 
\subsubsection{NAS for Hybrid Operator Search}

NAS methods can also be utilized to search for hybrid operators within a transformer backbone network. We review two case studies to demonstrate the usefulness of NAS in such scenarios. 
Liu et al. \cite{liu2022neural} include the conventional attention ($\mathcal{O}(n^{2})$ complexity) and linear attention ($\mathcal{O}(n)$ complexity) from Cosformer \cite{qin2022cosformer} in the search space. They propose a search methodology to find the best type of attention at each layer of the transformer network such that the models are balanced in terms of time complexity and accuracy. Compared to baseline transformers that have quadratic complexity, their searched networks have comparable accuracy on NLP and vision tasks, while having much better compute efficiency.

The networks based on convolution and attention operations achieve high accuracy; however, involve multiplications, which are expensive. To boost hardware efficiency, researchers have proposed networks that perform only addition \cite{chen2020addernet} or bitwise-shift operations \cite{you2020shiftaddnet}. Nevertheless, such non-multiplication networks attain inferior accuracy. ShiftAddNAS \cite{you2022shiftaddnas} employs a NAS method which searches the combination of multiplication or non-multiplication operators at every layer of the backbone transformer network to balance model accuracy and hardware efficiency.The searched transformers on WMT’14 En-Fr and WMT’14 En-De NLP datasets outperform the baseline transformers and HAT models in terms of latency, energy, and BLEU score. The searched CV transformer model also outperforms the ResNet50 and other CV transformers on ImageNet dataset.

\begin{figure}[H]
    \centering
    \includegraphics[scale=0.38]{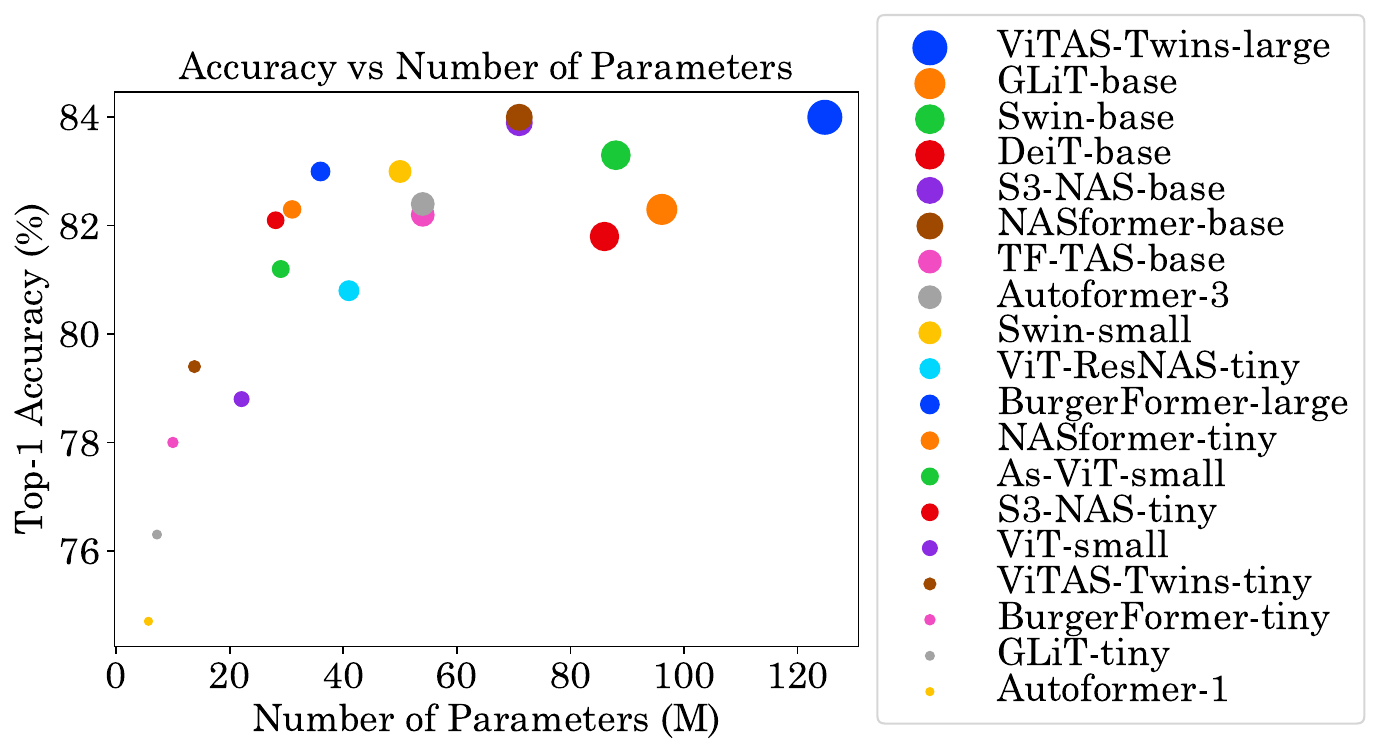}
    \captionsetup{justification=centering}
    \caption{ImageNet accuracy vs model size comparison of the searched CV transformer models with manually designed models. The searched CV transformers include:  
ViTAS \cite{su2021vitas}, GLiT \cite{chen2021glit}, S3-NAS \cite{chen2021searching}, TF-TAS \cite{zhou2022training}, NASformer \cite{ni2022nasformer}, AutoFormer \cite{chen2021autoformer}, ViT-ResNAS \cite{liao2021searching}, BurgerFormer \cite{yang2022searching}, As-ViT \cite{chen2022auto} } 
    \label{fig:ViT_NAS_plot}
\end{figure}

 \subsection{Quantitative comparison of searched CV transformers}

In Figure \ref{fig:ViT_NAS_plot}, we summarize the parameter count and top-1 accuracy on the ImageNet dataset of the searched models. Based on the plot, it appears that a model must possess a large number of parameters in order to achieve a high level of accuracy. However, the trend is not linear, rather increases with increasing parameter count and saturates after a point. Methods such as NASformer \cite{ni2022nasformer} have better accuracy and fewer parameters than manually designed models such as Swin \cite{liu2021swin} and DeiT \cite{touvron2021training}. This highlights the strengths of these search methods.

%Designing efficient transformer architectures manually
%Designing efficient transformer architectures using NAS 

\section{Hardware optimization techniques} \label{sec:HardwareOptimization}
\begin{table}[htbp]
  \centering \footnotesize
  \caption{Classification of hardware-level techniques}
    \begin{tabular}{|p{1.5cm}|p{6.5cm}|}
    \hline
Reducing computations & by skipping layers or input/output channels  \cite{sreedhar2022enabling}, exploiting patch locality \cite{li2022divit}, performing two multiplications in one go  \cite{zhang2021algorithm,wang202228nm}, early terminating negative computations  \cite{wang202228nm}, avoiding trivial operations \cite{ma2022axby}
\\\hline
Predicting attention score using & LRT of Q and K  \cite{qu2022dota}, cosine of Hamming distance between two vectors \cite{ham2021elsa}, accounting for only the largest positive and negative products \cite{ham20203}, ternary (0/-1/1) quantization of key matrix \cite{chen2022enabling} \\\hline
Loop-optimizations & unrolling \cite{qi2021accommodating}, reordering \cite{qi2021accommodating}, fusion \cite{peng2022length} \\\hline
Systolic array &  \cite{ye2022accelerating,lu2020hardware,li2022divit,shen2022salo,fang2022efficient,lu2021sanger} \\\hline
What is stationary in dataflow & output-block \cite{zhao2022fpga}, output \cite{shen2022salo}, weight \cite{shen2022salo,li2022divit}, key \cite{you2022vitcod}, sparse attention scores \cite{lu2021sanger} \\\hline

LUT for & log \cite{li2022accelerating}, multiplications \cite{li2022accelerating,peng2022length},  reciprocal \cite{ham2021elsa}, exponent \cite{fang2022efficient,li2022divit}, softmax \cite{wang202228nm}, cosine values \cite{ham2021elsa}, adder-tree \cite{zhao2022fpga}, inverse square root \cite{lu2020hardware}, GeLU \cite{hong2022dfx} \\\hline
Approximate computing & computing multiplication of large values exactly and small values approximately  \cite{wang202228nm}, truncating mantissa bits \cite{ma2022axby} \\\hline
GD of weights  & exploited for quantization \cite{zadeh2020gobo,zadeh2022mokey} and trivialization \cite{ma2022axby} \\\hline
Intelligent data storage & storing encoder/decoder layers on-chip and embedding layer off-chip \cite{li2020ftrans}, storing weight matrices in HBM and remaining data in DDR \cite{hong2022dfx} \\\hline
Memory optimizations & matrix reordering to expose reuse in dilated window attention \cite{shen2022salo}, splitting/compressing weight matrices to make them SA-friendly \cite{lu2020hardware} or FPGA buffer-friendly \cite{zhang2021algorithm},  custom data-layout to remove bank-conflicts \cite{fan2022adaptable},   formats for storing sparse matrices \cite{qi2021accommodating,peng2021accelerating} \\\hline
Others & FFT  \cite{li2020ftrans,fan2022adaptable,rizk2022resource,fan2022adaptable}, tiling \cite{hong2022dfx,you2022vitcod,shen2022salo}, early-exit \cite{li2022accelerating}, block-circulant matrix \cite{li2020ftrans,rizk2022resource}, double-buffering \cite{li2022divit,fan2022adaptable,li2022slice}, load-balancing \cite{ye2022accelerating,lu2021sanger,hong2022dfx,qu2022dota}, multi-FPGA accelerator \cite{hong2022dfx}, agglomerative clustering \cite{zadeh2022mokey} \\\hline
BERT &  \cite{chang2022pipebert,ham2021elsa,lu2021sanger,chen2022enabling} \\\hline
NLP & \cite{li2020ftrans,lu2020hardware,li2022accelerating,nagarajan2022axformer,wang2021spatten,shen2022salo,qu2022dota,peng2022length,wang202228nm,yang2022dtatrans,zadeh2020gobo,huang2021hmc,zhang2021algorithm,fan2022adaptable,peng2021accelerating,zadeh2022mokey,fang2022efficient,wang202228nm,ham20203} \\\hline
CV & \cite{zhao2022fpga,sreedhar2022enabling,li2022divit,you2022vitcod,ma2022axby,shen2022salo,qu2022dota} \\\hline

 \end{tabular}%
  \label{tab:Hardwaretable}%
\end{table}

Hardware optimization techniques for transformers play a crucial role in achieving efficient and high-performance computing. These techniques involve improving the design and architecture of hardware systems for efficient inference of the neural network models to optimize performance and energy efficiency. Table \ref{tab:Hardwaretable} shows key ideas of these techniques. In this section, we discuss hardware optimization techniques, such as pipelining (Section \ref{sec:pipelining}), optimizing matrix-multiplication operations (Section \ref{sec:optmatmul}) and skipping redundant or trivial operations (Section \ref{sec:redundantTrivial}). We also review dataflows to exploit reuse (Section \ref{sec:dataflows}), and  block-circulant matrix to compress weight storage (Section \ref{sec:blockCirculant}).

\subsection{Pipelining}\label{sec:pipelining}

Pipelining allows overlapping computation with data transfer or overlapping different sub-computations. It helps achieve load-balancing and is especially helpful for deep networks such as transformers, which have multiple encoder and decoder layers.

PipeBERT  \cite{chang2022pipebert} is a pipelining technique to accelerate BERT models on big.LITTLE processors. This processor has two clusters, big and LITTLE, each with four cores. They divide the BERT network into two subgraphs and map one subgraph each on big and LITTLE clusters. Mapping is realized using the \enquote{affinity} functionality of CPU. They propose a latency-aware binary search algorithm to achieve a load-balanced division of the subgraph. They first observe the ratio, say R, of the throughput of BERT on big and LITTLE clusters. Let the total number of operations in the entire graph be Z. They start with the initial allocation of [Z-Z/R, Z/R] operations to [big, LITTLE] clusters. Then, the latencies of both clusters are recorded. If they differ by more than a threshold, then half of the operations from the slower cluster are moved to the lighter cluster. This process is repeated till the latency difference falls below the threshold. Their binary search algorithm completes much faster than naive binary search and brute-force search. The subgraphs on two clusters operate in a pipelined fashion. On the HiKey970 system, their technique achieves much higher throughput than executing on four big cores and a much lower \enquote{energy-delay product} than the best single-cluster execution.

\textbf{Length Adpative Co-design:} Peng et al. \cite{peng2022length} note that when the sequence lengths of the inputs in a batch vary greatly, e.g., in SQuAD v2.0, the maximum and average lengths are 975 and 171, respectively. Padding of inputs creates unnecessary overheads. They predict relative attention scores at low-precision. 
Multiplication is realized through an LUT, e.g., with 4b integers, only 256 entry LUT is required. Then, $Q'K'^T$ is computed. After this, sorting is done to choose top-$L$ (i.e., most dominant) attention scores. For these $L$ values, exact MatMul and softmax are done to obtain $QK^T$. This is more efficient than computing the attention scores of all the elements. They split the encoder into three coarse-grain pipelines: (1) linear layer using MatMul and approximate attention computation. (2) attention computation and (3) FFN. Stages 1 and 3 utilize on-chip memory and loop allocation to overcome the memory wall. Stage 2 is further subdivided into a three-stage pipeline for trading resource utilization with data locality. Various attention operators are fused into a single loop using the fine-grain pipelining capability of FPGA. The reconfigurability of FPGA allows fusing loops with diverse iteration counts, whereas GPU can only fuse selected loops. 

Variable input sequence lengths lead to stalling of the pipeline. They propose a sequence-length aware coarse-grain pipeline, which adjusts the resource allocation according to the requirement of a stage. It works based on the observation that with sparse attention, all operators have linear complexity. It sorts the inputs of a batch in decreasing order of length. Transformers usually have multiple encoder layers, and the inputs sequentially go through those layers. Their technique patches the pipeline stages of different sequence length inputs and different encoder layers. This removes pipeline bubbles and improves hardware efficiency. Pipelining and data-prefetching facilitate the overlapping of compute with data transfer. Their technique leads to better throughput than padding and cutting schemes. Their technique provides higher throughput for a marginal accuracy loss than CPU and GPU implementation and higher energy efficiency than a GPU design using CUBLAS GEMM.

\subsection{Optimizing matrix-multiplication}\label{sec:optmatmul}

Transformer computations involve the multiplication of large matrices, and hence, optimizing matrix multiplication can significantly boost compute efficiency.

\begin{figure} [htbp]\centering
    \includegraphics[scale=0.33]{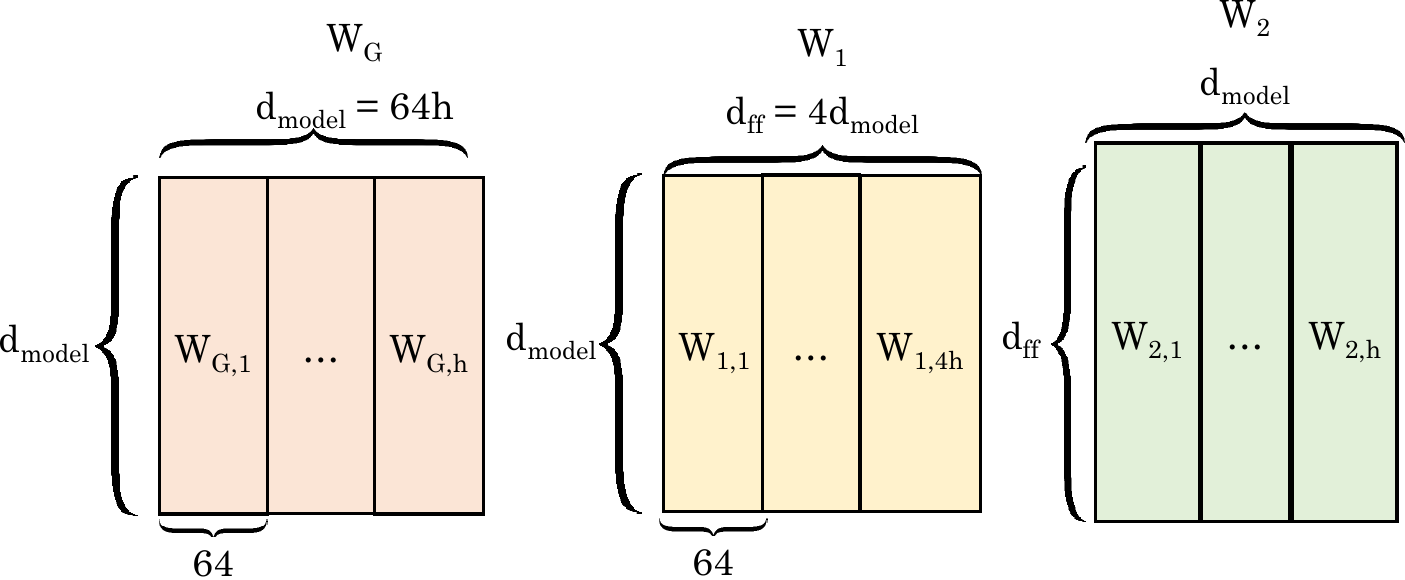}       
    \caption{Division of weight matrices in the technique of Lu et al. \cite{lu2020hardware} }
    \label{fig:SOCC_HardwareAccelerator}
\end{figure}

  Lu et al. \cite{lu2020hardware} present a hardware accelerator for MHA and FFN blocks in the transformer. The computation of FFN is shown as $FFN(x) = ReLU(xW_1 + b_1)W_2 + b_2$.
They note that all four tensors have a shape of $[batchSize, s, d_{model}]$. Both MHA and FFN blocks involve GEMM computations. In all the heads, the GEMM computations of linear sublayers can be done by an systolic array (SA) of size $s\times 64$. However, the use of this SA is still a challenge for large matrices such as $W_1$, $W_2$ and $W_G$. They note that in well-known transformers, $d_{model} = 64h$, and $d_{ff} = 4d_{model}$. Based on this, they divide the weight matrices $W_1$, $W_2$ and $W_G$ in a manner shown in Figure \ref{fig:SOCC_HardwareAccelerator}. Then, most GEMMs can be performed using an SA of size $s\times 64$, where $s$ denotes the max sequence length. The use of a single SA saves hardware resources.
 
The $s \times 64$ SA has $s$ rows and 64 columns, which produces the output matrix column by column. Then, the bias and the residual input are added using two sets of $s$ adders. The softmax calculations are done in parallel to the calculations of \enquote{value} to enhance the hardware usage of this SA. They ensure that the softmax produces the output before the SA computes the \enquote{value} so that the overall latency is dictated by the SA and layerNorm block. In the softmax block, they avoid LUT and multipliers by using the 
\enquote{log-sum-exp} strategy, which uses linear approximation for logarithmic and exponentiation functions. Both MHA and FFN blocks need to compute layerNorm; hence, layerNorm is on the critical path. They minimize its latency by starting their computation early. They implement their technique on FPGA and show that it provides high speedup compared to the GPU.

\textbf{DFX:} Hong et al. \cite{hong2022dfx} present a multi-FPGA accelerator for text-generation workloads such as GPT. The GPT-2 model produces tokens in two stages: summarization and generation. These stages have different properties. The parallel compute capability of GPU makes it effective for summarization, however, the sequential nature of generation renders GPU ineffective. Acceleration of GPT-2 requires accelerating not only attention and FFN but LayerNorm, LM head, residual and token embedding. While residual and LayerNorm account for only 0.1\% of operations, they account for 22.8\% of the GPT-2 latency. This highlights the need for a custom accelerator. Further, given the massive size of GPT-2, a single FPGA device is insufficient, and hence, they connect a cluster of four FPGAs with a single CPU host.

In pipelining, all the workers process one input completely, leading to large latency. Instead, they use intra-layer model parallelism, which reduces MatMul latency and incurs low synchronization delay. Every FPGA (worker) handles weight matrices at head granularity in MHA and column granularity in FC layers. Unlike GPU cores, which depend on massive parallelism, their compute core is optimized for processing individual tokens. Their ISA includes (1) matrix instructions for performing matrix-vector operations, (2) vector instructions for performing vector-vector and vector-scalar operations. The vector instructions execute LayerNorm and softmax, (3) DMA instructions and (4) router instructions. An FPGA has 8GB HBM and 32GB DDR with bandwidths of 460GB/s and 38GB/s. The weight matrices are kept in HBM since they are frequently accessed. The remaining infrequently accessed data are kept in DDR. 

In conventional attention operation, the key is transposed, however, they transpose `value' since it is read column-wise but written row-wise. Thus, transpose is performed inherently during the write operation, not the read operation. To hide the latency of the transpose operation, they start it in advance while Q and K are being computed. Both matrix and vector operations are mapped to DSPs, whereas non-linear operations are performed using DSP, LUT and combinational logic. GeLU is realized using piecewise linear approximation and an LUT. A parallel tree of comparators is used for finding the largest value of a vector. A router is used to synchronize different FPGAs over a ring network. They evaluate their technique using four Alveo U280 FPGAs and show that it is superior to four V100 GPUs in performance, energy efficiency and cost efficiency.

\subsection{Skipping redundant, ineffectual or trivial computations}\label{sec:redundantTrivial}

A large fraction of computations performed in transformers are either repeated (e.g., due to patch locality), have no impact on the final output (because only high attention scores decide the final result) or are trivial (e.g., multiplication with zero or one). Identifying and avoiding such computations can boost efficiency significantly.

Since softmax is a normalization approach, what matters is only the relative values of attention scores and not the absolute values. Based on this, some works \cite{peng2022length,lu2021sanger,qu2022dota} quantize K and Q from the FP32 value to a low-precision (1b or 4b) integer format. Since quantization and exponential are monotonic functions, the relative ordering of attention scores is maintained. Compared to 16-bit dense attention,  4-bit computations  incur only 1/16X overhead.
A few  works predict attention scores based on low-rank transformations (LRT) of Q and K \cite{qu2022dota}.

ELSA technique \cite{ham2021elsa} reduces computations in attention operation. Without performing full computations, they seek to select, for every query, the key that would lead to high attention scores. They compute approximate self-attention in three steps (1) They compute Hamming distance between  $k$-bit hashes of two vectors (say key and query) to guess the angle between them. The hash function is computed by using a structured orthogonal matrix.  (2) A higher value of the cosine of the angle shows a higher value of dot-product between them and hence, the higher similarity between them. (3) The relevance of a key to a query is ascertained by comparing the approximate similarity to a threshold. Then, irrelevant relations are filtered out, and the exact dot-product is computed only for significant relations. While computing exact dot-product requires $d$ multiplications, their algorithm requires only two multiplications, a cosine function and Hamming distance calculation. An LUT of cosine values can further reduce the computations.  Based on the amount of approximation acceptable to a user ($s$), their technique decides the threshold for each sub-layer. For this, the target model is run with the training set and for a sub-layer, the keys whose softmax-normalized attention score is more than $s/n$ are considered relevant, where $n$ is the count of input elements.  With only 1\% accuracy loss, their proposed accelerator provides high speedup and energy efficiency gain compared to the GPU.

DOTA  \cite{qu2022dota} technique reduces computation wastage. Although 90\% of connections in FC attention layers can be safely pruned, still, the attention scores and softmax need to be computed to find the prominent attention weights.  To avoid this wastage, they detect ineffectual connections based on idea that connections with low attention weights also have a low value of attention score $S=QK^T$. The authors propose a learning approach to identify the relative importance of connections in attention graphs. Their technique predicts attention scores ($\tilde{S}$) based on LRT of Q and K. Weak attentions are identified by comparing $\tilde{S}$ with a threshold. Masking weak attentions reduces the denominator in softmax and affects the quality. To address this, their technique optimizes the parameters of the network and LRT by minimizing the weighted sum of network loss and attention detection loss. At the end of the training, LRT parameters are trained to detect stronger connections and the network parameters are trained to work with the sparse attention graph and still achieve high accuracy. The rank of $\tilde{S}$ depends on the target task complexity.

Long-sequence transformers operate on large GEMM/GEMV computations with configurable hidden dimensions. Hence, they identify the building blocks of each computation and propose a single architecture for efficiently handling various layers. They propose a reconfigurable MM engine (RMME) to support multi-precision calculations at row-granularity. RMME is a 32*16 array of PEs that can support 16b fixed-point, INT2, INT4 and INT8 calculations. RMME initially estimates $\tilde{S}$  in INT2 or INT4 precision and then selects significant attentions. For the sake of hardware-friendliness, it selects an equal number of attention connections in all the rows of the attention matrix.  Then, only significant attentions are computed in 16b fixed-point precision. The $QK^T$ results are converted to FP before computing the softmax function to avoid overflow. The softmax output is again quantized to perform $A*V$ in fixed-point. Their technique provides large gains over a GPU.

The A$^{3}$ technique \cite{ham20203} proposes two approximation schemes for accelerating attention computations. (1) Since key and value matrices are acquired at the time of knowledge understanding and not query response, they preprocess the key matrix to reduce the number of operations required and query response time. Specifically, for a $d$-dimensional query, the dot-product between a query and a key matrix row is a sum of multiplications of each dimension. Their key idea is that if the multiplication of a single dimension is a large positive number, the final dot-product result is also expected to be large. If this value is a large negative number, the final result is unlikely to be a large positive number.

\begin{figure} [htbp]\centering
    \includegraphics[scale=0.43]{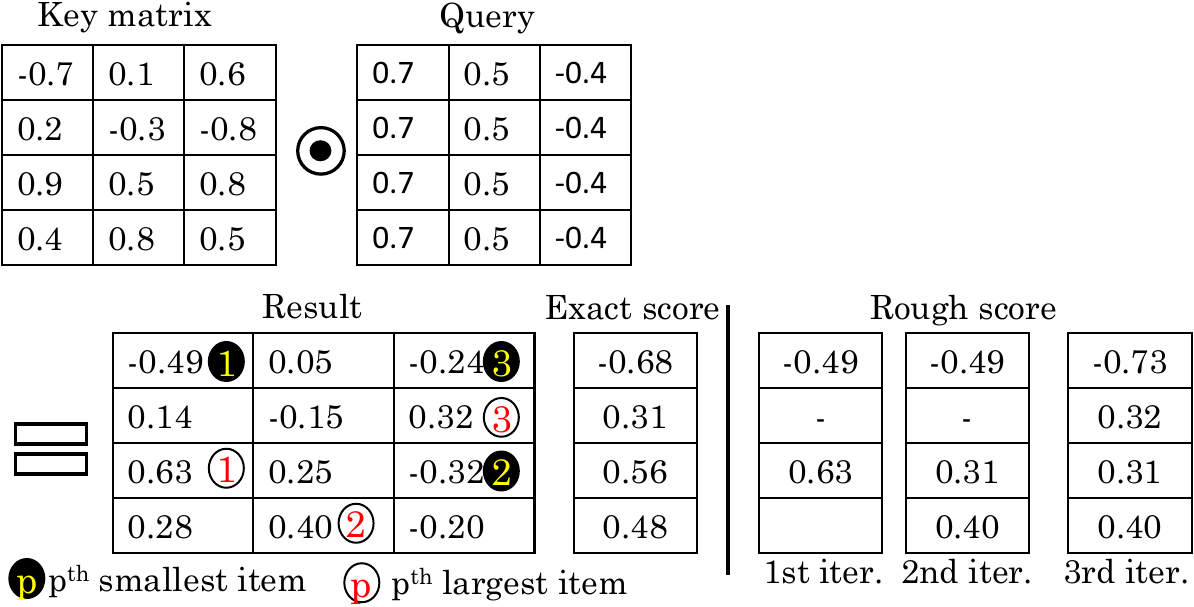}       
    \caption{Approximate computation of attention score ($Z=3$) in the technique of Ham et al. \cite{ham20203}}
    \label{fig:ham2023}
\end{figure}

Their technique seeks to estimate a \enquote{rough-score} for each row in a greedy fashion. This is illustrated in Figure \ref{fig:ham2023}.  In $J^{th}$ iteration, their technique checks the $J^{th}$ largest and smallest number in the result (key*query) and adds them to the rough-score of their row. This is repeated for $Z$ iterations. The rows with positive rough-score have large positive elements and vice versa. Then, their technique further processes only the rows having positive rough-score. 
They sort the key matrix during knowledge comprehension to reduce critical path delay. Then, when the query arrives, the above technique is applied. The preprocessing overhead is easily amortised in networks where many queries use the same key matrix (e.g., 320 in BERT).

(2) They compute the attention score of the above-selected rows in an exact manner and further note that the softmax function transforms score values into weights, such that small scores lead to small weights. These small weights do not impact the model validation accuracy. In fact, they arise because of using a differentiable version of the argmax function, which is required in training, not inference. Hence, these near-zero weights can be treated as zeros. Based on this idea, if a row's score is within $T$\% of the score of the top row, only then it is passed to the softmax function. They quantize FP inputs to FX and use different bitwidths in different layers. They also propose an accelerator for their technique, which achieves magnitude-order gains in performance and energy efficiency over CPU and GPU.

\begin{figure} [htbp]\centering
    \includegraphics[scale=0.36]{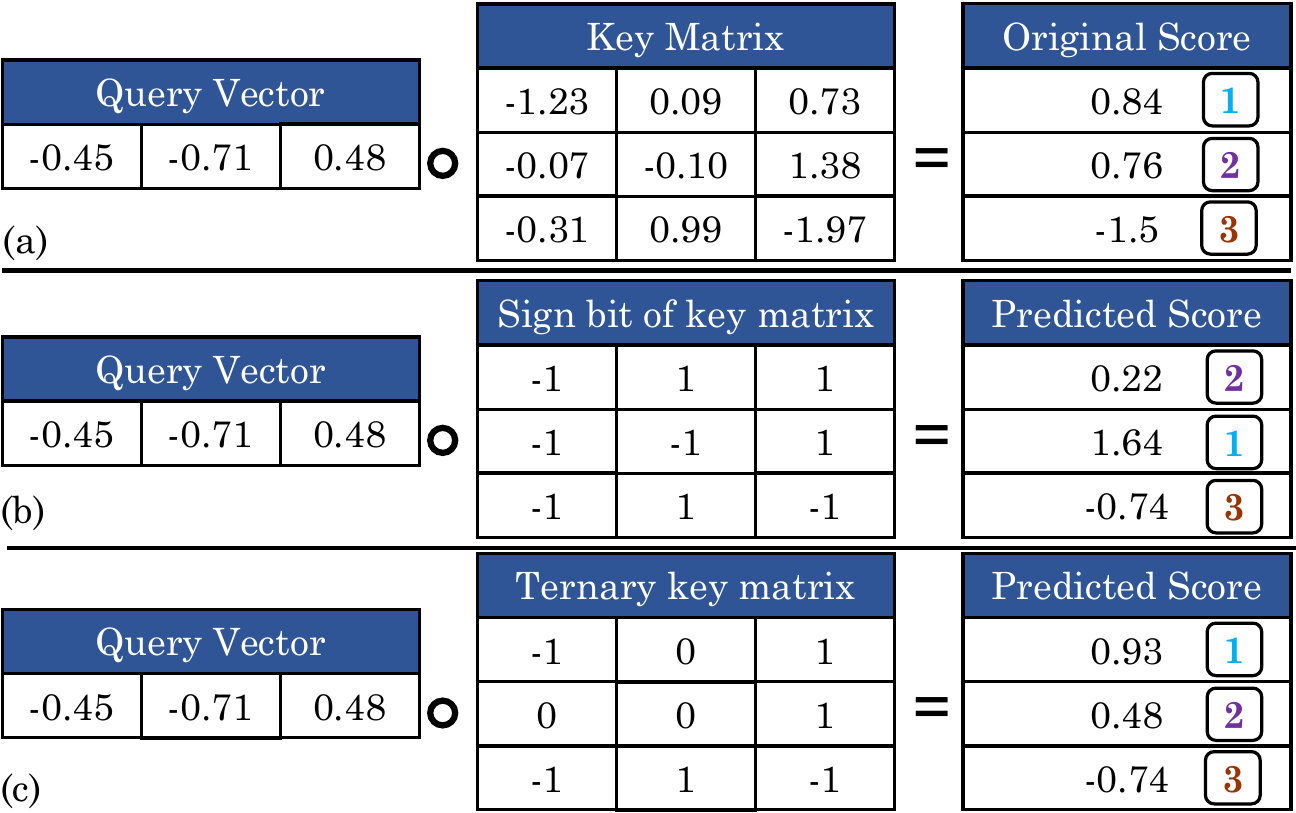}       
    \caption{Ternary pruning \cite{chen2022enabling}}
    \label{fig:TernaryPruning}
\end{figure}

Chen et al. \cite{chen2022enabling} present modifications to the A$^{3}$ technique\cite{ham20203}. They store a key' matrix, such that near-zero/positive/negative numbers in key matrix becomes 0/1/-1 in the key' matrix. On multiplying key' with the query, the relative ordering of attention scores does not change, as illustrated in Figure \ref{fig:TernaryPruning}. Their technique allows finding $Y$ rows with positive rough-score. Further, while Ham et al. \cite{ham20203} perform exact dot-product for these rows and then further shortlist top-$K$ rows, Chen et al. \cite{chen2022enabling} select top-$K$ rows from the $Y$ rows selected based on their rough-scores. Their technique causes a 1\% accuracy loss on the BERT model while boosting energy efficiency.

Wang et al. \cite{wang202228nm} present three approximate-computing techniques for accelerating transformer computations. (1) To avoid the ineffectual computations due to weakly-related tokens, they present a PE that calculates small values with high errors for saving energy but computes large values exactly since these strongly-related tokens have a high impact on the final attention score. It uses an inexact multiplier that gauges the magnitude from the MSB and then, for small values, gates the LSB computations. Although this multiplier produces a high error, the softmax operation reduces the magnitude of those errors.

(2) In $Q*K^T$ computation, near-zero values from softmax become zero after $N$-bit quantization. This sparsity pattern depends on $X_{max}$, which varies for every row. To deal with this issue, they present a \enquote{bidirectional asymptotic speculation engine}. It uses diagonal-prior computing to find row-wise $X_{max}$. It first performs positive MACs, and on observing sparsity, terminates negative computations. A token is strongly related to its neighboring tokens. Based on this locality property of attention, they first compute eight scores on the matrix diagonal to find an estimate of $X'_{max}$. An update-unit iteratively generates a new estimate of $X_{max}$ and compares it with $X'_{max}$, to finally obtain actual $X_{max}$. 
From this, it predicts sparsity and avoids redundant computations.

(3) For the near-zero probability values, many MSB values are zero. Due to them, in $P*V$, many partial products are zero. To address this issue, they present a scheduler that reorders operands to merge two operations in a single multiplication. This removes zero-valued partial products. They fabricate a die that takes only 6.8mm$^2$ area, 272mW power and achieves 14.28TOS/W at 510MHz. Their processor outperforms GPU and previous accelerators.

\textbf{Dynamic ViT Inference:} Sreedhar et al. \cite{sreedhar2022enabling} note that recent vision transformers \cite{liu2021swin,xie2021segformer,xie2021segmenting,zheng2021rethinking,carion2020end} use CONV backbones and CONV layers in the encoder. They evaluate the impact of selective skipping of computations in CONV and attention layers on the predictive performance of object detection and semantic segmentation applications. In DETR and deformable DETR \cite{carion2020end,zhu2020deformable} object detection models, the transformer accounts for nearly 15\% of execution time, whereas ResNet-50 accounts for the remaining time. Especially at higher batch-size, the latency of ResNet-50 increases further. 
In Segformer and Swin-tiny semantic segmentation models, \ApproxSign70 and \ApproxSign90\% of total computations are in CONV layers. However, CONVs account for only  26\% of latency on GPUs since CONV is implemented efficiently on GPUs. MatMul and softmax operations of attention account for 27\% of latency. Thus, both CONV and MatMul need to be accelerated in the overall semantic segmentation application.

They study the resilience of semantic segmentation models towards computation bypassing techniques. Here, they modify the model execution graph to execute subsets of the original model using (1) skipping a layer (2) reducing the input or output channels of a layer (3) decreasing sampling scale factors in spatial-downscaling attention. The Pareto-optimal sub-models are identified that achieve higher accuracy with lower resource consumption.  
In Segformer-B2, scheme (3) degrades accuracy without reducing latency. Rather, Pareto-optimal points are obtained by changing the number of encoder layers in every stage and the input channels to CONV layers. This technique reduces mIoU (mean intersection over union) by 6 percentage point on the ADE20K dataset while bringing a 17\% reduction in latency. This bypassing approach without retraining can save up to 25\% of execution latency. However, saving 50\% latency requires retraining the model. The Swin-Tiny model does not have similar resilience as Segformer-B2.   
They further propose a dynamic inference technique, which performs computation skipping by using an LUT to store Pareto-optimal configurations and their latency saving. During inference, a configuration is searched from this LUT based on the target and resource constraint. Their technique populates this LUT during inference and avoids the need for pruning+retraining.

\begin{figure} [htbp]\centering
    \includegraphics[scale=0.35]{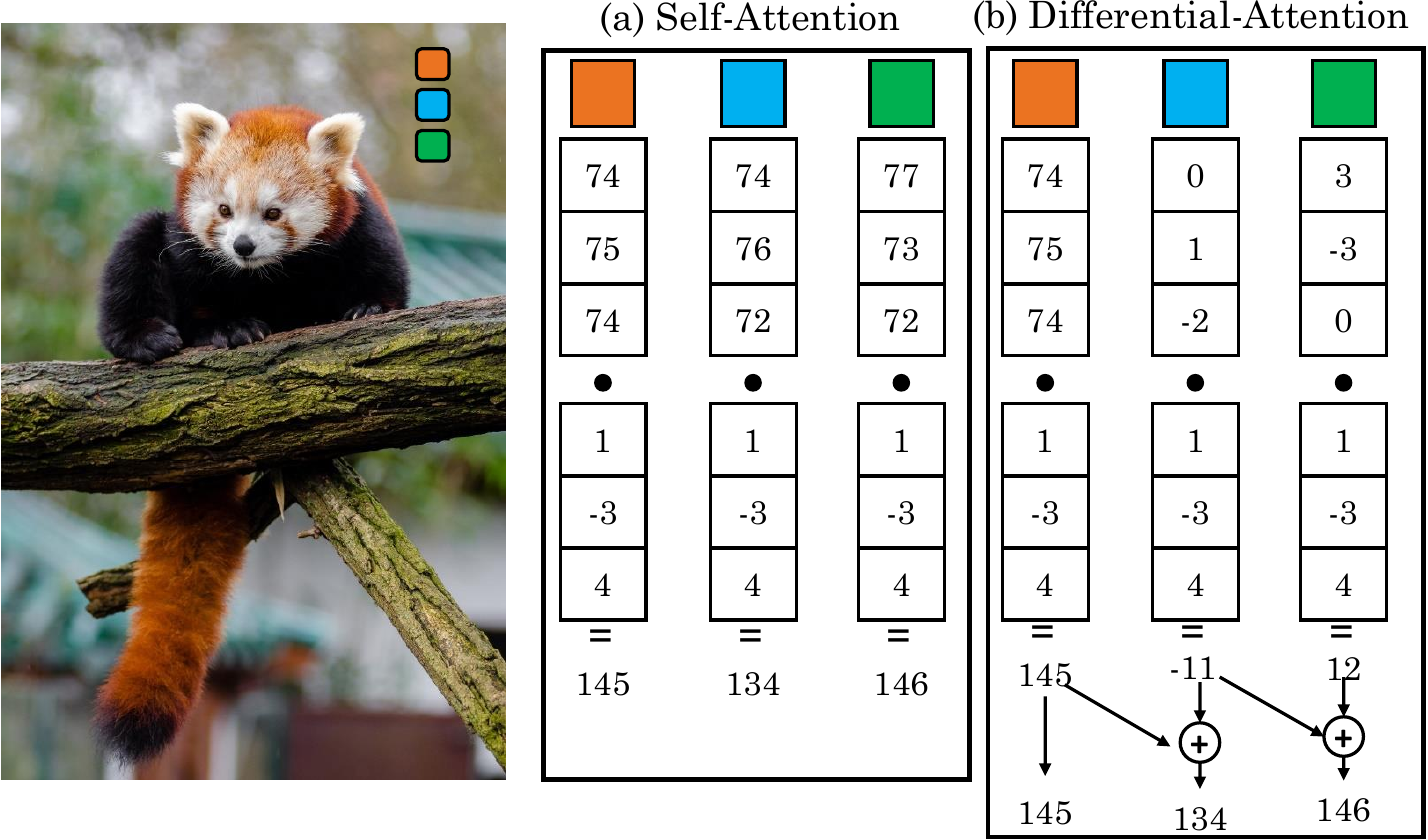}       
    \caption{Delta encoding in technique of Li et al. \cite{li2022divit}  }
    \label{fig:DifferentialAttention}
\end{figure}

DiVIT technique \cite{li2022divit} is based on observation that since neighboring patches of ViT have high locality, many computations in self-attention operation are redundant. It proposes a \enquote{delta encoding} technique to exploit patch locality, which is shown in Figure \ref{fig:DifferentialAttention}.
In the transformer, an input feature x is multiplied with the weight w. If a neighboring patch is $x'$, then $\delta x = x'-x$ shows the delta patch. In a bit-serial multiplier, the actual multiplication workload depends on the number of non-zero bits. With patch locality, $\delta x$ has a lower number of non-zero bits than $x'$. Based on this, delta encoding replaces $x'*w$ with $x*w + \delta x * w$. By reusing the value of x*w from a neighboring patch, their technique can significantly accelerate multiplications in MHA without losing accuracy.

Their proposed accelerator computes differential attention based on the \enquote{bit-pragmatic architecture} \cite{albericio2017bit}, which processes only non-zero bits of input features in a bit-serial manner. Their PE array computes the input patch features in a bit-serial manner. The offset generators change the patch features into powers of two. In every cycle, a weight is multiplied with a power-of-two using a shifter. Then, an add or subtract operation is performed, depending on the sign. Their engine can parallelly compute 32 patches while reusing the same group of weights.

They use a 32*32 PE array, where neighboring patches are assigned to neighboring columns. Thirty-two channels of patch features are assigned to every column. Instead of multipliers, this engine uses shifters and a 32-input adder tree. The PEs in the same row reuse the same group of 32 weights. Specifically, the first column works on the raw values of the patch. After finishing, it passes the result to the second column, which works on the second patch feature in a delta computation manner. This goes on till the last column (number 32), which passes its result back to column 1 for computation on the next set of 32 patches. This pipelining operation hides the latency of accumulating delta results.  The MLP units are handled by an SA. Their accelerator shows high speedup and energy efficiency gains over CPU and GPU.

AxBy-ViT  technique \cite{ma2022axby} avoids trivial operations in ViT. In the operation a*b, if a or b is zero, it is termed fully trivial, whereas if a or b equal $\pm$1, it is termed semi-trivial. They note that the operands in input-embedding, MHA, MLP and FC computations show Gaussian distribution such that more than 50\% operands are concentrated around zero. Also, a small fraction of values in input-embedding and FC are $\pm$1. They note that exactly matching the operands with 0 or $\pm$1 finds the scope of trivialization in only 0.1\% of the operands.

They propose two approximate matching techniques that increase this scope by 1000$\times$. (1) Assuming FP32 operands, values having biased exponents lower than a threshold ($\theta$) are considered zero. For example, if  $\theta$   is 121, then any $|x|$ below 0.015625 is approximated to zero. Say, for 0.01171875, the biased exponent is 120, which is smaller than 121, and hence, it is approximated to zero. This can be seen from the fact that $0.01171875 < 0.015625$. (2) 
To match a number with $\pm1$, they truncate certain mantissa bits and then compare the number. For example, on truncating 19 ($K$) bits,  1.015625 has the same representation as 1.0. However, on truncating only 16 bits, its representation is not similar to that of 1.0. By approximating 1.015625 to 1.0, a computation such as 1.0001*6 reduces to 1*6, which is trivial. On increasing $\theta$ and $K$, the error increases exponentially. On setting $\theta$ to 121, MHA and MLP show an accuracy drop below 3\%, but input-embedding and FC show a larger accuracy drop. Thus, different blocks in ViT have different error tolerance.

\subsection{Dataflows for exploiting reuse}\label{sec:dataflows}
In a hardware accelerator, it is important to choose the right dataflow to exploit reuse and improve hardware utilization. For example, input stationary (IS) and weight stationary (WS) dataflows reduce the energy of reading inputs and weights, respectively, whereas output stationary (OS) reduces the energy of accessing the partial sums.

\begin{figure} [htbp]\centering
    \includegraphics[scale=0.30]{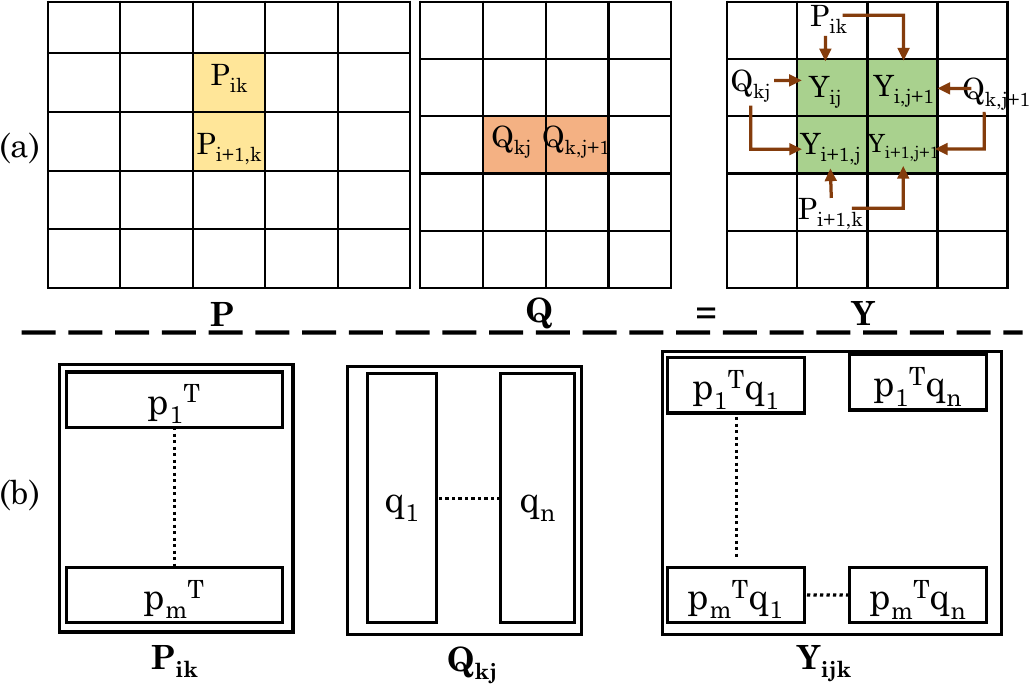}       
    \caption{In OBS dataflow \cite{zhao2022fpga}, broadcasting at (a) block and  (b) vector level   }
    \label{fig:blockStationary}
\end{figure}

\textbf{Output Block Stationary (OBS):} Zhao et al. \cite{zhao2022fpga} note that since CNN kernels show substantial weight-sharing, using WS dataflow is advantageous for CNN accelerators. However, transformer models show much less weight reuse than CNN, rather they show a comparable amount of input and weight sharing. They propose output-block stationary dataflow, which divides the matrix-multiplication operation into several smaller matrix multiplications. To enable reuse, it uses two levels of broadcasting. Consider the matrix-multiplication $P=A\times B$. (1) They divide A and B into $b\times b$ block matrices and compute the resultant block $P_{ij}$ as $\sum_k A_{ik}B_{kj}$. To compute four output block-matrices shown in Figure \ref{fig:blockStationary}, they broadcast four blocks, viz., $A_{ik}$, $A_{(i+1)k}$, $B_{kj}$, and $B_{k(j+1)}$. (2) Inside a block-matrix, they broadcast the rows of $A_{ik}$ and columns of $B_{kj}$, as shown in Figure \ref{fig:blockStationary}.  
In comparison with OS, the two-level broadcasting approach of OBS lowers the memory accesses for input and weight. Further, it reduces the requirement of output writing bandwidth. However, OBS has a slightly lower hardware utilization (96\% vs 89\%).

The partial sums of the output are accumulated in the DSP using an LUT-based adder tree. For a block size of $b$, a block-multiplication unit requires $b^3$ DSPs. At $b=1$, OBS dataflow reduces to OS dataflow. Given the limited number of DSPs, $b$ value can be at most 8. They observe that $b=8$ leads to lower energy consumption than $b=4$, and hence, they choose $b=8$. They experimentally compare WS with OBS for the ``transformer-in-transformer'' (TNT) network. In TNT, an \enquote{outer TE} embeds an \enquote{inner TE} and they extract patch- and pixel-level attention, respectively. The outer TE accounts for more than 83\% of operations, and OBS provides nearly 84\% utilization for this. OBS lowers memory accesses by more than 6.7$\times$ and power consumption by one-third.

\begin{figure} [htbp]\centering
    \includegraphics[scale=0.43]{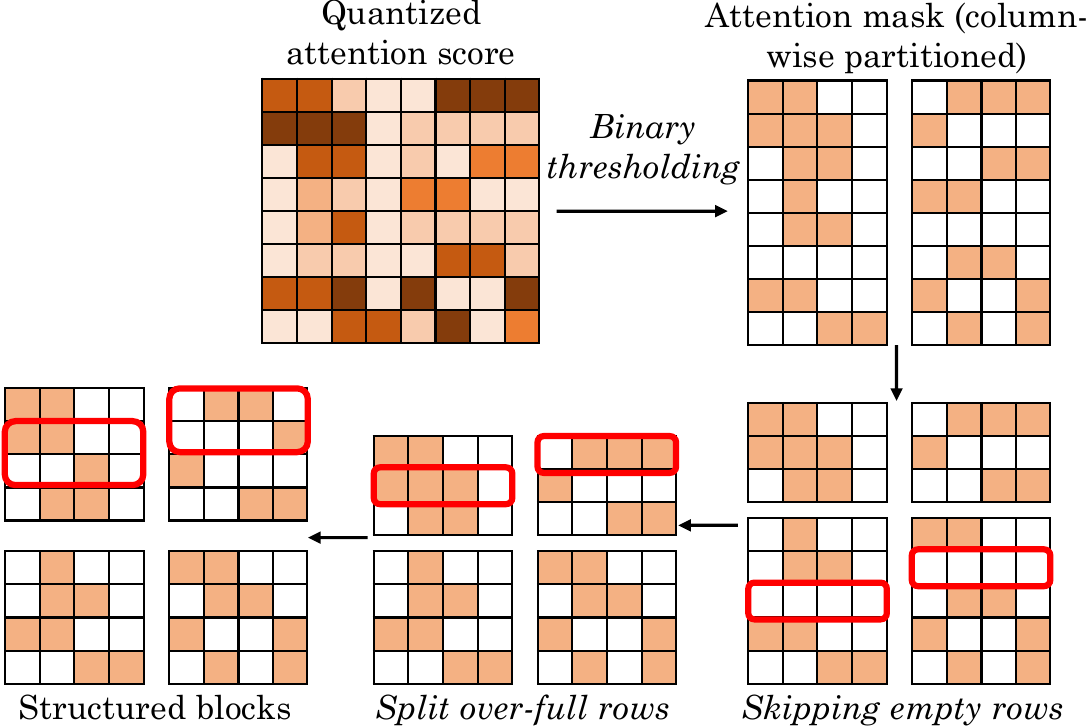}       
    \caption{Encoding of attention matrix into fine-granular structured blocks \cite{lu2021sanger}}
    \label{fig:pruningSanger}
\end{figure}

Sanger \cite{lu2021sanger} is a technique to accelerate sparse attention models. Unlike ReLU, which leads to exact-zero activations, the softmax operation does not output exact zero at any position. Hence, the authors use a threshold to zero out small attention weights. This  creates an  attention mask  with unstructured sparsity. Also, the dynamic sparsity patterns are unstructured and hence, hardware-unfriendly. To deal with this issue, they first vertically split the attention mask into sub-matrices, of the same width as the vertical input ports of the PE array. Then, sub-rows with all zero elements are removed. After this, to keep comparable non-zero elements in each sub-row, those with more non-zero elements than PE array row-width are split into multiple rows. This creates structured blocks with an almost equal load. This process is illustrated in Figure \ref{fig:pruningSanger}.

Still, the exact distribution of non-zero elements can be random. This creates non-uniform data accesses in SDDMM and SpMM operations. The output of SDDMM is the score matrix, which is fed as input to SpMM. Since Q/K/V matrices are dense, but the score matrix is sparse, it leads to decoding overheads. They propose a \enquote{score-stationary} dataflow, which stores sparse scores in PE till the end of computations. In a 4*4 block with every row containing two non-zero values, they use a 4*2 PE array. Then, corresponding to the zero locations, bubbles are inserted, representing a ghost PE. This alleviates decoding overheads. For this, the input data registers are dynamically connected with the PEs based on the attention mask.  Further, this unifies SDDMM and SpMM operations and hence, saves chip area since a single SA can be used for both of them. Still, there are some differences between these two operations; hence, they reconfigure the SA between them. Their technique can prune the network to 0.08-0.27$\times$ model size without accuracy loss and attains high speedup over CPU, GPU and previous attention accelerators.

SALO \cite{shen2022salo} is a technique to speed up attention for tasks with long input sequences.  Their accelerator has a data scheduler and a spatial engine. The sparse attention pattern is characterized by the window size (w) of sliding window attention and dilation (d) of dilated window attention. The spatial engine has a PE array that maximizes data reuse by computing sliding window attention using diagonal connections. The engine also has a global PE row/column for computing global attention.

\begin{figure} [htbp]\centering
    \includegraphics[scale=0.43]{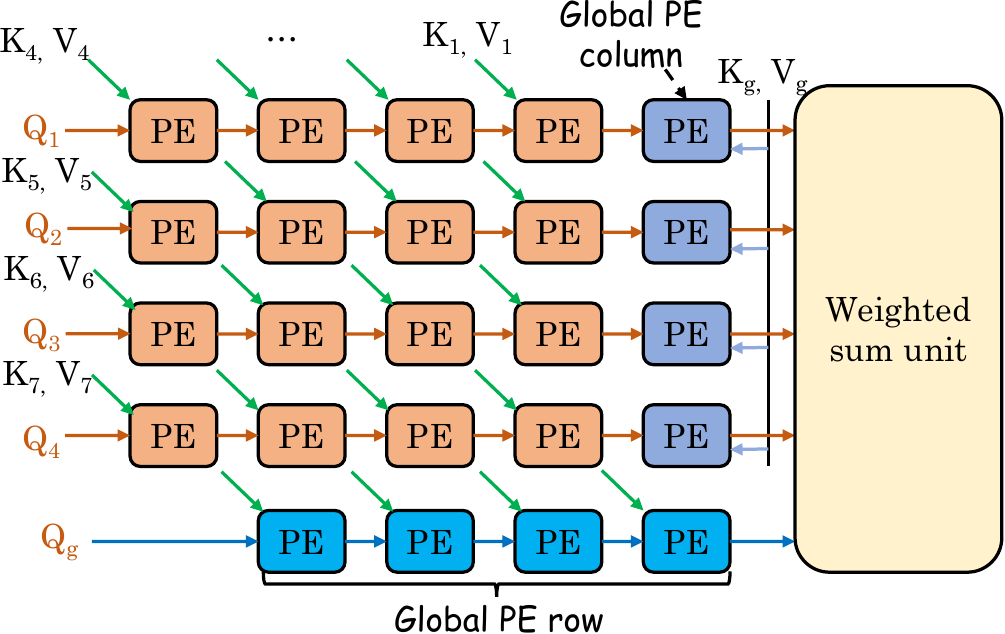}       
    \caption{Accelerator design of Shen et al. \cite{shen2022salo}}
    \label{fig:SaloDataFlow}
\end{figure}

Their datapath seeks to optimize data-reuse between (R1) various queries within sliding window attention and between (R2) sliding window attention and global attention. For (R1), they note that if $q_i$ attends to key vectors $[k_{i+a}, k_{i+a+1} \ldots k_{i+b}]$, then,   $q_{i+1}$ attends to key vectors $[k_{i+a+1}, k_{i+a+2} \ldots k_{i+b+1}]$. Hence, $b-a$ key vectors are reused. Their engine streams input $k$ and $v$ vectors diagonally and $q$ vectors horizontally, as shown in Figure \ref{fig:SaloDataFlow}. An incoming key vector $k_{i+b+1}$ replaces the outgoing vector  $k_{i+a}$ and the same strategy is used for replacing value vectors. 
For (R2), the K/Q of global tokens attend to all the K/Q of input sequences. Since SW attention works on all Q/K/V  elements, global attention needs no additional data access. Hence, both attention computations can be performed simultaneously.

They further propose data division and reorganization schemes to convert hybrid sparse attention into a pattern suitable for their accelerator. They divide matrices into tiles and compute attention in multiple passes. Specifically, they divide the sequence and the window based on the PE array dimensions. Window division splits an attention computation into multiple parts that are weighted and finally accumulated. This is done by the weighted sum unit. The dilated window attention cannot be directly executed on their dataflow. They propose reordering $Q$ matrix such that $q_{i}$ is grouped with $q_{i+d}$, $q_{i+2d}$, etc. This exposes reuse in dilated window attention with a period $d$, which is similar to sliding window attention. Hence, it can be accelerated on their spatial engine.

Their 2D SA performs computations in five stages: (1) The SA multiplies Q and K using output-stationary dataflow. (2) exponential is computed using the piece-wise linear function that the MAC unit can compute. (3) Exponential values of PEs in the same row are accumulated ($\sum Exp(S_{ij}$) to compute the denominator of softmax. At the last PE of a row, $(\sum Exp(S_{ij})^{-1}$ is computed and broadcast to PEs in a row. This avoids the need for a divider circuit. (4) Every PE computes $S'_{ij}= Exp(S_{ij} \times (\sum Exp(S_{ij})^{-1}$. (5) $PE_{ij}$ now receives value vector $v_j$. Using weight stationary dataflow, each $PE_{ij}$ computes $v_j \times S'_{ij}$, adds it to the previous partial sum and passes it to the next PE. The PE row output enters the weighted sum unit and is merged with the previous output. The global PE row computes the attention of queries of global tokens and hence, reuses the K/V vectors from the PE array. Analogously, the global PE column reuses Q vectors. Due to data-division, a single row/column suffices for handling global tokens found in real-world applications. On ViL and Longformer, their technique achieves higher speedup and energy efficiency compared to CPU and GPU.

ViTCoD \cite{you2022vitcod} technique seeks to mitigate data-movement and PE under-utilization issue in accelerators for ViT workloads. In the NLP transformers, the token count depends on the input, but ViTs use a fixed token count (e.g., 196). This avoids the need to predict sparse attention schemes and simplifies the design of hardware accelerators. Further, to keep the accuracy loss small, more than 90\% of attention maps in ViTs can be pruned with fixed sparsity layouts for all inputs, whereas in NLP transformers, only 50 to 70\% sparsity can be achieved even with dynamic sparsity layouts. However, this high sparsity in ViTs can create multiple challenges. In ViTs, the non-zero elements in sparse attention maps are found in the diagonal lines. This causes a large data movement and under-utilization of PEs.

\begin{figure} [htbp]\centering
    \includegraphics[scale=0.30]{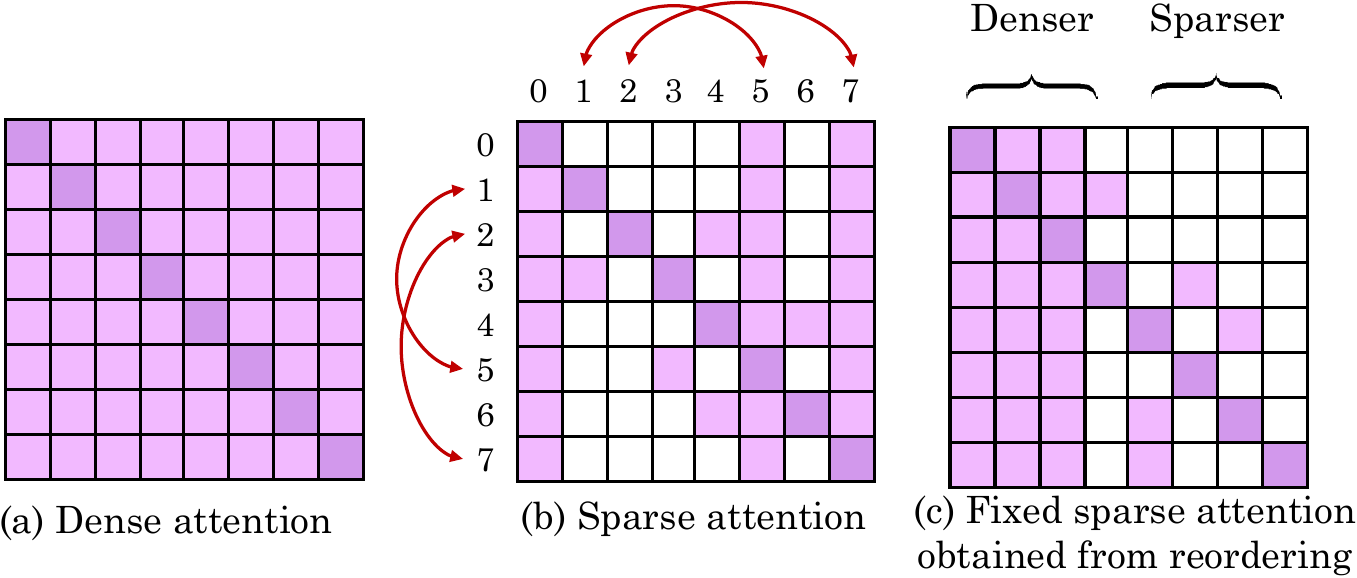}       
    \caption{Reordering of sparse attention weight matrices \cite{you2022vitcod} }
    \label{fig:SparseReordering}
\end{figure}

They first compute average and normalized attention maps by applying the pretrained models on all training images and then prune based on the amount of the remaining information. To lower the irregularity of resultant sparse maps, they find and cluster the query-key pairs into just two types: sparser or denser. Here, the tokens with more than $\theta$ non-zero elements are brought to the front as the denser pattern, and the remaining tokens are kept as the sparser pattern. This is shown in Figure \ref{fig:SparseReordering}. The denser layouts capture the tokens having a substantial correlation with the remaining tokens. The sparser layouts are those except the diagonal lines, where a majority of values are zero. This happens because a high correlation is found in neighbouring tokens. After reordering, they finetune the model to restore the accuracy. The above reordering operation enforces fixed sparse attention patterns, which do not change during finetuning. The sparser layouts suffer from high data-movement overhead. By exploiting the redundancy across attention heads, an auto-encoder compresses K and Q to 50\% of their original size. Evidently, the pruning+reordering technique reduces computations, and the auto-encoder trades computations for reducing memory accesses.

Their proposed accelerator has separate engines for processing dense and sparse layouts. The denser engine performs $Q.K^T$ (during sampled GEMM) and $S.V$ (during SpMM). The sparser engine performs the rest of the irregular computations, including activation functions. Since both denser and sparser engines work in parallel, they are likely to work on the same Q vectors. Hence, the sparser engine first queries the Q buffer of the denser engine and, only in case of a miss, accesses the off-chip memory.  
They use K-stationary dataflow, which reads K vectors and sequentially multiplies them with different Q vectors. It outputs attention values in a column-wise manner. This dataflow enables the full reuse of K vectors and requires a small on-chip storage for storing the interim results. This dataflow does not spatially map and multiply all Q/K features but only multiplies paired features depending on the non-zero indexes in S. 
Under 90\% attention sparsity, their accelerator provides much higher performance than CPU, GPU and previous accelerators.

\subsection{Block-circulant matrix for reducing weight storage}\label{sec:blockCirculant}
The block-circulant matrix (BCM) reduces the weight storage by replacing the original weight matrix with one or more blocks of circulant matrices, where every row/column is a cyclic reformulation of the other. This allows storing only one first row/column vector, called the index vector, instead of the entire matrix. Further, matrix vector multiplication can be replaced by an FFT operation, which reduces the complexity from  $O(b^2)$ to $O(b \log b)$, where $b$ is the row (or column) size of the circulant matrix.

Ftrans \cite{li2020ftrans} technique proposes an improved BCM approach for compressing transformer models. The previous works on BCM only take the first row/column as the index vector but fail to represent the remaining rows/columns. To better capture the parameter values and maintain accuracy, they take $P_{ij}$ = ($\sum_{j=1}^bW_{ij})/b$, where $W_{ij}$ is the circulant matrix. Thus, they also take the representation of the remaining rows/columns. The encoder/decoder layers, which account for two-thirds of total parameters, are stored on-chip, which reduces off-chip communication. The embedding layer has only one-third of the total parameters; hence, they store it off-chip. They develop two PEs for performing matrix-vector multiplication of different matrix sizes and one PE for performing FFT/IFFT. The matrix-vector multiplication PEs can also perform scaling and softmax in an overlapped manner with matrix multiplication. Computation of $KV^K$ etc., takes much larger time than the computation of MHA. To achieve a balanced pipelined execution of different layers, they use a scheduling algorithm to allocate more resources (e.g., the number of PEs) to the slowest layer.   Their technique brings up to 16X reduction in model size and also improves performance and energy efficiency over CPU.

%\section{Hardware  techniques for memory-optimization} \label{sec:HardwareMemory}

\section{Conclusion and future work} \label{sec:conclusion}

Transformers have become one of the most widely used architectures in NLP and computer vision domains due to their capability to capture long-range interdependencies. As the complexity of the transformers continues to grow, there is an increasing need for model compression and hardware optimization methods to accelerate these models. To address the enormous challenges faced by the transformers, researchers have developed several model enhancement techniques, including pruning, quantization, neural architecture search, distillation, and lightweight self-attention design. In this paper, we provide taxonomy and a comprehensive overview of the recently proposed inference optimization techniques for transformer-based networks. We discuss the optimization methods, both from transformer architecture and hardware perspectives. We plot the predictive performance against the number of parameters/FLOPs of several optimization techniques to provide meaningful insights. We conclude this paper by emphasizing the future directions in this rapidly evolving field of research. We hope our effort will create further interest in developing novel optimization methods to facilitate efficient inference of large models on a wide range of hardware.

The development of optimization techniques for efficient transformer inference has become an important area of research due to their broad applications across various architectures and domains. The current solutions have been effective in developing more robust and efficient algorithms for the family of transformer architectures. However, there is still ample room for improvement in both algorithmic and hardware optimization methods. There exist several challenges, such as scalability, interpretability, and fair comparison of the performance of different optimization techniques. The following list identifies the limitations and gaps in current research to enhance the development of efficient and effective methods:

\textbf{AutoML for Model Compression:} The automatic ML and NAS techniques are becoming mainstream for designing networks for several architectures, datasets and tasks. These methods have great potential to be applied to transformer model compression. This could involve exploring different search approaches, such as automatic identification of sparse weights and mixed precision quantization and obtaining optimal models in terms of accuracy and computational efficiency. While AutoML methods for CNNs has received significant attention, their use for CV transformers has not been studied in as much detail. There is a need to use AutoML for transformer model compression, to improve the scalability and efficiency and adapting them to a wider range of application areas.

\textbf{Hardware-aware NAS for CV transformers:} The HW-NAS methods for CNNs has been well-studied on various platforms, while the HW-NAS techniques for transformers is not fully explored, especially for computer vision. Although hardware-agnostic NAS algorithms on transformer architectures are useful for research, hardware-aware methods are crucial for real-time deployment. Thus, better hardware-aware search space and NAS algorithms are needed to improve accuracy and performance on multiple platforms.

\textbf{Transformer NAS benchmarks:} The computation resource demand is a major bottleneck for developing novel transformer NAS algorithms, as it requires training and evaluation of several neural architectures during the search process. Also, result reproducibility is challenging due to differences in search space and experiment setting; thus, comparing different methods becomes unfair. In this context, several NAS benchmarks \cite{chitty2023neural, ying2019bench} have been developed, which act as a look-up-table to obtain the validation accuracy without actually training them. HW-NAS benchmarks \cite{li2021hw} help non-hardware researchers by providing hardware performance metrics of the neural architectures on the selected platforms. Moving forward, transformer NAS and HW-NAS benchmarks will play a key role in developing better NAS algorithms.

\textbf{Transformer Inference Benchmarks:} The existing benchmark suites such as MLPerf  do not include transformer networks. There is a need to include SOTA transformer networks in such benchmark suites to accelerate and standardize the research on transformer.

\textbf{Evaluating existing optimization techniques on emerging topologies:}
The current optimization techniques have been shown to be effective on standard transformer models on widely used datasets and applications. Recently, transformer/attention modules have been utilized in emerging neural architecture topologies, such as image segmentation \cite{cao2021swin}, graph neural networks (GNN) \cite{dwivedi2020generalization}, graph convolutional network (GCN) \cite{dong2021dual}, 3D convolutional neural networks \cite{cao2017egocentric}, point clouds \cite{guo2021pct} and spiking transformer neural network \cite{mueller2021spiking}. Therefore, it is crucial to expand the scope of model compression technqiues to generalize over other tasks and emerging network topologies.

\textbf{Sparse transfer learning:} Sparse transfer learning refers to leveraging the sparsity pattern learned by a large and pre-trained teacher model to fine-tune a smaller model. In this way, the sparsity map learned by the large model can be transferred to the smaller model without the need for pruning and additional fine-tuning \cite{iofinova2022well}. This process is nascent and requires experimentation and investigation on several transformer architectures and hardware platforms.

\textbf{Reliability and security:} The model compression techniques can weaken fault-resilience and adversarial-robustness of a model by removing parameters that are unimportant for normal execution but are useful against perturbed input data or faulty weights. As transformers get deployed in critical applications, the future work must evaluate the adversarial-robustness and fault-tolerance of the proposed optimization methods.

\section*{Acknowledgement} The research reported in this paper was funded in part by the Philip and Virginia Sproul Professorship, and the National Science Foundation grants MRI 1726447 and MRI 2018594 at Iowa State University. This research was also supported by the Argonne Leadership Computing Facility, which is a DOE Office of Science User Facility supported under Contract DEAC02-06CH11357. All opinions, findings, and conclusions expressed are those of the authors.

{ 
\footnotesize
\bibliographystyle{IEEEtran}

\bibliography{Consolidated}
}

\begin{IEEEbiography}[{\includegraphics[width=1in,height=1.25in,clip,keepaspectratio]{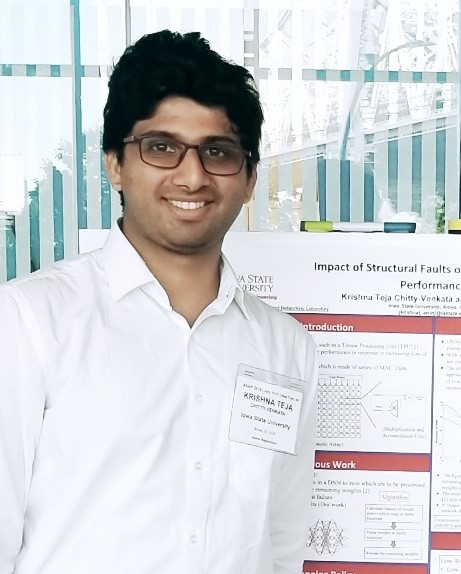}}]{Krishna Teja Chitty-Venkata} is a PhD student at Iowa State University, Ames, Iowa, USA. He received his Bachelor’s of Engineering degree in Electronics and Communication from University College of Engineering, Osmania University, Hyderabad, India in 2017. His research interests include designing network design algorithms and compression methods for Neural Network processing, such as Pruning, Quantization and Neural Architecture Search, for General and Special purpose hardware platforms. He interned at Argonne National Laboratory, Intel Corporation and AMD where he worked on various problems on efficient Deep Learning. 
\end{IEEEbiography}
\vskip -2.5\baselineskip plus -1fil

\begin{IEEEbiography}[{\includegraphics[width=1in,height=1.25in,clip,keepaspectratio]{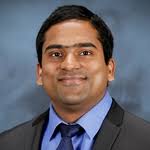}}]{Sparsh Mittal} received the B.Tech. degree from IIT, Roorkee, India, and the Ph.D. degree from Iowa State University (ISU), USA. He worked as a Postdoctoral Research Associate with the Oak Ridge National Laboratory (ORNL), USA. He was as an Assistant Professor at IIT Hyderabad. He is currently working as an Assistant Professor with the Department of Electronics and Communication Engineering, IIT Roorkee, and as a Joint Faculty with the Mehta Family School of Data Science and Artificial Intelligence. He was the graduating topper of his batch in B.Tech. degree and has received fellowship from ISU and performance award from ORNL. He has published more than 110 research papers at top venues. His research has been covered by insideHPC, HPCwire, Phys.org, and scientific computing.  He has received research funding from SERB, Intel, and SRC (USA).
\end{IEEEbiography}
\vskip -2.5\baselineskip plus -1fil
\begin{IEEEbiography}[{\includegraphics[width=1in,height=1.25in,clip,keepaspectratio]{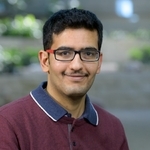}}]{Murali Emani} is an Assistant Computer Scientist in the Data Science group with the Argonne Leadership Computing Facility (ALCF) at Argonne National Laboratory. Prior to this, he was a Postdoctoral Research Staff Member at Lawrence Livermore National Laboratory. He obtained his Ph.D. from the School of Informatics, University of Edinburgh, UK in 2015. His research interests are in scalable machine learning, emerging HPC and AI architectures, AI for Science. He serves as a co-chair for MLPerf HPC group at MLCommons to benchmark large scale ML on HPC systems. He also co-leads the AI Testbed at ALCF to evaluate novel AI accelerators for scientific machine learning applications. Murali has organized workshops and participated in turorials that include benchmarking deep learning workloads on emerging hardware, MLPerf-Bench at MLSys’20, MLSys’21, ISPASS’20, ISPASS’21, ASPLOS’21. He has also co-chaired the MLPerf birds-of-a-feather sessions at SC’19, SC’20 and SC’21. 
\end{IEEEbiography}
\vskip -2.5\baselineskip plus -1fil
\begin{IEEEbiography}[{\includegraphics[width=1in,height=1.25in,clip,keepaspectratio]{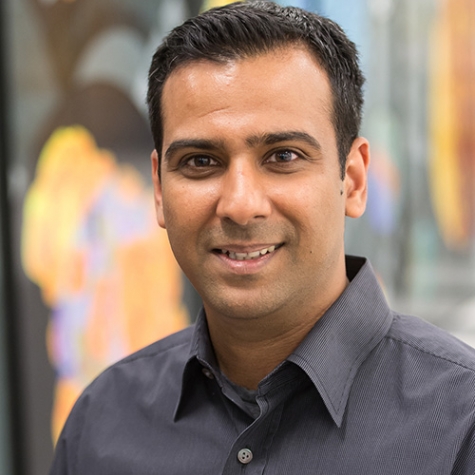}}]{Venkatram Vishwanath} is a computer scientist at Argonne National Laboratory. He is the Data Science Team Lead at the Argonne leadership computing facility (ALCF). His current focus is on algorithms, system software, and workflows to facilitate data-centric applications on supercomputing systems. His interests include scientific applications, supercomputing architectures, parallel algorithms and runtimes, scalable analytics and collaborative workspaces. He has received best papers awards at venues including HPDC and LDAV, and a Gordon Bell winner. Vishwanath received his Ph.D. in computer science from the University of Illinois at Chicago in 2009.
\end{IEEEbiography}
\vskip -2.5\baselineskip plus -1fil
\begin{IEEEbiography}[{\includegraphics[width=1in,height=1.25in,clip,keepaspectratio]{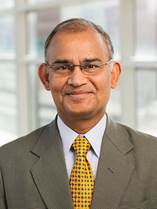}}]{Arun K. Somani} received M.S. and Ph.D. degrees in Electrical Engineering from McGill University, Montreal, in 1983 and 1985, respectively. He served as a Scientific Officer with the Government of India, New Delhi, and a Faculty Member with the University of Washington, Seattle. He is currently an Anson Marston Distinguished Professor of Electrical and Computer Engineering with Iowa State University. His research interests are in the areas of computer system design, architecture, fault tolerant computing, computer interconnection networks, optical networking, reconfigurable and parallel computer systems. He served as IEEE Distinguished Visitor, IEEE Distinguished Tutorial Speaker, and an IEEE Communication Society Distinguished Visitor. He delivered several keynote speeches, and distinguished and invited talks all over the world. He is a Life Fellow of IEEE, Distinguished Engineer of ACM, Eminent Engineer of Tau Beta Pi, and a fellow of AAAS.
\end{IEEEbiography}

%\section{Future Work} \label{sec:Future_work}

\end{document}